\documentclass{article}

 \usepackage[preprint]{neurips_2026}

\usepackage[utf8]{inputenc} %
\usepackage[T1]{fontenc}    %
\usepackage{hyperref}       %
\usepackage{url}            %
\usepackage{booktabs}       %
\usepackage{multirow}       %
\usepackage{amsfonts}       %
\usepackage{amssymb}        %
\usepackage{nicefrac}       %
\usepackage{microtype}      %
\usepackage{xcolor}         %
\usepackage{placeins}
\usepackage{amsmath}
\usepackage{graphicx}
\usepackage{wrapfig}        %
\usepackage{tikz}
\usetikzlibrary{positioning, arrows.meta, calc, shapes.geometric}

\title{Flexible Intensities Matter: \\ A Comprehensive Re-evaluation of Hawkes Processes and Neural Temporal Point Processes}

\author{%
  Hendrik Mehrtens\textsuperscript{1,2} \qquad
  Florian Büttner\textsuperscript{1,2,*} \qquad
  Oliver Stegle\textsuperscript{1,3,4,*} \\[4pt]
  \normalfont\small \textsuperscript{1}German Cancer Research Center (DKFZ), Heidelberg, Germany \quad
  \textsuperscript{2}Goethe-Universität Frankfurt, Germany \\
  \normalfont\small \textsuperscript{3}European Molecular Biology Laboratory (EMBL), Heidelberg, Germany \\
  \normalfont\small \textsuperscript{4}Ruprecht-Karls-Universität Heidelberg, Germany
}

\begin{document}

\newcommand{\tb}{\textbf}
\newcommand{\tu}{\underline}
\newcommand{\ti}{\textit}
\newcommand{\codeurl}{\url{https://github.com/PMBio/torchPP}}

\maketitle
{\let\thefootnote\relax\footnotetext{\raggedright\textsuperscript{*}These authors contributed equally. Correspondence: \texttt{hendrikalexander.mehrtens@dkfz-heidelberg.de}, \texttt{florian.buettner@dkfz-heidelberg.de}, \texttt{o.stegle@dkfz-heidelberg.de}.}}

\begin{abstract}
Temporal point processes (TPPs) are the canonical model class for event sequences in continuous time, with applications ranging from seismology to electronic health records. Recent benchmarks report that neural TPPs substantially outperform their classical counterparts, such as Hawkes processes. However, these comparisons set flexible neural models against Hawkes processes in their most restricted forms and covered only a narrow range of domains. We revisit this comparison under autoregressive next-event prediction, evaluating Poisson and Hawkes processes of increasing flexibility against neural TPPs on five established benchmarks and two large-scale medical tasks derived from UK Biobank disease histories and MIMIC-IV ICU records. First, once Hawkes processes are equipped with flexible, learnable baselines and kernels, the previously reported gap to neural TPPs largely closes. Second, neural TPPs are likewise limited by their intensity parametrization: a Transformer with a spline-based intensity head outperforms previous intensity-based neural TPPs and matches or exceeds the intensity-free model. Decomposing the log-likelihood into time and type terms shows that this gain lies almost entirely in modeling \emph{when} events occur ($\log p(t)$) rather than \emph{which} type occurs ($\log p(d\mid t)$). Third, only on MIMIC-IV do neural TPPs retain an advantage in predicting event types, underscoring the value of complex datasets for evaluating TPPs. We release \textsc{TorchPP}, an open-source PyTorch framework with unified implementations of classical and neural TPPs.
\end{abstract}

\section{Introduction}
\label{sec:intro}

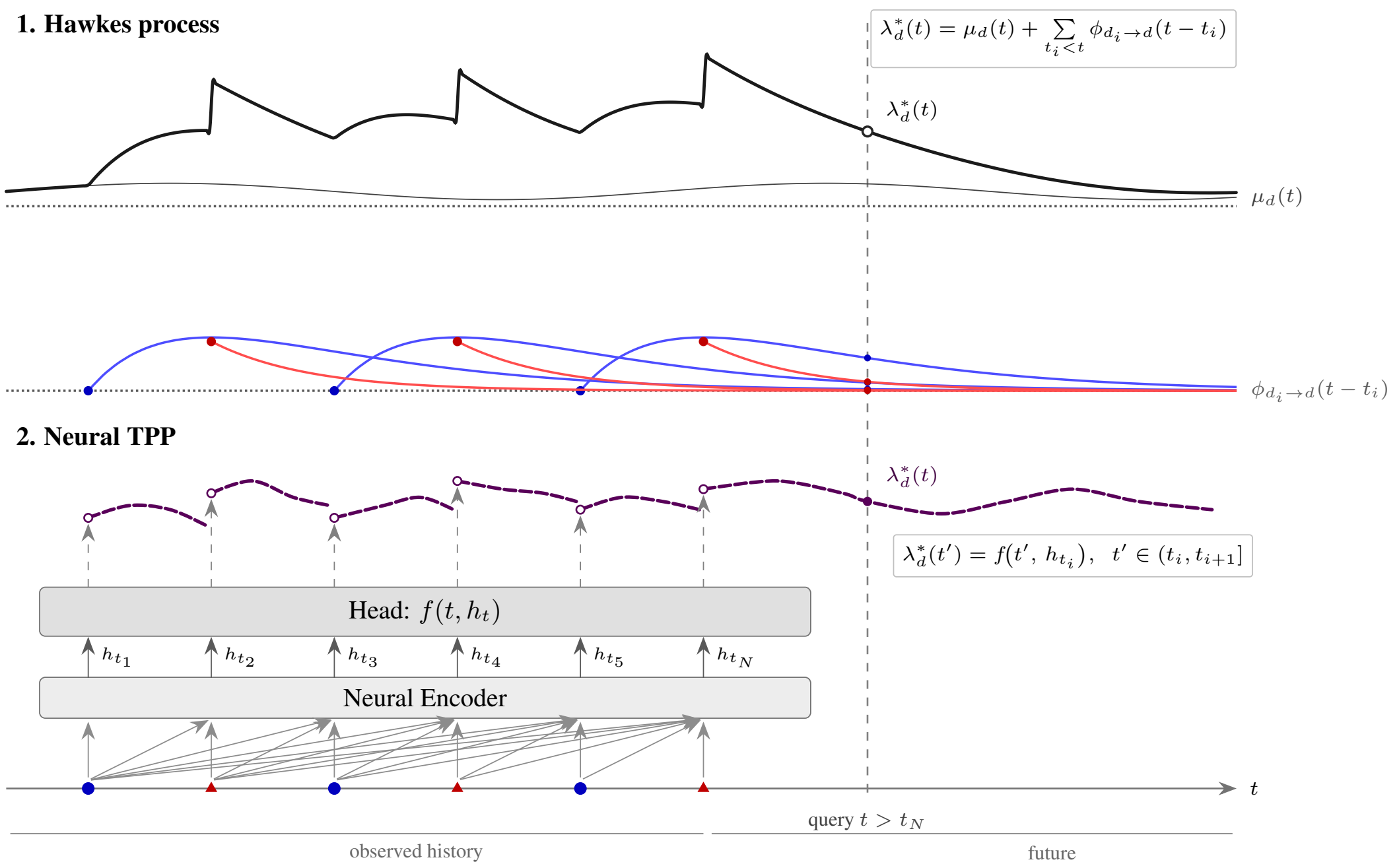
\begin{figure}[t]
\centering
\resizebox{\textwidth}{!}{%
\begin{tikzpicture}[
  >={Stealth[length=2.2mm]},
  font=\small,
  evtA/.style={circle, fill=blue!75!black, draw=blue!75!black, inner sep=1.4pt},
  evtB/.style={regular polygon, regular polygon sides=3, fill=red!75!black, draw=red!75!black, inner sep=1.0pt, scale=0.7},
  spawnA/.style={circle, fill=blue!75!black, draw=blue!75!black, inner sep=1.0pt},
  spawnB/.style={circle, fill=red!75!black, draw=red!75!black, inner sep=1.0pt},
  jumpdot/.style={circle, fill=black!85, inner sep=0.7pt},
  axline/.style={->, draw=black!55, semithick},
  baseline/.style={densely dotted, draw=black!65, thick},
  qline/.style={dashed, draw=black!50, semithick},
  kernelA/.style={draw=blue!70, thick},
  kernelB/.style={draw=red!70, thick},
  total/.style={draw=black!90, line width=1.1pt, line cap=round},
  freeform/.style={draw=violet!70!black, line width=1.2pt, line cap=round, dash pattern=on 4pt off 1.5pt},
  archbox/.style={rectangle, rounded corners=2pt, draw=black!55, fill=black!7,
                  align=center, font=\small, inner sep=4pt},
  feed/.style={->, draw=black!45, thin, shorten >=1pt, shorten <=1pt},
  flow/.style={->, draw=black!65, thick},
  paneltag/.style={font=\bfseries\small, anchor=west},
  eqbox/.style={rectangle, rounded corners=1pt, draw=black!25, fill=white,
                inner sep=3pt, font=\scriptsize}
]

\def\xmin{0.0}
\def\xmax{15.0}
\def\xt{10.5}
\def\xlast{8.5}     %
\def\mufun#1{0.18 + 0.10*sin(45*#1)}
\def\aA{0.65}
\def\tauA{1.5}
\def\aB{0.60}
\def\betaB{0.85}
\def\phiA#1{\aA*(max(0,#1)/\tauA)*exp(1-max(0,#1)/\tauA)}
\def\phiB#1{\aB*exp(-\betaB*max(0,#1))*(#1>=0?1:0)}

\draw[axline] (\xmin, 0) -- (\xmax, 0) node[right=1pt, font=\scriptsize]{\(t\)};

\node[evtA] at (1.0, 0) {};
\node[evtB] at (2.5, 0) {};
\node[evtA] at (4.0, 0) {};
\node[evtB] at (5.5, 0) {};
\node[evtA] at (7.0, 0) {};
\node[evtB] at (\xlast, 0) {};

\node[font=\scriptsize, anchor=north, text=black!75] at (\xt, -0.18) {query \(t > t_N\)};

\draw[draw=black!45, thin] (0.05, -0.55) -- (\xlast, -0.55);
\draw[draw=black!45, thin] (\xlast+0.10, -0.55) -- (\xmax-0.05, -0.55);
\node[font=\scriptsize, text=black!60] at (5.0, -0.78) {observed history};
\node[font=\scriptsize, text=black!60] at (12.75, -0.78) {future};

\draw[qline] (\xt, -0.05) -- (\xt, 9.4);

\node[archbox, minimum width=9.4cm, minimum height=0.45cm, anchor=south west]
      (enc) at (0.4, 0.85) {Neural Encoder};
\node[archbox, fill=black!12, minimum width=9.4cm, minimum height=0.45cm, anchor=south west]
      (head) at (0.4, 1.85) {Head: \(f(t, h_t)\)};

\foreach \xi [count=\ii] in {1.0, 2.5, 4.0, 5.5, 7.0, 8.5} {
  \foreach \xj [count=\jj] in {1.0, 2.5, 4.0, 5.5, 7.0, 8.5} {
    \ifnum\ii>\jj\relax\else
      \draw[feed] (\xi, 0.10) -- (\xj, 0.85);
    \fi
  }
}

\foreach \x/\lab in {1.0/{h_{t_1}}, 2.5/{h_{t_2}}, 4.0/{h_{t_3}}, 5.5/{h_{t_4}}, 7.0/{h_{t_5}}, 8.5/{h_{t_N}}} {
  \draw[flow, thin] (\x, 1.35) -- (\x, 1.85) node[midway, right=1pt, font=\tiny]{\(\lab\)};
}

\draw[freeform] plot[smooth, tension=0.6] coordinates {(1.0, 3.30) (1.5, 3.45) (2.0, 3.40) (2.45, 3.20)};
\draw[freeform] plot[smooth, tension=0.6] coordinates {(2.5, 3.60) (3.0, 3.75) (3.5, 3.55) (3.95, 3.45)};
\draw[freeform] plot[smooth, tension=0.6] coordinates {(4.0, 3.30) (4.6, 3.45) (5.05, 3.55) (5.45, 3.40)};
\draw[freeform] plot[smooth, tension=0.6] coordinates {(5.5, 3.75) (6.05, 3.65) (6.55, 3.60) (6.95, 3.50)};
\draw[freeform] plot[smooth, tension=0.6] coordinates {(7.0, 3.40) (7.5, 3.55) (8.0, 3.50) (8.45, 3.40)};
\draw[freeform] plot[smooth, tension=0.55] coordinates {
  (8.5, 3.65) (9.4, 3.75) (10.2, 3.60) (\xt, 3.50) (11.4, 3.35) (12.2, 3.50) (13.0, 3.65) (13.8, 3.50) (14.7, 3.40)
};

\foreach \x/\y in {1.0/3.30, 2.5/3.60, 4.0/3.30, 5.5/3.75, 7.0/3.40, 8.5/3.65} {
  \node[circle, draw=violet!70!black, fill=white, line width=0.6pt, inner sep=1.0pt] at (\x, \y) {};
}

\foreach \x/\y in {1.0/3.30, 2.5/3.60, 4.0/3.30, 5.5/3.75, 7.0/3.40, 8.5/3.65} {
  \draw[flow, thin, dashed, draw=black!50, shorten >=2pt] (\x, 2.45) -- (\x, \y);
}

\node[circle, draw=violet!70!black, fill=violet!70!black, inner sep=1.0pt] at (\xt, 3.50) {};
\node[font=\scriptsize, anchor=south west, text=violet!50!black] at (\xt+0.10, 3.55) {\(\lambda_d^*(t)\)};

\node[paneltag, anchor=west] at (\xmin, 4.30) {2. Neural TPP};

\node[eqbox, anchor=north east] at (\xmax+0.2, 3.10) {%
  \(\lambda_d^*(t') = f\!\big(t',\, h_{t_i}\big),\;\; t' \in (t_i, t_{i+1}]\)};

\pgfmathsetmacro{\kbase}{4.85}
\draw[baseline] (\xmin, \kbase) -- (\xmax, \kbase);
\node[font=\scriptsize, anchor=west, text=black!65] at (\xmax+0.05, \kbase) {\(\phi_{d_i \to d}(t-t_i)\)};

\foreach \tev in {1.0, 4.0, 7.0} {
  \draw[kernelA, smooth, samples=80, domain=\tev:\xmax]
    plot (\x, {\kbase + \phiA{\x-\tev}});
  \node[spawnA] at (\tev, \kbase) {};
}
\foreach \tev in {2.5, 5.5, 8.5} {
  \draw[kernelB, smooth, samples=80, domain=\tev:\xmax]
    plot (\x, {\kbase + \phiB{\x-\tev}});
  \node[spawnB] at (\tev, {\kbase + \aB}) {};
}

\pgfmathsetmacro{\sbase}{7.10}
\draw[baseline] (\xmin, \sbase) -- (\xmax, \sbase);
\draw[draw=black!75, thin] plot[smooth, samples=80, domain=\xmin:\xmax]
  (\x, {\sbase + \mufun{\x}});
\node[font=\scriptsize, anchor=west, text=black!75] at (\xmax+0.05, {\sbase + \mufun{\xmax}}) {\(\mu_d(t)\)};

\draw[total, smooth, samples=400, domain=\xmin:\xmax]
  plot (\x, {\sbase + \mufun{\x}
    + \phiA{\x-1.0} + \phiB{\x-2.5}
    + \phiA{\x-4.0} + \phiB{\x-5.5}
    + \phiA{\x-7.0} + \phiB{\x-8.5}});

\foreach \tev/\type in {1.0/A, 2.5/B, 4.0/A, 5.5/B, 7.0/A, 8.5/B}{
  \ifthenelse{\equal{\type}{A}}{
    \pgfmathsetmacro{\yval}{\kbase + \phiA{\xt-\tev}}
    \node[spawnA, scale=0.7] at (\xt, \yval) {};
  }{
    \pgfmathsetmacro{\yval}{\kbase + \phiB{\xt-\tev}}
    \node[spawnB, scale=0.7] at (\xt, \yval) {};
  }
}
\pgfmathsetmacro{\sumAtT}{\sbase + \mufun{\xt}
  + \phiA{\xt-1.0} + \phiB{\xt-2.5}
  + \phiA{\xt-4.0} + \phiB{\xt-5.5}
  + \phiA{\xt-7.0} + \phiB{\xt-8.5}}
\node[circle, draw=black!85, fill=white, line width=0.8pt, inner sep=1.2pt] at (\xt, \sumAtT) {};
\node[font=\scriptsize, anchor=south west] at (\xt+0.10, \sumAtT) {\(\lambda_d^*(t)\)};

\node[paneltag, anchor=west] at (\xmin, 9.30) {1. Hawkes process};

\node[eqbox, anchor=north east] at (\xmax, 9.50) {%
  \(\lambda_d^*(t) = \mu_d(t) + \sum\limits_{t_i < t}\phi_{d_i \to d}(t - t_i)\)};

\end{tikzpicture}}
\caption{Conceptual comparison of Hawkes processes (1) and neural TPPs (2), illustrated on an event sequence with two event types (\textcolor{blue!75!black}{\(\bullet\)}~type~A, \textcolor{red!75!black}{\(\blacktriangle\)}~type~B).
\textbf{Top (Hawkes process).} Each past event interacts via a kernel response \(\phi_{d_i \to d}(t - t_i)\) (colored curves) on top of a (possibly time-varying) baseline \(\mu_d(t)\) (thin solid line).
The total intensity (upper sub-panel) is the sum of baseline and active kernels.
At any query time \(t\), the intensity is thus the baseline plus the sum of kernel contributions from all past events.
\textbf{Bottom (Neural TPP).} A neural encoder processes the same event history and outputs a hidden state at each event location. The head of the neural TPP projects this encoding to an intensity function.
Conceptually, this lets neural TPPs capture higher-order, non-additive interactions between events, at the cost of the interpretable decomposition of the Hawkes process.}
\label{fig:hawkes_vs_neural}
\end{figure}

Temporal point processes (TPPs) provide a principled framework for modeling irregular event sequences in continuous time.
Applications range from seismology~\citep{ogata1988earthquake} and social media analysis~\citep{zhou2013triggering} to healthcare, where a patient's medical history forms a sequence of clinical events~\citep{Delphi}.
  
The Hawkes process~\citep{hawkes1971spectra} is a classical TPP that decomposes the intensity into a baseline (exogenous) component and triggering kernels (endogenous) through which past events excite future ones.

In recent years, neural TPPs have gained traction. They replace this additive structure with learned representations: a neural encoder summarizes the event history, and a flexible output head parametrizes either the conditional intensity or the inter-event time distribution~\citep{shchur2021review, xue2024easytpp}.
See Figure~\ref{fig:hawkes_vs_neural} for a conceptual visualization.

Neural TPPs have repeatedly been reported to outperform Hawkes baselines~\citep{RMTPP, nhp2017, xiao2017intensity, SAHP, shchur2020tritpp, soen2021unipoint}, most recently and by  substantial margins in the EasyTPP benchmark~\citep{xue2024easytpp}. 

However, the Hawkes baselines in these comparisons were typically restricted to a constant baseline and exponential kernels, far less flexible than the neural TPPs they were compared against.  These evaluations also cover a narrow range of domains and omit health data entirely, despite growing interest in sequence models for electronic health records~\citep{Delphi, MOTOR, behrt, medbert, ethos}.

In this work, we systematically re-evaluate Poisson processes, Hawkes processes, and neural TPPs on established benchmark datasets and on two large-scale medical tasks: disease histories from the UK Biobank~\citep{sudlow2015ukb} and ICU records from MIMIC-IV~\citep{mimic-iv}.

To our knowledge, this is the first systematic comparison of Hawkes processes with \textbf{flexible} baselines and kernels and \textbf{inhibitive} variants against modern neural TPPs under a unified protocol. Prior benchmarks \citep{xue2024easytpp, bosser2023predictive} considered Hawkes processes with a constant baseline and an exponential kernel \citep{hawkes1971spectra}, while \citet{zhao2025embedded} used a neural but purely excitatory kernel.
We deliberately restrict the set of classical models to Poisson and Hawkes processes; Cox-type models such as log-Gaussian Cox processes are beyond the scope of this work. 
On the neural side, we consider RNN- and Transformer-based intensity models with output heads of increasing flexibility, as well as an intensity-free model.
Flow-, diffusion-, and state-space-based TPPs are out of scope.
The code is available at \codeurl, together with checkpoints for the standard benchmarks.

\paragraph{Our contributions are as follows:}
\begin{itemize}
\item We show that the previously reported gap between Hawkes processes and neural TPPs is largely an artifact of under-specified baselines: with spline baselines, spline kernels, and an inhibitive variant, Hawkes processes match or exceed the performance of rigidly parametrized neural TPPs (THP, SAHP, NHP) and are competitive with the flexible neural models (IntensityFree, NeuralSplineTPP), indicating that these benchmarks are largely explained by pairwise interactions.
\item We identify the flexibility of the intensity parametrization as a main driver of performance: a Transformer with a spline-based intensity head (\textbf{NeuralSplineTPP}) outperforms softplus-headed intensity models (\textbf{THP}, \textbf{SAHP}, \textbf{NHP}) and performs on par with \textbf{IntensityFree}~\citep{IntensityFree}. Unlike the softplus-headed models, it has a closed-form cumulative intensity that allows exact sampling, and unlike IntensityFree, its type distribution depends on the event time.
\item We show that more flexible intensities improve the modeling of \emph{when} events occur, not \emph{which} type occurs. Decomposing the joint log-likelihood into a time term \(\log p(t)\) and a time-conditional type term \(\log p(d \mid t)\), and sweeping the number of spline knots, we find that added flexibility improves \(\log p(t)\) but leaves \(\log p(d \mid t)\) essentially unchanged. Across models, too, the standard benchmarks separate them far more on time prediction than on type prediction.
\item We evaluate both model families on two large-scale medical evaluation tasks: a 1{,}257-type terminating disease-history task derived from the UK Biobank, and a 117-type ICU-event task derived from MIMIC-IV. On UKB, flexible Hawkes processes nearly match the best neural TPP. On MIMIC-IV, the flexible neural TPP opens a gap in both joint log-likelihood and time-conditional accuracy.
\item We release \textsc{TorchPP}, a modular PyTorch framework for classical and neural TPPs. It supports terminating and right-censored sequences, exact sampling, and both analytical and numerical computation of the cumulative intensity (Appendix~\ref{sec:framework}).
\end{itemize}

\section{Background and Related Work}
\label{sec:background}

\subsection{Temporal Point Processes}
\label{sec:background/tpps}

A multivariate temporal point process over \(D\) event types on the interval \([0, T]\) is characterized by its conditional intensity function \(\boldsymbol{\lambda}^*(t) = (\lambda_1^*(t), \ldots, \lambda_D^*(t))\), where \(\lambda_d^*(t)\) gives the instantaneous rate of type-\(d\) events conditioned on the history \(\mathcal{H}_t = \{(t_i, d_i) : t_i < t\}\).
The log-likelihood of an observed sequence decomposes as:
\begin{align}
    \log p(\mathcal{H}_T)  &= \sum_{i=1}^{N} \log p(t_i, d_i \mid \mathcal{H}_{t_i}) + \log S^*(T - t_N) = \sum_{i=1}^{N} \log \lambda_{d_i}^*(t_i) - \int_0^T \sum_{d=1}^{D} \lambda_d^*(t) \, dt,
    \label{eq:loglik}
\end{align}
where the first term rewards high intensity at the observed event times, and the second, the cumulative intensity (CI), penalizes intensity throughout the observation window, accounting for the intervals in which no events occur. See Appendix~\ref{app:math_details} for a detailed derivation.

\paragraph{Poisson Processes}
\label{sec:background/poisson}
Poisson processes assume a history-independent intensity \(\lambda_d^*(t) = \mu_d(t)\), where \(\mu_d\) is a non-negative function of time. The homogeneous Poisson process corresponds to the assumption of a constant intensity function \(\mu_d(t) = \mu_d\); more expressive parametrizations, e.g.\ splines or neural networks, are equally admissible.

\paragraph{Hawkes Processes}
\label{sec:background/hawkes}
A Hawkes process~\citep{hawkes1971spectra} corresponds to an intensity function that is parametrized as a sum of a baseline rate and pairwise interactions from past events:
\begin{equation}
    \lambda_d^*(t) = \mu_d(t) + \sum_{t_i < t} \phi_{d_i \to d}(t - t_i).
    \label{eq:hawkes}
\end{equation}
Here, \(\mu_d(t)\) is the baseline intensity of type \(d\), i.e.\ the intensity of a Poisson process, and \(\phi_{d' \to d}(\cdot)\) is the triggering kernel from type \(d'\) to type \(d\). The classical choices are a constant baseline \(\mu_d(t) = \mu_d\), corresponding to a homogeneous Poisson process, and an exponential kernel \(\phi_{d' \to d}(\Delta t) = \alpha_{d'd}\, \beta_{d'd}\, e^{-\beta_{d'd} \Delta t}\); together, they yield a closed-form cumulative intensity~\citep{hawkes1971spectra}.

\paragraph{Neural Temporal Point Processes}
\label{sec:background/ntpps}
In their most common form, neural TPPs~\citep{shchur2021review, lin2025survey} replace the additive pairwise Hawkes structure with a flexible intensity function parametrized by learned representations. A history encoder maps \(\mathcal{H}_t\) to a context vector \(\mathbf{h}(t)\), which is projected to an intensity function by an output head. Common choices for encoders include recurrent networks (RMTPP~\citep{RMTPP}, NHP~\citep{nhp2017}) and attention-based architectures (THP~\citep{THP}, SAHP~\citep{SAHP}, AttNHP~\citep{AttNHP}). Most output heads apply a clamping function such as Softplus to enforce non-negativity, which precludes the closed-form calculation of the cumulative intensity and hence requires numerical integration during training and evaluation.
Alternatively, the cumulative intensity itself can be parametrized by a monotone network~\citep{omi2019fullynn, zhou2023autoint}.
Other approaches specify the distribution of the next event directly, e.g.\ via mixture models~\citep{IntensityFree} or triangular maps~\citep{shchur2020tritpp}, or generate whole sequences via flow matching~\citep{EventFlow, ludke2026editpp} or diffusion~\citep{DiffusionTPP, ludke2023addthin, mukherjee2026adiff4tpp}. Neural ODEs~\citep{song2024decoupled} and state-space models~\citep{S2P2} are further families. Of these, we include the intensity-free mixture model \textbf{IntensityFree}~\citep{IntensityFree} as a strong density-based baseline. The flow, diffusion, and state-space families remain complementary to the intensity/density-flexibility axis studied here and are left to future work.

See \citet{shchur2021review} for a survey of neural TPP architectures, and Figure~\ref{fig:hawkes_vs_neural} for a conceptual comparison with Hawkes processes.
While neural TPPs have the capacity to learn higher-order interactions beyond pairwise kernels, they sacrifice the intrinsic interpretability of classical methods such as Hawkes processes.

\subsection{Existing Benchmarks of TPPs}
\label{sec:background/benchmarks}
Few benchmarks have systematically compared different neural TPPs. Among these, EasyTPP~\citep{xue2024easytpp} is widely used and has evaluated several recent TPP architectures across five standard datasets (see Section~\ref{sec:datasets-experiments/datasets}). 
However, their Hawkes baseline is highly restricted: it uses a constant baseline intensity, exponential kernels with fixed, non-learned decay rates, and no inhibition. These choices severely limit its expressiveness (see Appendix~\ref{app:tick} for a reproduction).
\citet{bosser2023predictive} benchmark combinations of neural encoders and decoders, including an exponential-kernel Hawkes decoder, and \citet{lin2025survey} compare neural TPP components empirically. Neither considers baselines with flexible Hawkes kernels.

Another related benchmark is~\citet{DanBench},  a multi-modal TPP benchmark for text and video data, which, however, targets a different domain and model class. Earlier work~\citep{RMTPP, nhp2017, xiao2017intensity, SAHP, shchur2020tritpp, soen2021unipoint} has also compared neural TPPs with Hawkes processes. As above, these comparisons used only exponential-kernel Hawkes processes~\citep{hawkes1971spectra} or self-correcting processes~\citep{SelfCorrecting}, both of which have a fixed parametric form.

\section{Evaluation Design}

\subsection{Considered Methods}
\label{sec:models}
We systematically compare Hawkes processes and neural TPPs with increasingly flexible intensity parametrizations, under a unified training, tuning, and evaluation protocol, implemented in our \textsc{TorchPP} framework. For full model details see Appendix~\ref{app:model_details}.

\subsubsection{Hawkes Processes}
\label{sec:models/hawkes}
Previous TPP benchmarks have largely considered Hawkes baselines restricted to a constant baseline and an exponential kernel. We consider a more expressive family of Hawkes processes with flexible baseline intensities and kernels, including variants that capture inhibitive  effects.

\paragraph{Baseline intensity.}
For the per-type baseline intensity \(\mu_d(t)\), we consider two parametrizations:
\begin{enumerate}
    \item ~a constant baseline \(\mu_d(t) = \mu_d\) (\textbf{homogeneous Poisson}),
    \item ~a linear spline over time since sequence start, with learnable knot heights (\textbf{spline Poisson}), which captures non-stationary intensities; the number and spacing of the knots are hyperparameters.
\end{enumerate}

\paragraph{Triggering kernel.}
For the kernels \(\phi_{d' \to d}(\Delta t)\), we  compare:
\begin{enumerate}
    \item ~the exponential kernel \(\phi_{d' \to d}(\Delta t) = \alpha_{d'd} \beta_{d'd} e^{-\beta_{d'd} \Delta t}\) (\textbf{exponential kernel}),
    \item a learnable linear spline over the time lag \(\Delta t\) for each type pair \((d', d)\) (\textbf{spline kernel}), constrained to be non-negative via \textit{Softplus} in the non-inhibitive models.
\end{enumerate}
Both kernels admit closed-form cumulative intensities.

Based on these triggering kernels, we consider the following Hawkes processes with progressively more flexible baseline intensities and kernels:
\begin{enumerate}
    \item~constant baseline + exponential kernel  \(\to\) \textbf{Const./Exp Hawkes}
    \item~constant baseline + spline kernel  \(\to\) \textbf{Const./Spline Hawkes}
    \item~spline baseline + exponential kernel \(\to\)  \textbf{Spline/Exp Hawkes}
    \item~spline baseline + spline kernel \(\to\) \textbf{Spline/Spline Hawkes} 
\end{enumerate}

\paragraph{Inhibitive variants.}
The standard Hawkes intensity (Equation~\ref{eq:hawkes}) cannot model inhibitive interactions because intensities, as instantaneous rates, must remain non-negative.
To capture inhibitive effects, we wrap the sum of baseline and kernel contributions in a non-negative clamping function~\citep{bremaud1996nonlinear}. We compared ReLU and Softplus, and found Softplus more stable and better-performing. The resulting models lack a closed-form cumulative intensity. We estimate it numerically with a stochastic trapezoidal rule.
We report the \textbf{Inhib. Const./Exp Hawkes} and \textbf{Inhib. Spline/Spline Hawkes} variants.

\subsubsection{Neural TPPs}
\label{sec:models/neural_tpps}

As our primary aim is a meaningful comparison of Hawkes processes and neural TPPs, we select a representative set of neural TPPs based on the EasyTPP benchmark~\citep{xue2024easytpp}, which allows direct comparison with previously reported results. For a comprehensive comparison of alternative neural TPP architectures, we refer the reader to the EasyTPP paper.
The neural TPPs we consider share a common structure: a neural history encoder \(\mathrm{enc}\) that maps \(\mathcal{H}_t\) to a context vector \(h_t\), and an output head \(f(t,h_t)\) that maps this context to per-type intensity functions:
\begin{equation*}
    \lambda_d^*(t) = f\!\big(t,\, \mathrm{enc}(\mathcal{H}_t)\big).
\end{equation*}
Most of the models we consider use a Transformer encoder with a joint time/type event embedding and differ mainly in the design of the output head \(f\), which maps the hidden state to the conditional intensity (or, for IntensityFree, to the inter-event time distribution). Hyperparameters for all models are tuned on the validation set (Appendix~\ref{app:hyperparams}). We also include two models with a different structure: \textbf{NHP}, which uses a continuous-time recurrent encoder, and \textbf{IntensityFree}, which models the inter-event time distribution directly rather than the intensity. Mirroring the Hawkes variants above, we list the models in order of increasing flexibility:
\begin{itemize}
    \item \textbf{NeuralConstTPP}: the head predicts a history-dependent but time-constant intensity \(\lambda_d^*(t) = c_d(\mathcal{H}_t)\), yielding an analytical CI. This model corresponds to the simplest intensity parametrization that does not allow for time-varying effects.
\item \textbf{THP / SAHP}: re-implementations of the models from~\citep{THP, SAHP}, serving as representatives of common but inflexible neural TPP architectures. Both models employ a Softplus-wrapped intensity function on top of the encoder state, requiring numerical integration of the cumulative intensity. For a comparison with the original implementations, see Appendix~\ref{app:equivalence}. %

\item \textbf{NHP}~\citep{nhp2017}: a continuous-time LSTM encoder with a softplus intensity head. It keeps the numerically integrated softplus head of \textbf{THP}/\textbf{SAHP} but replaces the attention encoder with a recurrent one, probing the effect of the encoder architecture.

\item \textbf{NeuralSplineTPP}: to match the kernel flexibility introduced for the Hawkes family, we equip the Transformer encoder with a spline-based intensity head. The head predicts per-type linear splines parametrized by knot heights. Unlike \textbf{SAHP} and \textbf{THP}, this yields a closed-form cumulative intensity and exact sampling by inversion, while remaining arbitrarily flexible given enough knots. A similar but piecewise-constant parametrization is used in MOTOR~\citep{MOTOR}.

\item \textbf{IntensityFree}~\citep{IntensityFree}: an intensity-free model that parametrizes the inter-event-time distribution directly as a log-normal mixture, giving a closed-form likelihood with no numerical integration. In our default implementation it uses a Transformer encoder like the intensity models (the GRU encoder of the original model is compared in Appendix~\ref{app:full_results}).
\end{itemize}

Overall, we cover attention- and RNN-based intensity and intensity-free models with heads of increasing flexibility (NeuralConstTPP, NHP, THP, SAHP, IntensityFree, NeuralSplineTPP). Our goal is not an exhaustive benchmark of neural TPP architectures, but a targeted test of whether the flexibility of the intensity parametrization explains the previously reported gap to Hawkes processes. We therefore vary this flexibility while keeping the encoder fixed wherever possible. Further neural model families can be readily added to \textsc{TorchPP}, and we leave their evaluation to future work.

\subsection{Datasets and Evaluation Tasks}
\label{sec:datasets-experiments}
\label{sec:datasets-experiments/datasets}

\begin{table}[t]
  \caption{Datasets overview. Small- to medium-sized, low-dimensional standard benchmarks are shown in the top rows, and the medical evaluation tasks, with orders of magnitude more instances and event types, in the bottom rows. ``\(D\)'': number of event types. ``Term.'': whether each type occurs at most once. ``Mean Len'': mean number of events per sequence. Sequences are ICU-stay-level (with each patient only in one split) for MIMIC-IV and patient-level for UKB. }
  \label{tab:dataset_stats}
  \centering
  \small
  \begin{tabular}{lrrcrrr}
    \toprule
    Dataset & \(D\) & Term. & Train / Val / Test & Total Events & Mean Len\\
    \midrule
    Taxi          & 10    & No  & 1,400 / 200 / 400       & 74K   & 37.0\\
    StackOverflow & 22    & No  & 1,401 / 401 / 401       & 143K  & 64.6\\
    Retweet       & 3     & No  & 9,000 / 1,535 / 1,520   & 494K  & 41.1\\
    Amazon        & 16    & No  & 6,454 / 922 / 1,851     & 413K  & 44.7\\
    Taobao        & 17    & No  & 1,300 / 200 / 500        & 113K  & 56.7 \\
    \midrule
    UKB            & 1{,}257 & Yes & 351,656 / 100,473 / 50,238              & 6.2M          & 12.4 \\
    MIMIC-IV (ICU, trunc.) & 117     & No  & 67,320 / 8,474 / 8,531       & 7.96M & 94.4 \\
    \bottomrule
  \end{tabular}
\end{table}

Following the EasyTPP benchmark~\citep{xue2024easytpp}, we evaluate all the considered models on the Taxi~\citep{whong2014taxi}, Retweet~\citep{zhou2013triggering}, StackOverflow~\citep{leskovec2014snap}, Taobao~\citep{xue2022hypro}, and Amazon~\citep{ni2019amazon} datasets (Table~\ref{tab:dataset_stats}).%

\paragraph{Disease event sequences from the UK Biobank.}
Following~\citet{Delphi}, we consider longitudinal disease histories for more than 500,000 UK Biobank participants~\citep{sudlow2015ukb}: first-occurrence dates for 1{,}256 ICD-10-derived diagnosis categories and death, recorded in days since birth (mean sequence duration \(\approx\)~62~years). Details on data preparation and access are provided in Appendix~\ref{app:data_prep_ukb}.

\paragraph{Intensive-Care-Unit (ICU) event series data from the MIMIC-IV dataset.}
We evaluate our models on a task derived from the MIMIC-IV ICU dataset~\citep{mimic-iv}, containing event sequences (procedures, medications, and similar) from over 58{,}000 patients. We use the input and procedure events of each ICU stay, keep the 116 most frequent item types plus an \texttt{OTHER} class (\(D = 117\)), and right-truncate sequences to 316 events, the 90th percentile of sequence lengths. Event times are recorded to the minute, and we dequantize the events of each recorded minute uniformly within that minute. For more details see Appendix~\ref{app:data_prep_mimic}.

\subsection{Training and Evaluation Details}
\label{sec:datasets-experiments/setup} 
Model hyperparameters for each model--dataset combination are selected on the validation-set log-likelihood, employing a multi-stage search (Appendix~\ref{app:hyperparams}). The tuning budget per model family is compared in Table~\ref{tab:tuning_budget}.
Each model is trained with five random seeds (training configuration in Appendix~\ref{app:hyperparams_final}), and we report the mean and standard deviation over seeds.
All models are trained and selected on the full-sequence log-likelihood (Equation~\ref{eq:loglik}). For comparability with EasyTPP~\citep{xue2024easytpp}, test metrics are reported per event and conditioned on the first event of each sequence, excluding the survival term after the last event (Appendix~\ref{app:implementation_details}). We report:
\begin{itemize}
    \item the \textbf{log-likelihood} per event, decomposed as \(\log p(t, d \mid \mathcal{H}) = \log p(t \mid \mathcal{H}) + \log p(d \mid t, \mathcal{H})\) into a \textbf{time term} and a \textbf{type-conditional term};
    \item the \textbf{time-conditional accuracy} \(\mathrm{Acc}|t\): the top-1 accuracy of \(p(d \mid t, \mathcal{H})\) at the true next-event time \(t\);
    \item the \textbf{time RMSE} of the next-event time prediction \(\mathbb{E}[t \mid \mathcal{H}]\).
\end{itemize}
Additional metrics are reported in Appendix~\ref{app:full_results}: the \textbf{marginal type log-likelihood} \(\log p(d \mid \mathcal{H}) = \log \int p(t, d \mid \mathcal{H})\,dt\) and the corresponding marginal top-1 \textbf{accuracy}.

The quantities \(p(d \mid \mathcal{H})\) and \(\mathbb{E}[t \mid \mathcal{H}]\) require numerical integration over the next-event time. Because this is expensive, all test metrics are computed on the first 1{,}000 test sequences of each dataset (all test sequences for Taxi, StackOverflow, and Taobao) and, for the integrated metrics on MIMIC-IV, on the first 500.
See Appendix~\ref{app:math_details} for full details.

\section{Results}

\subsection{Revisiting Existing Benchmark Datasets}
\label{sec:benchmarks}
\label{sec:results}

Table~\ref{tab:main_results} reports the per-event test log-likelihood on the five standard benchmark datasets, split into the time term \(\log p(t)\) and the type term \(\log p(d \mid t)\), which sum to the joint log-likelihood. All further metrics are presented in Appendix~\ref{app:full_results}.

\begin{table}[t]
  \caption{Main results: per-event test log-likelihood on the standard benchmark datasets (nats, higher is better; mean over 5 seeds; EasyTPP convention, i.e.\ conditioned on the first event and without the final survival term, Section~\ref{sec:datasets-experiments/setup}). Each cell is \(\log p(t)\,/\,\log p(d\mid t)\) (time\,/\,type), and the joint \(\log p(t,d)=\log p(t)+\log p(d\mid t)\) is their sum (Table~\ref{tab:joint_ll_results}; \(\pm\)std in Appendix~\ref{app:full_results}). Models differ chiefly in the time term \(\log p(t)\), the axis that flexibility improves, while the type term \(\log p(d\mid t)\) is comparatively tight across history-aware models (near-flat on StackOverflow and Retweet). Row labels: baseline/kernel (C: constant, S: spline, E: exponential), I.: inhibitive, N.: neural. \textbf{Bold}: best per dataset and component; \underline{underline}: best joint log-likelihood; ties at the displayed precision are all marked.}
  \label{tab:main_results}
  \centering
  \small
  \setlength{\tabcolsep}{3pt}
  \begin{tabular}{lccccc}
    \toprule
    Model & Taxi & StackOv. & Retweet & Amazon & Taobao \\
    \midrule
    \multicolumn{6}{l}{\textit{Poisson processes}} \\
    H. Poisson & \(0.51/{-}1.14\) & \({-}0.86/{-}1.90\) & \({-}3.58/{-}0.82\) & \({-}0.36/{-}2.24\) & \(2.07/{-}2.18\) \\
    S. Poisson & \(0.51/{-}1.13\) & \({-}0.85/{-}1.88\) & \({-}3.43/{-}0.82\) & \({-}0.35/{-}2.24\) & \(2.11/{-}2.19\) \\
    \midrule
    \multicolumn{6}{l}{\textit{Hawkes processes}} \\
    C/E Hawkes & \(0.45/{-}0.74\) & \({-}0.76/{-}1.62\) & \({-}3.42/{-}0.79\) & \({-}0.10/{-}2.03\) & \(2.61/{-}1.42\) \\
    S/E Hawkes & \(0.45/{-}0.74\) & \({-}0.75/{-}1.60\) & \({-}3.38/\mathbf{{-}0.78}\) & \({-}0.11/{-}2.04\) & \(2.61/{-}1.43\) \\
    C/S Hawkes & \(0.50/{-}0.71\) & \({-}0.73/{-}1.60\) & \({-}3.40/{-}0.79\) & \(0.01/{-}1.96\) & \(2.65/\mathbf{{-}1.38}\) \\
    S/S Hawkes & \(0.49/{-}0.70\) & \underline{\({-}0.73/{-}1.59\)} & \({-}2.37/{-}0.80\) & \(1.73/{-}2.04\) & \(2.67/{-}1.41\) \\
    \midrule
    \multicolumn{6}{l}{\textit{Inhibitive Hawkes processes}} \\
    I. C/E Hawkes & \(0.54/{-}0.33\) & \({-}0.75/{-}1.60\) & \({-}3.41/\mathbf{{-}0.78}\) & \(0.05/{-}2.06\) & \(2.62/{-}1.43\) \\
    I. S/S Hawkes & \(0.68/{-}0.28\) & \({-}0.76/{-}1.60\) & \underline{\(\mathbf{{-}1.29}/{-}0.80\)} & \(2.05/{-}2.17\) & \(2.66/{-}1.47\) \\
    \midrule
    \multicolumn{6}{l}{\textit{Neural TPPs}} \\
    N. Const & \(0.55/{-}0.29\) & \({-}0.80/{-}1.59\) & \({-}3.42/\mathbf{{-}0.78}\) & \({-}0.34/{-}1.99\) & \(2.07/{-}1.43\) \\
    THP & \(0.56/{-}0.26\) & \({-}0.80/{-}1.59\) & \({-}3.41/\mathbf{{-}0.78}\) & \({-}0.31/{-}1.99\) & \(2.11/{-}1.45\) \\
    SAHP & \(0.64/{-}0.24\) & \underline{\({-}0.75/\mathbf{{-}1.57}\)} & \({-}3.38/{-}0.79\) & \(0.14/{-}2.20\) & \(2.66/{-}1.49\) \\
    NHP & \(0.60/{-}0.24\) & \({-}0.79/{-}1.60\) & \({-}3.51/{-}0.81\) & \({-}0.32/{-}2.06\) & \(2.56/{-}1.51\) \\
    \midrule
    \multicolumn{6}{l}{\textit{Flexible neural TPPs}} \\
    IntensityFree & \(\mathbf{0.73}/{-}0.29\) & \({-}0.73/{-}1.60\) & \({-}3.66/{-}0.79\) & \(2.41/{-}2.04\) & \underline{\(\mathbf{2.73}/{-}1.42\)} \\
    N. Spline & \underline{\(0.71/\mathbf{{-}0.22}\)} & \underline{\(\mathbf{{-}0.72}/{-}1.60\)} & \({-}1.86/\mathbf{{-}0.78}\) & \underline{\(\mathbf{2.56}/\mathbf{{-}1.91}\)} & \(2.71/{-}1.43\) \\
    \bottomrule
  \end{tabular}
\end{table}

\paragraph{Hawkes processes close the gap.}
Once Hawkes processes are equipped with flexible baselines, flexible kernels, and inhibitive effects, the previously reported performance gap to neural TPPs narrows substantially (Table~\ref{tab:joint_ll_results}). On every dataset, the best Hawkes variant matches or outperforms the softplus-headed neural TPPs (\textbf{THP}, \textbf{SAHP}, \textbf{NHP}). Against the flexible neural TPPs, Hawkes models are competitive on four of the five datasets: \textbf{Inhib.\ Spline/Spline Hawkes} is the best model overall on Retweet, \textbf{Spline/Spline Hawkes} ties with \textbf{NeuralSplineTPP} on StackOverflow and Taobao, and the remaining gap on Taxi is small (\(0.09\) nats). Amazon is the only dataset with a substantial gap (\(0.77\) nats). The performance gap reported in prior benchmarks on these datasets therefore largely reflects under-specified Hawkes models rather than an intrinsic advantage of neural architectures. The relevance of individual model components varies across datasets. A non-stationary baseline intensity (\textbf{Spline Poisson} vs.\ \textbf{homogeneous Poisson}) provided the biggest benefits on Retweet. Inhibition helped on Taxi, Retweet, and Amazon, but offered no gain over the best Hawkes model on StackOverflow, and reduced performance on Taobao. We attribute these losses to optimization difficulties: numerical integration of the cumulative intensity likely leads to noisier gradients and less stable training. Replacing the exponential kernel with a spline kernel improved the spline-baseline Hawkes process on all datasets, to varying degrees.

\begin{figure}[t]
  \centering
  \includegraphics[width=\textwidth]{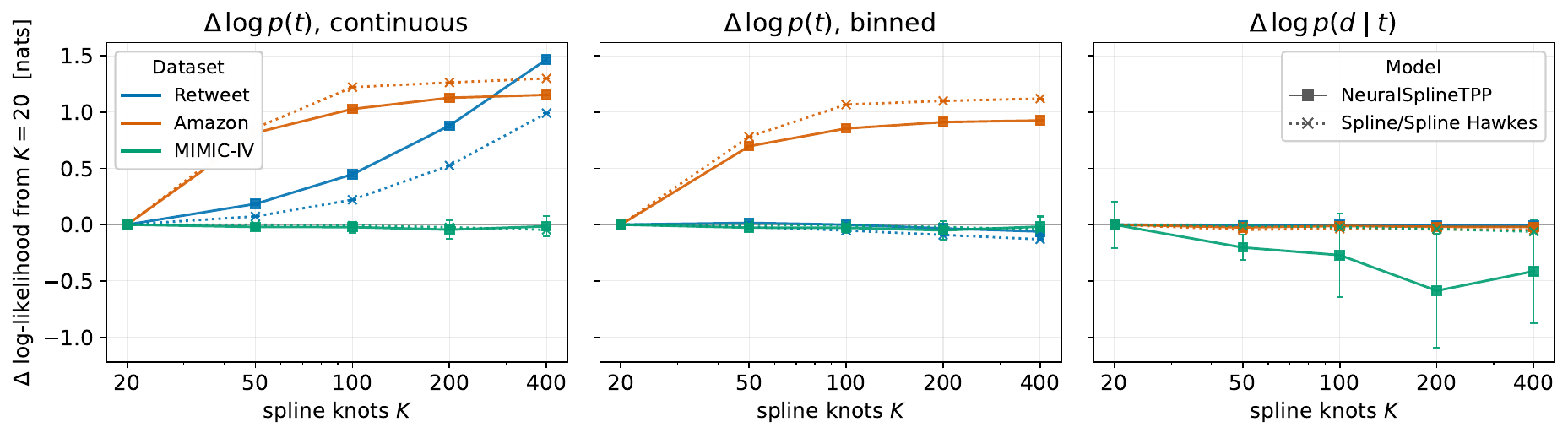}
  \caption{Change in per-event test \(\log p(t)\) and \(\log p(d\mid t)\) (nats) vs.\ spline knot count \(K\) (relative to \(K{=}20\)), for \textbf{NeuralSplineTPP} (intensity-head knots) and \textbf{Spline/Spline Hawkes} (kernel knots), on three datasets (all datasets in Tables~\ref{tab:ksweep_all} and~\ref{tab:binned_ksweep}). \emph{Left:} the continuous time term \(\log p(t)\) increases with \(K\) on Retweet and Amazon. \emph{Middle:} scored by the probability of the recorded time bin (Retweet: 1, Amazon: 0.01, MIMIC-IV: 1 min), it increases only on Amazon. \emph{Right:} the type term \(\log p(d\mid t)\) does not improve with \(K\).}
  \label{fig:flex_decomp}
\end{figure}

\paragraph{Flexible intensity parametrization matters.}
Flexible intensity parametrizations benefit neural TPPs as well. \textbf{NeuralSplineTPP} outperforms \textbf{THP}, \textbf{SAHP}, \textbf{NeuralConstTPP}, and \textbf{NHP} by a considerable margin on all datasets except StackOverflow, where it ties with \textbf{SAHP}. As the Transformer models differ mainly in their output head, this is consistent with the flexible but analytically tractable spline head being the decisive factor. The \textbf{IntensityFree} model, which likewise fits a flexible next-event-time density, performs best on Taobao and second best on Taxi and Amazon, but falls behind on Retweet.
Notably, the constant-intensity \textbf{NeuralConstTPP} performs comparably to \textbf{THP}, while requiring no numerical integration. Equipped with the GRU encoder of the original model, \textbf{IntensityFree} performs similarly or worse on Taxi, StackOverflow, Retweet, and Amazon, but better on Taobao, MIMIC-IV, and UKB (Table~\ref{tab:iftpp_encoder}). On Retweet, the differences in \(\log p(t)\) mostly reflect how strongly a model exploits the discretized time grid. A continuous density can place narrow peaks on such a grid, and its log-likelihood then grows without bound. Scored by the probability of the recorded time bin instead (Appendix~\ref{app:binned}), the gain from more knots vanishes on Retweet but persists on Amazon, whose time stamps lie on no grid (Figure~\ref{fig:flex_decomp}).

\paragraph{The flexibility gain is a sharper time density.}
Decomposing each model's log-likelihood into a time term \(\log p(t)\) and a time-conditional type term \(\log p(d \mid t)\) reveals where the gains from flexibility arise. We vary the number of spline knots \(K\) from 10 to 400, both in the intensity head of \textbf{NeuralSplineTPP} and in the kernels of \textbf{Spline/Spline Hawkes} (Figure~\ref{fig:flex_decomp}, Table~\ref{tab:ksweep_all}). For both models, \(\log p(t)\) increases by 0.9--1.6 nats on Retweet and Amazon and remains essentially unchanged on the other datasets, whereas \(\log p(d \mid t)\) does not improve on any dataset. On Retweet, Amazon, and Taobao, the spread between history-dependent models is several times larger in \(\log p(t)\) than in \(\log p(d \mid t)\) (Table~\ref{tab:main_results}). The exception is Taxi, whose event types strictly alternate between pick-up and drop-off events. Here, the type-conditional term separates the models that can represent inhibition (inhibitive Hawkes and neural TPPs) from the excitatory Hawkes processes. On these benchmarks, the flexibility that separates modern neural TPPs from classical Hawkes processes thus mostly sharpens \emph{when} the next event occurs, not \emph{which} type it is.

\paragraph{Benchmark datasets are close to saturation.}
A flexible baseline alone (\textbf{Spline Poisson}) performs considerably worse, whereas adding pairwise interactions (\textbf{(Inhib.) Spline/Spline Hawkes}) closes most of the gap to the neural TPPs. A goodness-of-fit test based on the time-rescaling theorem~\citep{brown2002timerescaling} (Table~\ref{tab:calibration}) shows that the flexible models are the best calibrated in time on four of the five datasets. As such, current benchmark datasets offer neural TPPs little opportunity to model interactions beyond pairwise ones.

\subsection{Large-Scale Medical Evaluation Tasks}
\label{sec:medical}

Table~\ref{tab:medical_results} shows the results on the two medical tasks (full metrics in Tables~\ref{tab:full_mimic_results} and~\ref{tab:ukb_full_results}).

\paragraph{Disease histories are well explained by additive interactions.}
Recent work has shown that transformer models capture population-scale disease trajectories across more than a thousand disease types~\citep{Delphi}. Our UKB task is a simplified variant of this setting, excluding the sex and lifestyle covariates, evaluated on next-event likelihood and accuracy. On this task, flexible Hawkes processes with purely excitatory, additive interactions come close to the best neural model: \textbf{Spline/Spline Hawkes} trails \textbf{NeuralSplineTPP} by only \(0.07\) nats in joint log-likelihood and \(0.08\) percentage points in \(\mathrm{Acc}|t\) (Table~\ref{tab:ukb_full_results}), and outperforms other neural TPPs. This suggests that pairwise interactions capture most of the next-event structure in first-occurrence disease histories, making flexible Hawkes processes a strong and interpretable model choice. It does not speak to other capabilities of large generative models of health trajectories, such as long-horizon generation, or multi-modal extensions.
We did not evaluate inhibitive Hawkes processes on UKB. However, the flexible neural models can represent inhibitive and non-additive effects, and they do not improve substantially on the excitatory Hawkes process. %

\begin{wraptable}{r}{0.50\textwidth}
  \caption{Large-scale medical results (nats; cell = \(\log p(t)/\log p(d\mid t)\); row labels, bold and underline as in Table~\ref{tab:main_results}). On UKB, inhibitive Hawkes were not evaluated and NHP did not converge. Joint, \(\pm\)std and full metrics in Appendix~\ref{app:full_results}.}
  \label{tab:medical_results}
  \centering
  \footnotesize
  \setlength{\tabcolsep}{3pt}
  \begin{tabular}{lcc}
    \toprule
    Model & UKB & MIMIC-IV \\
    \midrule
    H. Poisson & \({-}8.08/{-}5.59\) & \(2.60/{-}3.78\) \\
    S. Poisson & \({-}7.51/{-}5.36\) & \(2.64/{-}3.71\) \\
    \midrule
    C/E Hawkes & \({-}7.36/{-}5.15\) & \(3.38/{-}2.71\) \\
    S/E Hawkes & \({-}7.26/{-}5.07\) & \(3.39/{-}2.70\) \\
    C/S Hawkes & \({-}7.30/{-}5.12\) & \(3.78/{-}2.62\) \\
    S/S Hawkes & \({-}7.23/{-}5.05\) & \(3.83/{-}2.61\) \\
    \midrule
    I. C/E Hawkes & -- & \(3.87/{-}2.67\) \\
    I. S/S Hawkes & -- & \(3.83/{-}2.68\) \\
    \midrule
    N. Const & \({-}7.27/{-}5.09\) & \(3.08/{-}2.65\) \\
    THP & \({-}7.39/{-}5.41\) & \(3.26/\mathbf{{-}2.36}\) \\
    SAHP & \({-}7.64/{-}5.76\) & \(3.66/{-}2.74\) \\
    NHP & -- & \(3.60/{-}2.98\) \\
    \midrule
    IntensityFree & \({-}7.42/{-}5.44\) & \(\mathbf{4.06}/{-}2.61\) \\
    N. Spline & \underline{\(\mathbf{{-}7.19}/\mathbf{{-}5.02}\)} & \underline{\(4.05/{-}2.39\)} \\
    \bottomrule
  \end{tabular}
\end{wraptable}

\paragraph{MIMIC-IV: where do the gains come from?}
On MIMIC-IV, NeuralSplineTPP leads the best Hawkes variant (Spline/Spline Hawkes) by \(0.45\)~nats in joint log-likelihood, with \(0.22\)~nats in the time term and \(0.23\)~nats in the type term, and has a higher time-conditional accuracy (\(38.3\%\) vs.\ \(33.5\%\) for the most type-accurate Hawkes variant, Inhib.\ Spline/Spline Hawkes, Table~\ref{tab:full_mimic_results}). \textbf{THP} and \textbf{NeuralSplineTPP} reach the best type terms (\(-2.36\) and \(-2.39\)). \textbf{IntensityFree}, whose type distribution is independent of event time, performs at the level of the Hawkes processes (\(-2.61\), and \(-2.53\) with a GRU encoder, Table~\ref{tab:iftpp_encoder}). The advantage of neural TPPs is concentrated on events recorded within the same minute as the preceding event, which make up \(41.6\%\) of all events. On these events, most neural TPPs predict the event type more accurately than the Hawkes processes. On other events, NeuralSplineTPP and Spline/Spline Hawkes differ by only \(0.02\)~nats in \(\log p(d\mid t)\), and Spline/Spline Hawkes has the higher \(\log p(t)\) (Appendix~\ref{app:mimic_controls}). %
NeuralSplineTPP places its knots in the time elapsed since the last event, whereas the Hawkes spline kernels are anchored at each trigger event, which may favor the neural model on very short-range structure.

\section{Discussion}
\label{sec:discussion}

Our results suggest that prior benchmarks evaluated Hawkes processes in an overly restricted form. With flexible baseline intensities, learned kernels, and inhibition, Hawkes processes match or come close to the performance of neural TPPs on four of the five standard datasets. This indicates that these benchmarks are dominated by pairwise interactions. Amazon is the only standard benchmark on which the flexible neural models retain a clear advantage, mainly in the time term. On MIMIC-IV, the flexible neural TPPs retain an advantage in both terms, concentrated on events recorded in the same minute.

Neural TPPs such as THP~\citep{THP}, SAHP~\citep{SAHP}, NHP~\citep{nhp2017}, RMTPP~\citep{RMTPP}, and AttNHP~\citep{AttNHP} differ in their encoders but, like classical Hawkes processes, lack flexibility in the intensity parametrization. Splines offer a practical sweet spot for both families, as they are flexible yet allow exact sampling without thinning.

\paragraph{Implications for practitioners.} Our experiments indicate that most of the datasets we consider are well described by pairwise interactions, and we expect this observation to extend to other TPP datasets~\citep{TPPSurvey}. Consequently, a flexible, well-tuned Hawkes process is a sensible starting point; the additional cost and complexity of a neural TPP are justified only for data with structure beyond pairwise interactions.

\paragraph{Implications for benchmarking and future developments.} To assess neural TPPs, future benchmarks should use datasets with long-range dependencies and structure beyond pairwise interactions, such as electronic health records~\citep{EHRSHOT, mimic-iv}, including rich event attributes beyond categorical types, such as lab measurements, genomic data, images, and clinical reports. Handling datasets requires care in how event times are recorded. On datasets with discretized timestamps, such as Retweet, the continuous-time likelihood is unbounded. Such data therefore require either dequantization or a discrete-time model. A related problem arises with simultaneous events, to which a continuous-time TPP assigns zero probability. Handling them requires discretized time, set-valued marks that represent all events at a time point jointly, or an explicit tie-breaking rule. More complex neural TPP methods~\citep{S2P2, EventFlow}, especially those requiring substantial tuning or compute, should also be compared against simple but flexible alternatives like \textbf{NeuralSplineTPP}.

\paragraph{Limitations.}
(i) \emph{Scope.} Our conclusions apply to Poisson and Hawkes processes, Transformer-based intensity models, NHP, and IntensityFree. They do not necessarily extend to Cox-type models or to flow-, diffusion-, or state-space-based TPPs, which we did not evaluate.
(ii) \emph{Task.} We evaluate autoregressive next-event prediction only. We do not assess long-horizon simulation or use of the models for planning and decision making, and calibration is checked only for four models with a closed-form cumulative intensity (Table~\ref{tab:calibration}).
(iii) \emph{Datasets.} Our conclusions rest on seven datasets, and other domains may favor neural models more prominently. Our aim is a fair comparison of TPP families, not state-of-the-art forecasting models for electronic health records. Both medical tasks are based on access-controlled data, and due to privacy regulations we can neither release any data derived from them nor the trained models. We use MIMIC-IV as a testbed for the modeling capabilities of TPPs, not to build a clinically useful forecasting model, and therefore retain only a subset of the recorded event types (Section~\ref{sec:datasets-experiments/datasets}). For use in continuous-time TPPs, we dequantize simultaneous events. On UKB, the sex and lifestyle covariates of the original pipeline~\citep{Delphi} are not included for simplicity.

\section{Conclusion}
\label{sec:conclusion}

We have shown that the reported gap between classical Hawkes processes and neural TPPs is largely an artifact of under-specified models. Most of our datasets show little evidence of structure beyond pairwise interactions, highlighting the need for benchmarks that contain such structure. With flexible baseline intensities, flexible kernels, and inhibition, Hawkes processes match or come close to the performance of neural TPPs on four of the five standard benchmarks and remain competitive on UK Biobank disease histories. Neural TPPs are likewise limited by the flexibility of their intensity parametrization. We release \textsc{TorchPP}, a modular PyTorch framework with unified implementations of classical and neural TPPs, including support for terminating and right-censored sequences.

\begin{ack}
This work was supported by the European Research Council (ERC) under the European Union's Horizon Europe research and innovation programme (project TAIPO, grant agreement No.~101088594). Funded by the European Union. Views and opinions expressed are however those of the author(s) only and do not necessarily reflect those of the European Union or the European Research Council Executive Agency. Neither the European Union nor the granting authority can be held responsible for them. This work was also funded by the Priority Program Translational Oncology of the Deutsche Krebshilfe (grant 70115167). Computations were performed on the GPU cluster of the DKFZ and on the UK Biobank Research Analysis Platform.
This research has been conducted using the UK Biobank Resource under Application Number 81358. MIMIC-IV was accessed through PhysioNet under its credentialed-access data use agreement.
\end{ack}

{
\small
\bibliographystyle{plainnat}

}

\appendix

\section{Implementation Overview}
\label{sec:framework}

\textsc{TorchPP} is organized around a single \texttt{TemporalPointProcess} abstraction that exposes intensity evaluation, cumulative-intensity computation, likelihood evaluation, density computation, distribution queries, and event sampling. Concrete models only need to provide the conditional intensity \(\lambda_d^*(t \mid \mathcal{H}_t)\) and, when available, a closed-form cumulative intensity. This keeps the training and evaluation code shared across Poisson, Hawkes, and neural families.

\subsection{Core Abstractions}

The base class separates three concerns. First, model-specific code computes the per-type intensity at an arbitrary query time. Second, the library provides shared likelihood and sampling utilities on top of that interface. Third, cumulative intensities can either be supplied analytically or approximated numerically through a common integration path. In practice, this means that simple models such as Poisson or exponential-kernel Hawkes processes use exact cumulative intensity computations, whereas inhibitive and fully neural models can still be trained and sampled through the same API with numerical quadrature.

The Hawkes family is implemented compositionally. A baseline module \(\mu_d(t)\) and a kernel module \(\phi_{d' \to d}(\tau)\) are instantiated independently and combined as
\begin{equation}
  \lambda_d^*(t) = \mu_d(t) + \sum_{j : t_j < t} \phi_{d_j \to d}(t - t_j),
\end{equation}
or, for inhibitive variants, after an outer positivity map \(g\),
\begin{equation}
  \lambda_d^*(t) = g\!\left(\mu_d(t) + \sum_{j : t_j < t} \phi_{d_j \to d}(t - t_j)\right).
\end{equation}
This makes constant, linear, and spline baselines interchangeable with exponential, spline, and neural kernels without duplicating likelihood code.

Neural-network-based models use the same pattern at a different abstraction level. The event history is encoded once into a context representation \(h_\ell\) per event at time \(t_\ell\), and a lightweight head maps \((h_\ell, t - t_\ell)\) to model parameters or directly to the intensity. Analytical heads such as the constant and spline heads provide closed-form cumulative intensities; fully neural heads fall back to the numerical path. Methods that integrate over the intensity function use a per-batch cache of the encoder output. Integration then requires only repeated evaluations of the lightweight head and a single forward pass through the encoder, which is substantially cheaper.

\subsection{Support for Termination and Right-Censoring}

Two implementation details are central for the medical data. First, terminating processes, as in the UK Biobank task, are handled explicitly: once an event type has occurred for a patient, the corresponding intensity is masked out and its cumulative intensity is truncated at the event time. This allows the same likelihood code to handle repeated-event and first-occurrence settings.

Second, right-censoring is supported. If a sequence is observed up to a sequence-specific censoring time \(C \geq t_N\), the survival term is integrated up to \(C\). We use this for MIMIC-IV, where the observation window ends at ICU discharge (for stays that are not truncated). For UKB, we set \(C = t_N\). Previous neural TPP libraries assume that the censoring time coincides with the last event.

\subsection{Sampling, Integration, and Reproducibility}

TorchPP uses cumulative-intensity inversion as its default sampler, which is exact whenever the cumulative intensity is available in closed form. Given \(u \sim \mathrm{Exp}(1)\), the next event time is obtained by solving \(\Lambda^*(t_\ell, t) = u\), followed by categorical sampling over event types using the per-type intensities at the sampled time. Because the same cumulative-intensity code is also used in training, analytical and numerical models share the same execution path and the same validation checks. Ogata thinning remains available for comparison. For numerical integration, trapezoidal, Monte Carlo, and Monte Carlo trapezoidal rules are available.

At the systems level, all experiments are configured by composable YAML fragments with dotted command-line overrides. Training uses bucketed batches and dynamic batch sizing to reduce padding overhead and keep the memory footprint constant. The same model code is therefore used in the analytical derivations, the controlled benchmark runs, and the UKB experiments discussed in the main text.

\subsection{Comparison to Previous Frameworks}
\label{app:implementation_details}

EasyTPP is the closest public comparison point for the benchmark experiments, so we use it to highlight the design choices that materially affect evaluation.

\paragraph{Likelihood definition.}
EasyTPP evaluates an autoregressive likelihood that conditions on the first event,
\begin{equation}
    \log p\big(\{(t_i, d_i)\}_{i=2}^{N} \mid (t_1, d_1)\big) = \sum_{i=2}^{N} \log p(t_i, d_i \mid \mathcal{H}_{t_i}),
    \label{eq:easytpp_ll}
\end{equation}
whereas TorchPP optimizes the full sequence likelihood
\begin{equation}
    \log p(\mathcal{H}_T) = \sum_{i=1}^{N} \log \lambda_{d_i}^*(t_i) - \int_0^T \lambda^*(s)\,ds,
\end{equation}
which additionally contains the first-event term \(\log p(t_1, d_1)\) (including the survival on \([0, t_1]\)) and the survival term \(-\int_{t_N}^{T} \lambda^*(s)\,ds\) after the last observed event. All TorchPP models are trained and selected on the full likelihood. All reported test log-likelihoods and accuracies follow Equation~\eqref{eq:easytpp_ll}, averaged per predicted event (\(i = 2, \dots, N\)), for comparability with EasyTPP.

\paragraph{Support for terminating and censored sequences.}
EasyTPP assumes recurrent events on a shared observation window. TorchPP instead supports type-specific termination and per-sequence censoring times. In the terminating setting, each type-specific cumulative intensity is truncated at its first occurrence time; in the censored setting, the survival term is integrated up to the sequence-specific observation end.

\paragraph{Cumulative intensity evaluation.}
TorchPP switches between analytical and configurable numerical cumulative intensities (Monte Carlo, trapezoidal, and Monte Carlo trapezoidal) depending on the model. For neural models, encoder outputs are cached per batch and only the lightweight head is re-evaluated at quadrature points, which substantially lowers the cost of accurate numerical integration.

\subsubsection{Improvements over the EasyTPP Benchmark Results}
\label{app:easytpp_issues}

\citet{xue2024easytpp} used a fixed thinning over-sampling rate and a fixed search window for the proposal bound over all datasets (according to their repository). When the bound is too low, predicted event times are biased and RMSE is inflated, which we observed when we reran the tests with new configurations and obtained lower RMSE values than previously reported (see Appendix~\ref{app:equivalence}). Our experiments also used a deterministic per-interval quadrature (Appendix~\ref{app:hyperparams_final}) and a longer training schedule with warmup and cosine learning-rate decay.

\subsection{Validation of Implementation Correctness}
\label{app:equivalence}

To check the correctness of our implementations, we re-implemented the THP and SAHP architectures in TorchPP, matching their encoder dimensions, attention structure, activation pattern, and Softplus-wrapped intensity heads. Evaluation on the EasyTPP benchmark datasets additionally reports the same conditional likelihood convention used by EasyTPP, while TorchPP training remains on the full likelihood.

Table~\ref{tab:equivalence} shows that THP agrees closely across frameworks on accuracy, RMSE, and log-likelihood. SAHP matches in accuracy but exhibits larger likelihood discrepancies, especially on StackOverflow, which is consistent with the training-instability behavior we observed in the EasyTPP implementation.  In these runs, we integrate with 100 points during training, which stabilizes SAHP. The lower TorchPP RMSE values are expected once biased thinning is removed.

\begin{table}[ht]
  \caption{Code equivalence validation: TorchPP re-implementations vs.\ EasyTPP framework on four benchmark datasets, with reference numbers from \citep{xue2024easytpp}.
  LL: joint log-likelihood (nats/event), \(\mathrm{Acc}|t\): accuracy given the true next event time, as used by EasyTPP (as a fraction), RMSE: next-event time RMSE.
  TorchPP values are means over 5 seeds. EasyTPP (rerun) values are single-framework runs we performed locally. EasyTPP (paper) values are taken from the original EasyTPP benchmark; the reported type error rates are converted to accuracies. The TorchPP values use the EasyTPP-matched training configuration of Appendix~\ref{app:hyperparams_easytpp} (20{,}000 steps, 100 integration points) and therefore differ from Appendix~\ref{app:full_results}. EasyTPP's reported Hawkes baseline is analyzed separately in Appendix~\ref{app:tick}.}
  \label{tab:equivalence}
  \centering
  \small
  \begin{tabular}{ll rrr rrr rr}
    \toprule
    & & \multicolumn{3}{c}{TorchPP} & \multicolumn{3}{c}{EasyTPP (rerun)} & \multicolumn{2}{c}{EasyTPP (paper)} \\
    \cmidrule(lr){3-5} \cmidrule(lr){6-8} \cmidrule(lr){9-10}
    Dataset & Model & LL & Acc\(|t\) & RMSE & LL & Acc\(|t\) & RMSE & Acc\(|t\) & RMSE \\
    \midrule
    \multirow{2}{*}{Taxi}
    & SAHP                 & 0.25    & 0.918 & 0.290 & \(-0.39\) & 0.897 & 0.313 & 0.903 & 0.372 \\
    & THP                  & 0.25    & 0.913 & 0.285 & 0.24    & 0.909 & 0.305 & 0.913 & 0.370 \\
    \midrule
    \multirow{2}{*}{StackOverflow}
    & SAHP                 & \(-2.36\) & 0.462 & 1.040 & \(-8.23\) & 0.425 & 1.307 & 0.439 & 1.375 \\
    & THP                  & \(-2.40\) & 0.452 & 1.029 & \(-2.54\) & 0.443 & 1.130 & 0.450 & 1.374 \\
    \midrule
    \multirow{2}{*}{Retweet}
    & SAHP                 & \(-4.14\) & 0.596 & 15.95 & \(-5.14\) & 0.540 & 19.69 & 0.584 & 22.40 \\
    & THP                  & \(-4.17\) & 0.596 & 16.02 & \(-4.16\) & 0.597 & 17.57 & 0.585 & 22.01 \\
    \midrule
    \multirow{2}{*}{Amazon}
    & SAHP                 & \(-2.15\) & 0.356 & 0.334 & \(-2.53\) & 0.339 & 0.397 & 0.323 & 0.619 \\
    & THP                  & \(-2.31\) & 0.357 & 0.331 & \(-2.22\) & 0.349 & 0.374 & 0.339 & 0.621 \\
    \bottomrule
  \end{tabular}
\end{table}

The close agreement suggests that the TorchPP re-implementations are faithful. Where metrics diverge, the remaining gap is explained by identified framework differences and by the sampling and integration issues in EasyTPP (see the difference between our EasyTPP re-runs and the paper values), rather than by a mismatch in model equations.

\subsection{Reproducing EasyTPP's Hawkes Baseline}
\label{app:tick}

EasyTPP~\citep{xue2024easytpp} reports a multivariate Hawkes (MHP) baseline whose next-type accuracy is far below ours on four of the five datasets and above our non-inhibitive Hawkes processes on Taxi (\(90.5\%\)). According to their Appendix~B, this baseline was fitted externally with the \texttt{tick} library~\citep{bacry2017tick} using an exponential kernel with fixed parameters: ``the base intensity is set to 0.2 for all dimensions while the decay matrix is set to be a unit matrix plus a small noise''. The code that produced these numbers is not part of the EasyTPP repository, which contains neither a Hawkes model nor \texttt{tick}-fitting code, so the numbers cannot be re-derived from the released code. We therefore re-fitted the same model with \texttt{tick} (version 0.8.0.2) following this recipe on the native EasyTPP time units, and evaluated \(\mathrm{Acc}|t\) as \(\arg\max_d \lambda_d^*(t_i)\) at the true next-event time with a causal history, the convention of EasyTPP's Appendix~B.2 and of our own evaluation. As \texttt{tick}'s likelihood-based estimator fails in this release, we additionally fitted the identical model by positivity-constrained maximum likelihood.

Table~\ref{tab:tick_repro} shows the results. Our \texttt{tick} fits match or exceed EasyTPP's reported MHP accuracy on StackOverflow, Retweet, Amazon, and Taobao (the one exception is the fit with the base intensity fixed to 0.2 on Retweet, far from that dataset's natural rate), where the reported values lie 6--12 points \emph{below} the majority-class rate. No properly fitted model should fall below this value, since always predicting the most frequent type attains it. On Taxi, no faithful excitation-only fit exceeds \(\approx 85\%\), in line with our learned Const./Exp Hawkes. The \(90\%\) regime on Taxi is reached only by our inhibitive Hawkes processes and the neural models, since a Hawkes process with non-negative interactions cannot suppress a type that has just occurred, which the Taxi data appears to require. The same pipeline reproduces EasyTPP's \emph{neural} accuracies closely (Table~\ref{tab:equivalence}). We conclude that the discrepancy is specific to EasyTPP's MHP baseline and that we cannot compare to it in good faith. We therefore do not include its numbers in our main tables.

\begin{table}[t]
  \caption{Reproducing EasyTPP's Hawkes (MHP) baseline. Time-conditional next-type accuracy \(\mathrm{Acc}|t\) (\%). The \texttt{tick} fits, the majority rate and EasyTPP use the full test sets, our models the first 1{,}000 test sequences (Section~\ref{sec:datasets-experiments/setup}). \emph{Majority}: always predicting the most frequent type. \emph{tick LS}: \texttt{tick} \texttt{HawkesExpKern} with fixed decay 1.0 and its default least-squares fit; \emph{MLE}: the same exponential Hawkes fitted by positivity-constrained maximum likelihood; \emph{MLE, \(\mu{=}0.2\)}: additionally fixing the base intensity to 0.2, the literal recipe of EasyTPP's Appendix~B. Ours: our learned Const./Exp Hawkes and the best of our (inhibitive) Hawkes variants, from Table~\ref{tab:cond_acc_results}.}
  \label{tab:tick_repro}
  \centering
  \small
  \resizebox{\textwidth}{!}{%
  \begin{tabular}{lcccccc}
    \toprule
    & & \multicolumn{3}{c}{\texttt{tick} MHP (our fit)} & & \\
    \cmidrule(lr){3-5}
    Dataset & Majority & tick LS & MLE & MLE, \(\mu{=}0.2\) & EasyTPP (reported) & Ours: C/E \;/\; best Hawkes \\
    \midrule
    Taxi & 44.3 & 79.6 & 84.2 & 84.6 & 90.5 & 85.0 \;/\; 92.0 \\
    StackOverflow & 42.5 & 44.5 & 43.6 & 44.2 & 35.0 & 46.2 \;/\; 46.7 \\
    Retweet & 54.0 & 53.8 & 54.1 & 42.2 & 44.3 & 57.5 \;/\; 58.6 \\
    Amazon & 30.0 & 29.9 & 28.9 & 32.8 & 24.1 & 35.0 \;/\; 39.1 \\
    Taobao & 43.6 & 54.1 & 53.7 & 54.5 & 31.9 & 61.0 \;/\; 61.5 \\
    \bottomrule
  \end{tabular}}
\end{table}

\FloatBarrier
\newpage

\section{Mathematical Details on Temporal Point Processes}
\label{app:math_details}

\begin{figure}[t]
\centering
\resizebox{\textwidth}{!}{%
\begin{tikzpicture}[
  >={Stealth[length=2.2mm]},
  font=\small,
  intens/.style={rectangle, rounded corners=2pt, draw=black!55, fill=blue!8,
                 align=center, minimum width=2.6cm, minimum height=1.0cm, inner sep=2pt},
  trans/.style={rectangle, rounded corners=2pt, draw=black!55, fill=teal!10,
                align=center, minimum width=2.6cm, minimum height=1.0cm, inner sep=2pt},
  dens/.style={rectangle, rounded corners=2pt, draw=black!55, fill=orange!12,
               align=center, minimum width=2.6cm, minimum height=1.0cm, inner sep=2pt},
  densExp/.style={rectangle, rounded corners=2pt, draw=red!60, fill=red!15,
                  align=center, minimum width=2.6cm, minimum height=1.0cm, inner sep=2pt},
  jnode/.style={circle, draw=black!60, fill=white, inner sep=0.5pt, minimum size=5mm, font=\small},
  arr/.style={->, thick, draw=black!70},
  pipe/.style={-, thick, draw=black!60},
  lab/.style={fill=white, inner sep=2pt, font=\scriptsize}
]
\node[intens] (Ld)  at (0,    0) {\(\lambda_d^*(t)\)\\\scriptsize per-type intensity};
\node[intens] (L)   at (4.0,  0) {\(\lambda^*(t)\)\\\scriptsize ground intensity};
\node[trans]  (Lam) at (8.5,  0) {\(\Lambda^*(t_{i-1}, t)\)\\\scriptsize cumulative intensity};
\node[trans]  (S)   at (12.8, 0) {\(S^*(\tau)\)\\\scriptsize survival function};
\node[trans]  (F)   at (16.6, 0) {\(F^*(\tau)\)\\\scriptsize CDF of \(\tau\)};

\draw[arr] (Ld)  -- node[lab, above]{\(\sum_{d=1}^{D}\)}                (L);
\draw[arr] (L)   -- node[lab, above]{\(\int_{t_{i-1}}^{t}\!\cdot\,ds\)} (Lam);
\draw[arr] (Lam) -- node[lab, above]{\(\exp(-\,\cdot\,)\)}              (S);
\draw[arr] (S)   -- node[lab, above]{\(1 - \cdot\)}                     (F);

\node[jnode] (jpdt) at (2.0,  -2.0) {\(/\)};
\node[jnode] (jptd) at (6.4,  -2.0) {\(\times\)};
\node[jnode] (jpt)  at (10.6, -2.0) {\(\times\)};

\node[dens] (pdt) at (2.0,  -4.0) {\(p^*(d \mid t)\)\\\scriptsize type conditional};
\node[dens] (ptd) at (6.4,  -4.0) {\(p^*(t, d)\)\\\scriptsize joint density};
\node[dens] (pt)  at (10.6, -4.0) {\(p^*(t)\)\\\scriptsize time marginal};
\node[densExp] (pd)  at (14.8, -4.0) {\(p^*(d)\)\\\scriptsize type marginal};

\draw[pipe] (Ld.south)      to[out=-90, in=130] (jpdt.north west);
\draw[pipe] (L.south)       to[out=-90, in=50]  (jpdt.north east);

\draw[pipe] (Ld.south east) to[out=-30, in=130] (jptd.north west);
\draw[pipe] (S.south west)  to[out=-150, in=50] (jptd.north east);

\draw[pipe] (L.south east)  to[out=-30, in=130] (jpt.north west);
\draw[pipe] (S.south)       to[out=-90, in=50]  (jpt.north east);

\draw[arr] (jpdt.south) -- node[lab, right]{\(\lambda_d^*(t)/\lambda^*(t)\)}     (pdt.north);
\draw[arr] (jptd.south) -- node[lab, right]{\(\lambda_d^*(t)\cdot S^*(\tau)\)}   (ptd.north);
\draw[arr] (jpt.south)  -- node[lab, right]{\(\lambda^*(t)\cdot S^*(\tau)\)}     (pt.north);

\draw[arr] (ptd) -- node[lab, above]{\(\sum_{d=1}^{D}\)} (pt);
\draw[arr] (ptd.south east) to[out=-40, in=-140] node[lab, below, pos=0.5]{\(\int_{t_{i-1}}^{\infty}\!\cdot\,dt\)} (pd.south west);

\node[jnode] (jtcond) at (10.6, -6.6) {\(/\)};
\node[densExp]  (tcond)  at (10.6, -8.1) {\(p^*(t \mid d)\)\\\scriptsize time conditional};

\draw[pipe] (ptd.south) to[out=-60, in=170] (jtcond.north west);
\draw[pipe] (pd.south)  to[out=-120, in=10] (jtcond.north east);
\draw[arr]  (jtcond.south) -- node[lab, right]{\(p^*(t,d)\,/\,p^*(d)\)} (tcond.north);

\end{tikzpicture}}
\caption{Relationships between the key quantities of a multivariate temporal point process.
The top row constructs the survival objects from the per-type conditional intensities \(\lambda_d^*(t)\): summation across types yields the ground intensity \(\lambda^*(t)\), integration over time yields the cumulative intensity \(\Lambda^*(t_{i-1}, t)\), and the survival \(S^*(\tau)\) and CDF \(F^*(\tau)\) of the next-event time follow from \(\Lambda^*\) by \(\exp(-\,\cdot\,)\) and \(1 - \cdot\).
The middle row shows how the next-event densities are constructed at merge nodes (\(\times\), \(/\)): each density is formed by combining two top-row quantities, and the resulting expression is the label on the outgoing arrow.
In addition, the time marginal is recovered as \(p^*(t) = \sum_d p^*(t, d)\), the type marginal as \(p^*(d) = \int_{t_{i-1}}^{\infty} p^*(t, d)\,dt\), and the time conditional as \(p^*(t \mid d) = p^*(t, d) / p^*(d)\).
Quantities shown in red (\(p^*(d)\) and \(p^*(t \mid d)\)) are expensive to evaluate when starting from intensities: both require the unbounded integral defining \(p^*(d)\), which has no closed form except for trivial cases such as the homogeneous Poisson process and otherwise demands numerical quadrature with \(\Lambda^*\) evaluated at every quadrature point.
Stars denote conditioning on the history \(\mathcal{H}_{t_{i-1}}\); \(\tau = t - t_{i-1}\).}
\label{fig:tpp_relationships}
\end{figure}
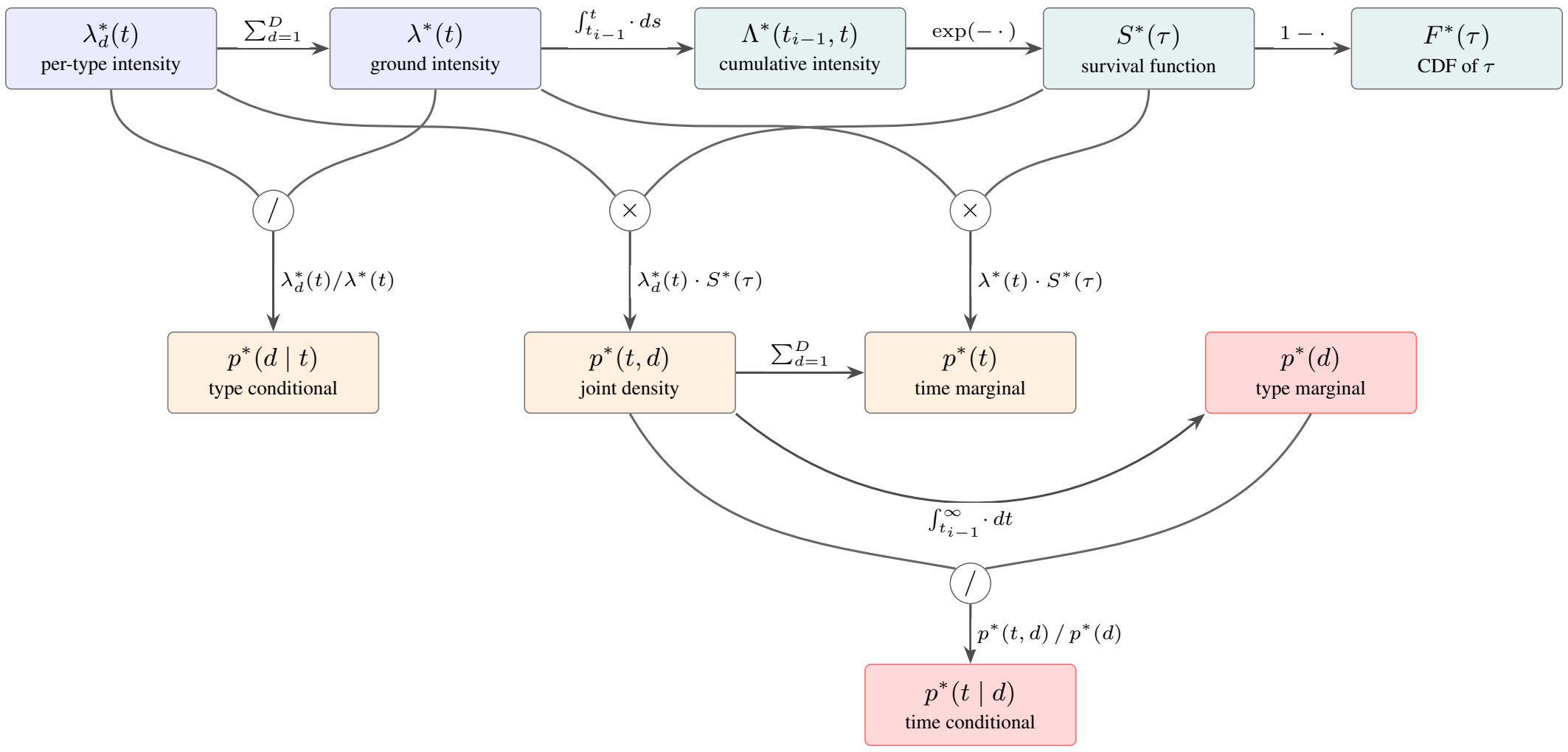

This appendix provides a self-contained derivation of the key quantities used throughout the paper: the conditional density, cumulative intensity, survival function, log-likelihood, marginal distributions, and sampling algorithms.
Figure~\ref{fig:tpp_relationships} summarizes how these objects relate to one another.

\subsection{Conditional Intensity and Cumulative Intensity}

A temporal point process on \([0, T]\) is fully characterized by its \emph{conditional intensity function}
\begin{equation}
    \lambda^*(t) = \lim_{\Delta t \to 0} \frac{\Pr(\text{event in } [t, t + \Delta t) \mid \mathcal{H}_t)}{\Delta t},
\end{equation}
where \(\mathcal{H}_t = \{(t_i, d_i) : t_i < t\}\) denotes the history of events before time \(t\).
In the multivariate case with \(D\) event types, the ground (total) intensity is \(\lambda^*(t) = \sum_{d=1}^{D} \lambda_d^*(t)\).

The \emph{cumulative intensity} between times \(a\) and \(b\) is defined as
\begin{equation}
    \Lambda^*(a, b) = \int_a^b \lambda^*(s) \, ds = \sum_{d=1}^{D} \int_a^b \lambda_d^*(s) \, ds = \sum_{d=1}^{D} \Lambda_d^*(a, b),
    \label{eq:ci}
\end{equation}
which is the compensator of the counting process. Within one inter-event interval, where the history is fixed, it is the cumulative hazard of the next event.

\subsection{Survival Function and the Inter-Event Distribution}

Consider the \(i\)-th inter-event interval.
Let \(\tau = t - t_{i-1}\) denote the time elapsed since the last event. Here and below, conditioning on \(\mathcal{H}_{t_{i-1}}\) is shorthand for \(\mathcal{H}_t\) with \(t \in (t_{i-1}, t_i]\), i.e.\ the history including \((t_{i-1}, d_{i-1})\), which is fixed within the interval.
The probability that \emph{no} event of any type occurs in \((t_{i-1}, t_{i-1} + \tau]\) is given by the \emph{survival function}:
\begin{align}
    S^*(\tau) &= \Pr\!\big(\text{no event in } (t_{i-1},\, t_{i-1} + \tau] \mid \mathcal{H}_{t_{i-1}}\big) \nonumber \\
              &= \exp\!\Big(-\Lambda^*(t_{i-1},\, t_{i-1} + \tau)\Big) = \exp\!\Big(-\int_{t_{i-1}}^{t_{i-1}+\tau} \lambda^*(s)\,ds\Big).
  \label{eq:survival}
\end{align}
This is the multivariate analogue of the standard survival function in survival analysis: the cumulative intensity acts as the cumulative hazard.

The density of the \emph{next event} occurring at time \(t\) with type \(d\), conditioned on history, is
\begin{equation}
  p^*(t, d \mid \mathcal{H}_{t_{i-1}}) = \lambda_d^*(t) \cdot S^*(t - t_{i-1}) = \lambda_d^*(t) \exp\!\Big(-\Lambda^*(t_{i-1}, t)\Big).
  \label{eq:joint_density}
\end{equation}
Intuition: the event must happen at time \(t\) (rate \(\lambda_d^*(t)\)) \emph{and} nothing must have happened before \(t\) (survival \(S^*\)).

\subsection{Deriving the Log-Likelihood}

The log-likelihood of a complete sequence \(\mathcal{H}_T = \{(t_1, d_1), \ldots, (t_N, d_N)\}\) observed on \([0, T]\) factorizes over inter-event intervals.
We sum the \(N\) event contributions from Equation~\ref{eq:joint_density} together with the survival probability that no event occurs between \(t_N\) and the observation end \(T\):
\begin{align}
  \log p(\mathcal{H}_T) &= \sum_{i=1}^{N} \log p^*(t_i, d_i \mid \mathcal{H}_{t_{i-1}}) + \log S^*(T - t_N) \nonumber \\
  &= \sum_{i=1}^{N} \Big[\log \lambda_{d_i}^*(t_i) - \Lambda^*(t_{i-1}, t_i)\Big] - \Lambda^*(t_N, T) \nonumber \\
  &= \sum_{i=1}^{N} \log \lambda_{d_i}^*(t_i) - \sum_{i=1}^{N} \Lambda^*(t_{i-1}, t_i) - \Lambda^*(t_N, T) \nonumber \\
  &= \sum_{i=1}^{N} \log \lambda_{d_i}^*(t_i) - \Lambda^*(0, T) \nonumber \\
  &= \underbrace{\sum_{i=1}^{N} \log \lambda_{d_i}^*(t_i)}_{\text{event log-intensities}} - \underbrace{\int_0^T \lambda^*(s)\,ds}_{\text{cumulative intensity}}.
  \label{eq:loglik_derived}
\end{align}
The combination in the penultimate step uses \(\Lambda^*(0,T) = \sum_{i=1}^{N}\Lambda^*(t_{i-1}, t_i) + \Lambda^*(t_N, T)\) with \(t_0 \equiv 0\).
This recovers the main-text log-likelihood expression.

The first term rewards high intensity at observed event times, while the second term penalizes high intensity everywhere, acting as a normalizer.

\subsection{Decomposing the Joint Density\texorpdfstring{: \(p(t, d) = p(t) \cdot p(d \mid t)\)}{}}

The joint conditional density from Equation~\ref{eq:joint_density} admits a natural decomposition into a \emph{time marginal} and a \emph{type conditional}.

\paragraph{Marginal time density \(p^*(t \mid \mathcal{H})\).}
Summing Equation~\ref{eq:joint_density} over all types \(d\):
\begin{equation}
  p^*(t \mid \mathcal{H}_{t_{i-1}}) = \sum_{d=1}^{D} p^*(t, d \mid \mathcal{H}_{t_{i-1}}) = \lambda^*(t) \exp\!\Big(-\Lambda^*(t_{i-1}, t)\Big).
  \label{eq:time_marginal}
\end{equation}
This is the density of the next-event time regardless of its type.
It has the familiar form of hazard times survival.

\paragraph{Conditional type distribution \(p^*(d \mid t, \mathcal{H})\).}
Dividing Equation~\ref{eq:joint_density} by Equation~\ref{eq:time_marginal}:
\begin{equation}
  p^*(d \mid t, \mathcal{H}) = \frac{p^*(t, d \mid \mathcal{H})}{p^*(t \mid \mathcal{H})} = \frac{\lambda_d^*(t)}{\lambda^*(t)} = \frac{\lambda_d^*(t)}{\sum_{d'=1}^{D} \lambda_{d'}^*(t)}.
  \label{eq:type_conditional}
\end{equation}
Given the next-event time \(t\), the type is sampled categorically with probabilities proportional to the per-type intensities.
Note that the survival terms cancel exactly, so the type distribution depends only on the intensity ratios at time \(t\).

\paragraph{Marginal type distribution \(p^*(d \mid \mathcal{H})\).}
The probability that the next event is of type \(d\), regardless of when it occurs, is obtained by marginalizing out the time:
\begin{equation}
  p^*(d \mid \mathcal{H}_{t_{i-1}}) = \int_{t_{i-1}}^{\infty} p^*(t, d \mid \mathcal{H}_{t_{i-1}}) \, dt = \int_{t_{i-1}}^{\infty} \lambda_d^*(t) \exp\!\Big(-\Lambda^*(t_{i-1}, t)\Big) \, dt.
  \label{eq:type_marginal}
\end{equation}
This integral generally has no closed form, except when the type distribution does not depend on \(t\) (e.g.\ for time-constant intensities such as the homogeneous Poisson process or NeuralConstTPP, or for IntensityFree), and must otherwise be computed numerically, for example by quadrature over the next-event time density.
\subsection{Sampling: Inversion Method}
\label{app:sampling_inversion}

The time-rescaling theorem~\citep{brown2002timerescaling} states that if \(\{t_i\}\) are event times of a point process with intensity \(\lambda^*(t)\), then the rescaled times \(\Lambda^*(0, t_i)\) form a unit-rate Poisson process, that is, the increments \(u_i = \Lambda^*(t_{i-1}, t_i)\) are i.i.d.\ \(\mathrm{Exp}(1)\). This requires \(\Lambda^*(t_{i-1}, \infty) = \infty\), which holds for the closed-form models evaluated in Table~\ref{tab:calibration}, as their baseline and head splines are extrapolated beyond the last knot as a constant.
To see this, note the relationship between CDF, survival, and cumulative intensity:

\begin{enumerate}
  \item The survival function is \(S^*(\tau) = \exp(-\Lambda^*(t_{i-1}, t_{i-1} + \tau))\).
  \item The CDF is \(F^*(\tau) = 1 - S^*(\tau) = 1 - \exp(-\Lambda^*(t_{i-1}, t_{i-1} + \tau))\).
  \item Substituting \(u = \Lambda^*(t_{i-1}, t_{i-1} + \tau)\) gives \(\Pr(u_i \leq u \mid \mathcal{H}_{t_{i-1}}) = 1 - e^{-u}\), so \(u_i \sim \mathrm{Exp}(1)\) independently of the history.
\end{enumerate}

This provides an exact sampling algorithm via \emph{cumulative intensity inversion}:

\begin{enumerate}
  \item Draw \(u \sim \mathrm{Exp}(1)\), equivalently \(u = -\log(1 - v)\) with \(v \sim \mathrm{Uniform}(0,1)\).
  \item Find \(t^*\) such that \(\Lambda^*(t_{i-1}, t^*) = u\), that is, invert the cumulative intensity.
  \item Sample the event type \(d^* \sim \mathrm{Categorical}\big(\lambda_1^*(t^*)/\lambda^*(t^*), \ldots, \lambda_D^*(t^*)/\lambda^*(t^*)\big)\).
  \item If \(t^* > T\), no further events occur; otherwise record \((t^*, d^*)\) and repeat.
\end{enumerate}

Step 2 requires solving a root-finding problem: find \(t^*\) such that \(\int_{t_{i-1}}^{t^*} \lambda^*(s)\,ds = u\).
When the cumulative intensity is available in closed form, this is solved with fast bracketed Newton iterations.
For models without a closed-form cumulative intensity, numerical quadrature is used to evaluate \(\Lambda^*\) within the root finder.

\paragraph{Advantages over thinning.}
Inversion does not require an upper bound on the intensity, so it is unbiased by construction whenever the cumulative intensity can be evaluated.
It is also more efficient when the intensity varies by orders of magnitude over the observation window, as is common in medical event data with long quiescent periods.
However, if no closed-form cumulative intensity is available, and an upper bound on the intensity can be estimated, Ogata's thinning algorithm is more reliable, as it does not compound integration errors.

\subsection{Sampling: Ogata's Thinning Algorithm}
\label{app:sampling_thinning}

Ogata's thinning algorithm~\citep{ogata1981lewis} is the classical approach for sampling from a TPP with intensity \(\lambda^*(t)\):

\begin{enumerate}
  \item Starting from the last event time \(t_{i-1}\), compute an upper bound \(\bar{\lambda} \geq \sup_{s > t_{i-1}} \lambda^*(s)\), or a local upper bound on some interval.
  \item Draw a candidate inter-event time \(\Delta t \sim \mathrm{Exp}(\bar{\lambda})\).
  \item Accept the candidate \(s = t_{i-1} + \Delta t\) with probability \(\lambda^*(s) / \bar{\lambda}\); otherwise reject it and repeat from \(s\).
  \item If accepted, sample the type as in the inversion method.
\end{enumerate}

The overall acceptance rate is the average of \(\lambda^*(s) / \bar{\lambda}\) over the proposed points.
When \(\bar{\lambda}\) is much larger than the typical intensity, for example due to rare high-intensity bursts, the algorithm becomes very inefficient, generating many rejected proposals.
Adaptive thinning with piecewise upper bounds mitigates this but adds implementation complexity.
Critically, if the upper bound is set too low, below the true supremum, the algorithm produces biased samples. This is especially a problem for neural TPPs, as even finding the upper bound over an interval can be challenging.

\section{Model Details}
\label{app:model_details}

Let \(t_\ell\) denote the most recent event time before a query time \(t\), let \(\tau = t - t_\ell\), and let \(h_\ell\) be the context embedding summarizing the history up to \(t_\ell\) whenever a neural encoder is used.

\subsection{Baseline-Only Models}

The homogeneous Poisson model uses a constant per-type rate,
\begin{equation}
  \lambda_d^*(t) = \mu_d.
\end{equation}
The inhomogeneous baseline-only models replace the constant rate by a learned time-dependent baseline \(\mu_d(t)\), implemented in TorchPP as spline interpolation over a small set of knots. For spline Poisson models this yields
\begin{equation}
  \lambda_d^*(t) = \mu_d(t),
\end{equation}
with analytical cumulative intensity given by the integral of the corresponding spline. 
The spline knot heights and the constant base rate are passed through a softplus to ensure non-negativity. Beyond the last knot, the baseline spline is extrapolated as a constant.

\subsection{Additive Hawkes Models}

All Hawkes variants share the additive structure
\begin{equation}
  \lambda_d^*(t) = \mu_d(t) + \sum_{j : t_j < t} \phi_{d_j \to d}(t - t_j).
\end{equation}
The  models differ only in the chosen baseline \(\mu_d\) and kernel \(\phi_{d' \to d}\).
For exponential-kernel Hawkes models,
\begin{equation}
  \phi_{d' \to d}(\tau) = \alpha_{d'd} \beta_{d'd} \exp(-\beta_{d'd} \tau), \qquad \tau > 0,
\end{equation}
which gives a closed-form cumulative intensity. Spline-kernel variants keep the same additive decomposition but replace the exponential kernel by a linear spline over elapsed time. Its integral, and hence the cumulative intensity, is a monotone piecewise-quadratic function in closed form.

\subsection{Inhibitive Hawkes Models}

To model inhibition, TorchPP wraps the additive pre-activation in a positivity map \(g\), typically \textit{Softplus},
\begin{equation}
  \lambda_d^*(t) = g\!\left(\mu_d(t) + \sum_{j : t_j < t} \phi_{d_j \to d}(t - t_j)\right).
\end{equation}
This allows negative intermediate contributions while preserving a non-negative intensity. Because the outer nonlinearity destroys the closed-form integral, these models use numerical quadrature for the cumulative intensities.

\subsection{Transformer-Conditioned Neural TPPs}

The simplest neural family predicts a history-dependent intensity that is constant between events,
\begin{equation}
  \lambda_d^*(t) = \exp\big(w_d^\top h_\ell + b_d\big),
\end{equation}
which corresponds to the \emph{NeuralConstTPP} model. The \emph{NeuralSplineTPP} model predicts spline knot heights from \(h_\ell\) and evaluates the intensity as a positive spline function of \(\tau\), with analytical integration over each spline segment.
The same Transformer backbone also supports THP- and SAHP-style heads, whose softplus intensities are integrated numerically.

\subsection{Recurrent and Intensity-Free Neural TPPs}
\emph{NHP}~\citep{nhp2017} encodes the history with a continuous-time LSTM whose cell state decays between events, and maps the hidden state at time \(t\) to per-type intensities with a scaled softplus. Its cumulative intensity is integrated numerically.
\emph{IntensityFree}~\citep{IntensityFree} is a density-based model: instead of an intensity it parametrizes the next inter-event time directly as a mixture of \(64\) log-normal components, \(p(\tau \mid h_\ell) = \sum_k w_k(h_\ell)\, \mathrm{LogNormal}(\tau; \mu_k(h_\ell), \sigma_k(h_\ell))\), and the next type by a categorical distribution \(p(d \mid h_\ell)\) that does not depend on \(\tau\). Both are available in closed form, so no numerical integration is needed. By default we use a Transformer encoder, and the GRU encoder of the original model is compared in Table~\ref{tab:iftpp_encoder}.

\section{Hyperparameter Tuning and Training Details}
\label{app:hyperparams}

Hyperparameters were selected by staged grid searches implemented directly in the repository scripts. The EasyTPP benchmarks, UK Biobank, and MIMIC-IV use separate search pipelines because they differ substantially in sequence length, censoring, and the stability of numerical models. All three pipelines share the same overall structure described below; per-dataset deviations are listed in the corresponding subsections.

Scripts with the exact configurations for reproduction can be found in our repository.

\paragraph{Common training settings.}
All tuning runs use AdamW with linear warmup followed by cosine annealing to \(\eta_{\min} = \mathrm{LR}/100\), gradient clipping at \(1.0\), and per-event normalized loss. Validation log-likelihood is computed every \(250\) optimization steps on the standard benchmarks (every \(500\) on UKB and MIMIC-IV), and early stopping with patience \(4\) (\(6\) on UKB and MIMIC-IV) is applied on the validation joint log-likelihood. Bucketed dynamic batching is used throughout: a small-bucket preset (\texttt{bucket\_default\_small}) for analytical Hawkes and standard neural runs, and a smaller numerical preset (\texttt{bucket\_numerical\_small}) for the inhibitive Hawkes variants, which require numerical cumulative-intensity integration. On the standard benchmarks, spline baselines use \(T_\mathrm{factor}=1.25\), and spline kernels use a knot-power schedule of \(4.0\) (denser knots near \(0\) to capture short-range excitation). The UKB and MIMIC-IV settings are given below.

\paragraph{Two-stage search.}
The grid is partitioned into two stages so that expensive models inherit architectural and time-scale choices from cheaper ones:
\begin{itemize}
\item \textbf{Stage~1}: cheap analytical families, i.e.\ Poisson, Spline Poisson (sweeping the baseline knot count \(K_b\)), constant-baseline exponential Hawkes, constant-baseline spline-kernel Hawkes (sweeping kernel knots \(K_k\) and time-scale factor \(T_f\)), and NeuralConstTPP (sweeping Transformer depth \(L\), heads \(H\), and embedding dimension \(d\)). Some pipelines also include variants that are not reported in this paper. Stage~1 selects the best \(K_b\), \((K_k, T_f)\), and \((L, H, d)\) per dataset, which fixes the architectural backbone and the inter-event time scale for the rest of the search.
\item \textbf{Stage~2}: expressive analytical models and neural heads conditioned on the Stage~1 winners, i.e.\ spline-baseline exponential Hawkes, spline-baseline spline-kernel Hawkes, the inhibitive Hawkes variants with softplus clamping, and NeuralSplineTPP (with a small refinement grid for \((K_k, T_f)\) on top of the Stage~1 backbone).
\end{itemize}
The exact per-dataset grids are summarized in the subsections below.

\subsection{EasyTPP Benchmarks}
\label{app:hyperparams_easytpp}

The benchmark search is orchestrated by \path{scripts/tuning/tune_easytpp_combined_local.sh}. Tuning runs use \(20{,}000\) optimization steps, learning rate \(0.1\) for Hawkes-family models and \(0.01\) for neural models, \(500\)-step warmup, and the smaller bucket preset. Stage~1 sweeps \(K_b \in \{5, 10, 15\}\) for Spline Poisson; \(K_k \in \{2,4,8\}\) and \(T_f \in \{5,10,15,20\}\) (multiples of the mean inter-event time) for the spline-kernel Hawkes; and the transformer backbone over \(L \in \{2,4\}\), \(H \in \{2,4\}\), \(d \in \{32,64\}\). Stage~2 reuses the Stage~1 winners and adds a refinement grid \(K_k \in \{2,4,8\}\), \(T_f \in \{5,10,15,20\}\) for NeuralSplineTPP. The configuration of the reported runs is given in Appendix~\ref{app:hyperparams_final}.

\subsection{UK Biobank}
\label{app:hyperparams_ukb}

The UKB search is implemented in \path{scripts/tuning/tune_ukb_real_combined_local.sh}. Because the vocabulary is much larger and sequences span decades, the defaults are more conservative: \(50{,}000\) optimization steps and learning rate \(0.01\) for Hawkes families and \(0.001\) for neural families. Stage~1 sweeps \(K_b \in \{15, 20\}\) for Spline Poisson, \((K_k, T_f) \in \{5,8,10\} \times \{15,20\}\) for the spline-kernel Hawkes, and the transformer backbone over \(L \in \{8,12\}\), \(H \in \{8,16\}\), \(d \in \{64,128,256\}\) (substantially wider than the EasyTPP grid, reflecting the \(1{,}257\)-type vocabulary). Stage~2 reuses the Stage~1 backbone and sweeps NeuralSplineTPP over \((K_k, T_f) \in \{5,8,10\} \times \{15,20\}\). Inhibitive Hawkes processes were not evaluated on UKB because of the cost of numerical integration at \(D = 1{,}257\). The reported UKB runs are defined in \path{scripts/tuning/generate_replica_commands_ukb.sh}. They use spline time-scale factors \(T_\mathrm{factor} = 1.0\), a kernel knot power of \(4.5\), \(100\) integration points for the numerically integrated neural models, and validation on \(2{,}000\) sequences. THP and SAHP were trained for \(20{,}000\) steps with \(50\) integration points and learning rates \(0.01\) and \(0.1\), respectively.

\subsection{MIMIC-IV ICU}
\label{app:hyperparams_mimic}

The MIMIC-IV search is implemented in \path{scripts/tuning/tune_mimic_iv_icu_combined_local.sh} on the top-\(116\) ICU intervention vocabulary. Tuning runs use \(20{,}000\) optimization steps, \(1{,}000\)-step warmup, and learning rate \(0.001\). Hawkes and standard neural runs use batch size \(64\) with \(2\)-step gradient accumulation, and inhibitive runs use batch size \(32\) with \(4\)-step accumulation.

\paragraph{Grids.} The search sweeps \(K_b \in \{5, 8, 10\}\) for Spline Poisson, \(K_k \in \{4, 8, 12\}\) and \(T_f \in \{2, 5, 10\}\) for the constant-baseline spline-kernel Hawkes, and the Transformer backbone over \(L \in \{2, 4\}\), \(H \in \{2, 4\}\), \(d \in \{32, 64, 128\}\), alongside the Poisson and exponential Hawkes baselines.

\paragraph{Reported configuration.} The reported MIMIC-IV runs use \(K_b = 20\) and \(T_\mathrm{factor} = 1.0\) for Spline Poisson and Spline/Exp Hawkes, \(T_f = 10\) for the spline kernels, \(T_f = 2\) for NeuralSplineTPP, and a Transformer with \(L = 4\), \(H = 4\), \(d = 64\), dropout \(0.1\) and weight decay \(10^{-4}\) for all Transformer models (IntensityFree: \(d = 32\), no weight decay). The kernel and head knot counts are selected as described in Appendix~\ref{app:hyperparams_final}.

\subsection{Training Configuration}
\label{app:hyperparams_final}
The results on the standard benchmarks and MIMIC-IV use the following configuration, with five seeds per model. All models are trained for \(40{,}000\) steps with early stopping on the validation log-likelihood, and the best validation checkpoint is restored before testing. The effective batch size is 128 throughout, using gradient accumulation where needed. Spline/Spline Hawkes and its inhibitive variant use 100 baseline knots, and the number of kernel/head spline knots of these models and of NeuralSplineTPP is selected on the validation log-likelihood from \(K \in \{10, 20, 50, 100, 200, 400\}\). Numerically integrated neural models use per-interval integration with 10 points, and the inhibitive Hawkes processes use 100 grid points. Increasing the points of the inhibitive Hawkes processes to 10{,}000 (MIMIC-IV: 3{,}000) changes \(\log p(t)\) of seed 0 by at most 0.08 nats and leaves the type terms unchanged. Taobao and MIMIC-IV event times are stored in float64. The selected kernel knots are 20/10/400/400/20/20 for Spline/Spline Hawkes and 50/20/400/400/20/10 for NeuralSplineTPP on Taxi/StackOverflow/Retweet/Amazon/Taobao/MIMIC-IV. The inhibitive variant inherits them. Learning rates are \(0.1\) for Poisson and Hawkes models and \(0.01\) for neural models on the standard benchmarks, and \(0.1\) and \(0.005\) on MIMIC-IV. They were adjusted per model and dataset where training did not converge (e.g.\ inhibitive Hawkes \(0.03\)--\(1.0\), NHP \(0.003\) on MIMIC-IV, and NeuralSplineTPP \(0.003\) on Taobao). IntensityFree uses \(0.01\) throughout. On MIMIC-IV, NeuralConstTPP uses a Softplus head, as its exponential head diverged for individual seeds. All values are part of the released configurations (\path{scripts/tuning/generate_gold_run_commands.sh}). The UKB models use the configuration of Appendix~\ref{app:hyperparams_ukb}. The knot-count sweep (Table~\ref{tab:ksweep_mimic}, Figure~\ref{fig:flex_decomp}) is run separately (20{,}000 steps on the standard benchmarks, learning rates 0.1 and 0.01 without per-model adjustment).

\subsection{Tuning Budget per Model Family}
\label{app:tuning_budget}
Table~\ref{tab:tuning_budget} summarizes the search effort per model family. The neural families received more search configurations than the Hawkes families (\(\approx 27\) vs.\ \(\approx 20\) per standard benchmark dataset for the models in this paper), and at the validation-selected knot count the flexible Hawkes process and NeuralSplineTPP have about the same number of parameters on MIMIC-IV.

\begin{table}[t]
  \caption{Tuning budget per model family on the standard benchmarks (search configurations per dataset of the systematic grid), integration method, and trainable parameters on MIMIC-IV (\(D = 117\)) at the final configuration. Early stopping (patience 4 on the standard benchmarks, 6 on UKB/MIMIC-IV, best validation checkpoint restored) is identical for all families. \(K_b\), \(K_k\): baseline and kernel knots; \(T_f\): kernel support in mean inter-event times; \(L, H, d\): Transformer layers, heads, embedding dimension.}
  \label{tab:tuning_budget}
  \centering
  \small
  \resizebox{\textwidth}{!}{%
  \begin{tabular}{llllr}
    \toprule
    Family & Configs / dataset & Swept hyperparameters & Integration & Params (MIMIC-IV) \\
    \midrule
    Poisson, Const./Exp Hawkes & 2 & -- & closed form & 117 -- 27{,}495 \\
    Spline Poisson, flexible Hawkes & \(\approx 16\) & \(K_b\), \(K_k \times T_f\); \(K\) validation-selected & closed form & 2{,}340 -- 285{,}481 \\
    Inhibitive Hawkes & 2 & inherited from analytical twin, clamping function & numerical (grid) & 27{,}495 -- 285{,}481 \\
    NeuralConstTPP, NeuralSplineTPP & \(\approx 20\) & \(L \times H \times d\); \(K_k \times T_f\); \(K\) validation-selected & closed form & 216{,}554 -- 286{,}052 \\
    THP, SAHP & -- & architecture from EasyTPP, validated & numerical (per interval) & 216{,}905 / 231{,}764 \\
    NHP, IntensityFree & 7 & encoder (GRU/Transformer), \(L\), \(d\), learning rate & numerical / closed form & 73{,}066 / 64{,}746 \\
    \bottomrule
  \end{tabular}}
\end{table}

\subsection{Search Philosophy}
The tuning procedure is intentionally not a single flat Cartesian search over all models. Instead, the cheapest families are tuned first to identify reasonable optimization scales, transformer widths, and inter-event time scales; more expressive analytical or neural variants then inherit those choices and only spend their compute budget on model-specific parameters. This staged design substantially reduced the number of unstable runs on UKB and MIMIC-IV and made it feasible to compare analytical, numerical, and fully neural TPPs under a common experimental budget.

\section{Compute Resources and Libraries}
\label{app:compute}

The experiments were run on a heterogeneous in-house GPU cluster with NVIDIA TITAN RTX (\(24\)\,GB), NVIDIA L40S (\(48\)\,GB), NVIDIA A100 (\(40\)\,GB, both PCIe and SXM4), and NVIDIA H200 (\(141\)\,GB) cards. MIMIC-IV and UKB runs defaulted to L40S or A100 to fit the larger transformer backbones, and part of the UKB experiments were run on the UK Biobank Research Analysis Platform. All runs used one GPU per job. No multi-node or multi-GPU training was used, and the longest single run took about \(19\) wall-clock hours.

Table~\ref{tab:compute} summarizes the number of training runs and the total wall-clock GPU time per dataset, separated into hyperparameter tuning and replica runs (5 seeds per model and dataset, including models and variants that are not reported in the tables). The tuning numbers are extracted from the WandB logs and include pilot configurations that were discarded. In total, these runs comprise approximately \(1{,}210\) training runs and \(\sim 760\) GPU-hours (\(\sim 32\) GPU-days). We additionally reran the EasyTPP-internal baselines (THP, SAHP) using the original \texttt{EasyTPP} framework on the same TITAN RTX hardware for the equivalence check in Appendix~\ref{app:equivalence}; that retraining contributes another \({\sim}80\) runs (5 seeds \(\times\) 2 models \(\times\) 5 datasets, plus a small Taobao tuning sweep) and is not included in Table~\ref{tab:compute}.

Failed and preempted runs are not accounted for, as the WandB logs we summarize only retain runs that reached the final evaluation step. The knot-count sweeps, the non-stationarity control (35 runs), and learning-rate adjustment runs used the same hardware and are not included in Table~\ref{tab:compute}.

\begin{table}[ht]
  \caption{Compute used for the experiments in this paper. Run counts and wall-clock GPU-hours are aggregated from WandB logs of the tuning sweeps and of the replica runs, which include the runs that produced the numbers in the tables. ``Runs'' counts only runs that reached the final evaluation step, and runtime is wall-clock time on a single GPU. Some tuning sweeps logged only metrics and not runtime. These are included in the run counts but not in the GPU-hour totals, which therefore slightly underestimate the true cost.}
  \label{tab:compute}
  \centering
  \small
  \begin{tabular}{lrrrr}
    \toprule
    & \multicolumn{2}{c}{Hyperparameter tuning} & \multicolumn{2}{c}{Replica runs} \\
    \cmidrule(lr){2-3}\cmidrule(lr){4-5}
    Dataset & Runs & GPU-h & Runs & GPU-h \\
    \midrule
    Standard benchmarks & 396 & 144 & 370 & 73 \\
    MIMIC-IV ICU & 28 & 4 & 75 & 156 \\
    UK Biobank & 205 & 289 & 133 & 98 \\
    \midrule
    Total & 629 & 437 & 578 & 327 \\
    \bottomrule
  \end{tabular}
\end{table}

Our experiments used EasyTPP 0.2.1 for comparisons. It is available under the Apache-2.0 License.
The exact software environment is available in our repository as a frozen \texttt{uv.lock} and \texttt{pyproject.toml}.

\section{UK Biobank Data Preparation and Access}
\label{app:ukb_generation}
\label{app:data_prep_ukb}

\paragraph{Access.}
Access to the UK Biobank requires an approved research application. We do not redistribute UK Biobank data, data derived from it, or models trained on it. The UKB results can therefore only be reproduced by approved UK Biobank users.

\paragraph{Preparation.} We adapt the cohort and event construction of~\citet{Delphi} but restrict the vocabulary to first-occurrence ICD-10 disease tokens and a single terminal \textit{Death} token, omitting the lifestyle and sex tokens of the original setup. Disease events are derived from ICD-10 codes truncated to three or four characters and joined to UK Biobank field IDs through the \texttt{icd10\_codes\_mod.tsv} mapping of the Delphi repository. After this mapping the vocabulary contains \(1{,}256\) disease codes plus a single \textit{Death} token, giving \(D = 1{,}257\) event types after the data loader's final remapping.

\paragraph{Cohort selection.}
A participant is included if they have a valid sex record. Records without a sex field are dropped, since sex is used both for cohort definition and as a sequence-start token in the original Delphi pipeline. We do not apply any further demographic or follow-up filters. Sex itself is not given to the models, and no other covariates are modeled. We list this as a limitation (Section~\ref{sec:discussion}).

\paragraph{Sequence construction and ordering.}
Each participant contributes exactly one sequence. Raw event timestamps are converted to days since the participant's date of birth and sorted by \texttt{(participant\_id, time)}. Strictly simultaneous events are kept. \textsc{TorchPP} handles them through causal masking.

\paragraph{Train/val/test split.}
The split is patient-level with fixed fractions \(0.7 / 0.2 / 0.1\). All events of a given participant are assigned to the same split, so there is no patient-level leakage across splits.

\section{MIMIC-IV ICU Data Preparation}
\label{app:data_prep_mimic}

The MIMIC-IV ICU task is built from the MIMIC-IV-3.1 release~\citep{mimic-iv, mimiciv31, goldberger2000physionet}, available to credentialed users via PhysioNet. The pipeline is split between \path{scripts/data_prep/prepare_mimic_iv_datasets.py}, which materializes the canonical sequences, and \path{scripts/data_prep/analyze_mimic_iv_datasets.py}, which computes vocabulary statistics. Both are orchestrated by \path{scripts/data_prep/recreate_mimic_iv_benchmarks.py}.

\paragraph{Source tables.}
We use \texttt{icustays.csv.gz} to define the per-stay observation windows \([\texttt{intime}, \texttt{outtime}]\), and merge events from \texttt{inputevents.csv.gz} and \texttt{procedureevents.csv.gz} into a single event stream per ICU stay. Item metadata (label, abbreviation, category) is read from \texttt{d\_items.csv.gz}, and no semantic re-grouping of items is applied.

\paragraph{Sequence unit.}
One sequence per ICU stay (\texttt{stay\_id}), \emph{not} per patient. A patient with multiple ICU stays therefore contributes multiple sequences. Within a stay, only events whose \texttt{starttime} falls inside \([\texttt{intime}, \texttt{outtime}]\) are retained, and stays with fewer than two events are dropped. Event times are recorded to the minute. We add to each event an offset drawn uniformly within its recorded minute (truncated at \texttt{outtime}, fixed seed) and sort the events by the resulting times. Event times are stored in float64.

\paragraph{Event vocabulary and \texttt{OTHER} class.}
The raw merged ICU vocabulary contains \(486\) \texttt{itemid}s. We retain the top-\(116\) \texttt{itemid}s by training-set frequency, which together cover \(\approx 95.1\%\) of all training events, and remap the remaining \(370\) items to a single \texttt{OTHER} class. The task vocabulary therefore has \(D = 117\). The \texttt{OTHER} class is treated as a regular event type during training and evaluation; it inherits a synthetic metadata entry and is not masked out.

\paragraph{Sequence-length truncation.}
The data preparation step retains all events within the observation window; pre-truncation sequence lengths reach a \(95\)\textsuperscript{th} percentile of roughly \(570\) events. To bound the quadratic cost of the Transformer encoders and the all-pairs Hawkes kernels, sequences are right-truncated to \(316\) events (\(90\)\textsuperscript{th} percentile of sequence lengths), reducing the average sequence length from \(138.8\) to \(94.4\) on the top-\(116\) vocabulary.

\paragraph{Train/val/test split.}
The split is subject-level with fractions \(0.8 / 0.1 / 0.1\). All ICU stays for a given subject are placed in the same split. With the full vocabulary of \(486\) items, the splits contain \(67{,}320 / 8{,}474 / 8{,}531\) stays and \(9{,}304{,}960 / 1{,}180{,}908 / 1{,}221{,}930\) events; pre-truncation mean sequence lengths are \(138.2\), \(139.4\), and \(143.2\), respectively. The top-\(116\) variant uses identical splits and stay assignments; only the vocabulary mapping differs.

\section{Full Result Tables}
\label{app:full_results}

Tables~\ref{tab:joint_ll_results}--\ref{tab:rmse_results} report all metrics with standard deviations for the standard benchmarks, Table~\ref{tab:full_mimic_results} for MIMIC-IV, and Table~\ref{tab:ukb_full_results} for UKB. Log-likelihoods and \(\mathrm{Acc}|t\) are computed in closed form (or with the training-time integration for the numerically integrated models). In contrast, \(\log p(d)\), the marginal accuracy, and the RMSE require integration over the next-event time on a dense grid (Section~\ref{sec:datasets-experiments/setup}). The RMSE is sensitive to this integration on datasets with heavy-tailed inter-event times (notably Retweet). On Taobao, whose test gaps are all below one, the RMSE does not separate the models.

\begin{table}[t]
  \caption{Joint test log-likelihood per event \(\log p(t, d \mid \mathcal{H})\) (nats, higher is better) on the standard benchmarks. It is the sum of Tables~\ref{tab:time_ll_results} and \ref{tab:cond_type_ll_results}. Mean \(\pm\) std over 5 seeds. \textbf{Bold}: best per dataset; \underline{underline}: best classical (Poisson or Hawkes) model; ties at the displayed precision are all marked.}
  \label{tab:joint_ll_results}
  \centering
  \small
  \setlength{\tabcolsep}{4pt}
  \begin{tabular}{lccccc}
    \toprule
    Model & Taxi & StackOverflow & Retweet & Amazon & Taobao \\
    \midrule
    \multicolumn{6}{l}{\textit{Poisson processes}} \\
    Homogeneous Poisson & \(-0.63_{\pm .00}\) & \(-2.76_{\pm .00}\) & \(-4.41_{\pm .00}\) & \(-2.60_{\pm .00}\) & \(-0.11_{\pm .00}\) \\
    Spline Poisson & \(-0.62_{\pm .00}\) & \(-2.73_{\pm .00}\) & \(-4.25_{\pm .00}\) & \(-2.59_{\pm .00}\) & \(-0.07_{\pm .00}\) \\
    \midrule
    \multicolumn{6}{l}{\textit{Hawkes processes}} \\
    Const./Exp Hawkes & \(-0.29_{\pm .00}\) & \(-2.38_{\pm .00}\) & \(-4.20_{\pm .00}\) & \(-2.13_{\pm .02}\) & \(1.19_{\pm .00}\) \\
    Spline/Exp Hawkes & \(-0.29_{\pm .00}\) & \(-2.35_{\pm .01}\) & \(-4.16_{\pm .00}\) & \(-2.15_{\pm .01}\) & \(1.19_{\pm .00}\) \\
    Const./Spline Hawkes & \(-0.21_{\pm .00}\) & \(-2.34_{\pm .00}\) & \(-4.19_{\pm .00}\) & \(-1.95_{\pm .00}\) & \underline{\(1.27_{\pm .00}\)} \\
    Spline/Spline Hawkes & \(-0.21_{\pm .00}\) & \underline{\(\mathbf{-2.32}_{\pm .01}\)} & \(-3.17_{\pm .00}\) & \(-0.31_{\pm .00}\) & \underline{\(1.27_{\pm .00}\)} \\
    \midrule
    \multicolumn{6}{l}{\textit{Inhibitive Hawkes processes}} \\
    Inhib. Const./Exp Hawkes & \(0.21_{\pm .00}\) & \(-2.35_{\pm .00}\) & \(-4.19_{\pm .00}\) & \(-2.01_{\pm .02}\) & \(1.18_{\pm .00}\) \\
    Inhib. Spline/Spline Hawkes & \underline{\(0.40_{\pm .00}\)} & \(-2.35_{\pm .00}\) & \underline{\(\mathbf{-2.09}_{\pm .02}\)} & \underline{\(-0.12_{\pm .01}\)} & \(1.19_{\pm .01}\) \\
    \midrule
    \multicolumn{6}{l}{\textit{Neural TPPs}} \\
    NeuralConstTPP & \(0.27_{\pm .00}\) & \(-2.40_{\pm .01}\) & \(-4.21_{\pm .01}\) & \(-2.33_{\pm .00}\) & \(0.64_{\pm .01}\) \\
    THP & \(0.30_{\pm .00}\) & \(-2.39_{\pm .01}\) & \(-4.19_{\pm .02}\) & \(-2.31_{\pm .01}\) & \(0.66_{\pm .01}\) \\
    SAHP & \(0.40_{\pm .02}\) & \(\mathbf{-2.32}_{\pm .01}\) & \(-4.17_{\pm .03}\) & \(-2.05_{\pm .06}\) & \(1.17_{\pm .05}\) \\
    NHP & \(0.36_{\pm .04}\) & \(-2.40_{\pm .02}\) & \(-4.32_{\pm .06}\) & \(-2.38_{\pm .04}\) & \(1.06_{\pm .01}\) \\
    \midrule
    \multicolumn{6}{l}{\textit{Flexible neural TPPs}} \\
    IntensityFree & \(0.44_{\pm .00}\) & \(-2.33_{\pm .02}\) & \(-4.45_{\pm .26}\) & \(0.37_{\pm .04}\) & \(\mathbf{1.30}_{\pm .01}\) \\
    NeuralSplineTPP & \(\mathbf{0.49}_{\pm .00}\) & \(\mathbf{-2.32}_{\pm .01}\) & \(-2.64_{\pm .01}\) & \(\mathbf{0.65}_{\pm .01}\) & \(1.28_{\pm .01}\) \\
    \bottomrule
  \end{tabular}
\end{table}

\begin{table}[t]
  \caption{Time log-likelihood per event \(\log p(t \mid \mathcal{H})\) (nats, higher is better) on the standard benchmarks (Equation~\ref{eq:time_marginal}). Mean \(\pm\) std over 5 seeds. \textbf{Bold}: best per dataset; \underline{underline}: best classical (Poisson or Hawkes) model; ties at the displayed precision are all marked.}
  \label{tab:time_ll_results}
  \centering
  \small
  \setlength{\tabcolsep}{4pt}
  \begin{tabular}{lccccc}
    \toprule
    Model & Taxi & StackOverflow & Retweet & Amazon & Taobao \\
    \midrule
    \multicolumn{6}{l}{\textit{Poisson processes}} \\
    Homogeneous Poisson & \(0.51_{\pm .00}\) & \(-0.86_{\pm .00}\) & \(-3.58_{\pm .00}\) & \(-0.36_{\pm .00}\) & \(2.07_{\pm .00}\) \\
    Spline Poisson & \(0.51_{\pm .00}\) & \(-0.85_{\pm .00}\) & \(-3.43_{\pm .00}\) & \(-0.35_{\pm .00}\) & \(2.11_{\pm .00}\) \\
    \midrule
    \multicolumn{6}{l}{\textit{Hawkes processes}} \\
    Const./Exp Hawkes & \(0.45_{\pm .00}\) & \(-0.76_{\pm .00}\) & \(-3.42_{\pm .00}\) & \(-0.10_{\pm .00}\) & \(2.61_{\pm .00}\) \\
    Spline/Exp Hawkes & \(0.45_{\pm .00}\) & \(-0.75_{\pm .00}\) & \(-3.38_{\pm .00}\) & \(-0.11_{\pm .00}\) & \(2.61_{\pm .00}\) \\
    Const./Spline Hawkes & \(0.50_{\pm .00}\) & \underline{\(-0.73_{\pm .00}\)} & \(-3.40_{\pm .00}\) & \(0.01_{\pm .00}\) & \(2.65_{\pm .00}\) \\
    Spline/Spline Hawkes & \(0.49_{\pm .00}\) & \underline{\(-0.73_{\pm .01}\)} & \(-2.37_{\pm .00}\) & \(1.73_{\pm .00}\) & \underline{\(2.67_{\pm .00}\)} \\
    \midrule
    \multicolumn{6}{l}{\textit{Inhibitive Hawkes processes}} \\
    Inhib. Const./Exp Hawkes & \(0.54_{\pm .00}\) & \(-0.75_{\pm .00}\) & \(-3.41_{\pm .00}\) & \(0.05_{\pm .02}\) & \(2.62_{\pm .00}\) \\
    Inhib. Spline/Spline Hawkes & \underline{\(0.68_{\pm .00}\)} & \(-0.76_{\pm .00}\) & \underline{\(\mathbf{-1.29}_{\pm .02}\)} & \underline{\(2.05_{\pm .01}\)} & \(2.66_{\pm .01}\) \\
    \midrule
    \multicolumn{6}{l}{\textit{Neural TPPs}} \\
    NeuralConstTPP & \(0.55_{\pm .00}\) & \(-0.80_{\pm .01}\) & \(-3.42_{\pm .00}\) & \(-0.34_{\pm .00}\) & \(2.07_{\pm .00}\) \\
    THP & \(0.56_{\pm .00}\) & \(-0.80_{\pm .00}\) & \(-3.41_{\pm .02}\) & \(-0.31_{\pm .01}\) & \(2.11_{\pm .00}\) \\
    SAHP & \(0.64_{\pm .02}\) & \(-0.75_{\pm .00}\) & \(-3.38_{\pm .01}\) & \(0.14_{\pm .06}\) & \(2.66_{\pm .01}\) \\
    NHP & \(0.60_{\pm .03}\) & \(-0.79_{\pm .01}\) & \(-3.51_{\pm .06}\) & \(-0.32_{\pm .03}\) & \(2.56_{\pm .01}\) \\
    \midrule
    \multicolumn{6}{l}{\textit{Flexible neural TPPs}} \\
    IntensityFree & \(\mathbf{0.73}_{\pm .00}\) & \(-0.73_{\pm .01}\) & \(-3.66_{\pm .26}\) & \(2.41_{\pm .01}\) & \(\mathbf{2.73}_{\pm .00}\) \\
    NeuralSplineTPP & \(0.71_{\pm .00}\) & \(\mathbf{-0.72}_{\pm .01}\) & \(-1.86_{\pm .01}\) & \(\mathbf{2.56}_{\pm .00}\) & \(2.71_{\pm .00}\) \\
    \bottomrule
  \end{tabular}
\end{table}

\begin{table}[t]
  \caption{Time-conditional type log-likelihood per event \(\log p(d \mid t, \mathcal{H})\) at the true next-event time (nats, higher is better; Equation~\ref{eq:type_conditional}). Mean \(\pm\) std over 5 seeds. \textbf{Bold}: best per dataset; \underline{underline}: best classical (Poisson or Hawkes) model; ties at the displayed precision are all marked.}
  \label{tab:cond_type_ll_results}
  \centering
  \small
  \setlength{\tabcolsep}{4pt}
  \begin{tabular}{lccccc}
    \toprule
    Model & Taxi & StackOverflow & Retweet & Amazon & Taobao \\
    \midrule
    \multicolumn{6}{l}{\textit{Poisson processes}} \\
    Homogeneous Poisson & \(-1.14_{\pm .00}\) & \(-1.90_{\pm .00}\) & \(-0.82_{\pm .00}\) & \(-2.24_{\pm .00}\) & \(-2.18_{\pm .00}\) \\
    Spline Poisson & \(-1.13_{\pm .00}\) & \(-1.88_{\pm .00}\) & \(-0.82_{\pm .00}\) & \(-2.24_{\pm .00}\) & \(-2.19_{\pm .00}\) \\
    \midrule
    \multicolumn{6}{l}{\textit{Hawkes processes}} \\
    Const./Exp Hawkes & \(-0.74_{\pm .00}\) & \(-1.62_{\pm .00}\) & \(-0.79_{\pm .00}\) & \(-2.03_{\pm .03}\) & \(-1.42_{\pm .00}\) \\
    Spline/Exp Hawkes & \(-0.74_{\pm .00}\) & \(-1.60_{\pm .00}\) & \underline{\(\mathbf{-0.78}_{\pm .00}\)} & \(-2.04_{\pm .00}\) & \(-1.43_{\pm .00}\) \\
    Const./Spline Hawkes & \(-0.71_{\pm .00}\) & \(-1.60_{\pm .00}\) & \(-0.79_{\pm .00}\) & \underline{\(-1.96_{\pm .00}\)} & \underline{\(\mathbf{-1.38}_{\pm .00}\)} \\
    Spline/Spline Hawkes & \(-0.70_{\pm .00}\) & \underline{\(-1.59_{\pm .00}\)} & \(-0.80_{\pm .00}\) & \(-2.04_{\pm .00}\) & \(-1.41_{\pm .00}\) \\
    \midrule
    \multicolumn{6}{l}{\textit{Inhibitive Hawkes processes}} \\
    Inhib. Const./Exp Hawkes & \(-0.33_{\pm .00}\) & \(-1.60_{\pm .00}\) & \underline{\(\mathbf{-0.78}_{\pm .00}\)} & \(-2.06_{\pm .01}\) & \(-1.43_{\pm .00}\) \\
    Inhib. Spline/Spline Hawkes & \underline{\(-0.28_{\pm .00}\)} & \(-1.60_{\pm .00}\) & \(-0.80_{\pm .00}\) & \(-2.17_{\pm .01}\) & \(-1.47_{\pm .00}\) \\
    \midrule
    \multicolumn{6}{l}{\textit{Neural TPPs}} \\
    NeuralConstTPP & \(-0.29_{\pm .00}\) & \(-1.59_{\pm .00}\) & \(\mathbf{-0.78}_{\pm .01}\) & \(-1.99_{\pm .00}\) & \(-1.43_{\pm .01}\) \\
    THP & \(-0.26_{\pm .00}\) & \(-1.59_{\pm .00}\) & \(\mathbf{-0.78}_{\pm .00}\) & \(-1.99_{\pm .00}\) & \(-1.45_{\pm .01}\) \\
    SAHP & \(-0.24_{\pm .00}\) & \(\mathbf{-1.57}_{\pm .01}\) & \(-0.79_{\pm .02}\) & \(-2.20_{\pm .02}\) & \(-1.49_{\pm .04}\) \\
    NHP & \(-0.24_{\pm .01}\) & \(-1.60_{\pm .01}\) & \(-0.81_{\pm .01}\) & \(-2.06_{\pm .02}\) & \(-1.51_{\pm .00}\) \\
    \midrule
    \multicolumn{6}{l}{\textit{Flexible neural TPPs}} \\
    IntensityFree & \(-0.29_{\pm .00}\) & \(-1.60_{\pm .01}\) & \(-0.79_{\pm .01}\) & \(-2.04_{\pm .03}\) & \(-1.42_{\pm .01}\) \\
    NeuralSplineTPP & \(\mathbf{-0.22}_{\pm .00}\) & \(-1.60_{\pm .01}\) & \(\mathbf{-0.78}_{\pm .00}\) & \(\mathbf{-1.91}_{\pm .01}\) & \(-1.43_{\pm .01}\) \\
    \bottomrule
  \end{tabular}
\end{table}

\begin{table}[t]
  \caption{Time-conditional top-1 type accuracy \(\mathrm{Acc}|t\) (\%, higher is better): \(\arg\max_d p(d \mid t, \mathcal{H})\) at the true next-event time, the next-type metric reported by EasyTPP~\citep{xue2024easytpp}. Mean \(\pm\) std over 5 seeds. \textbf{Bold}: best per dataset; \underline{underline}: best classical (Poisson or Hawkes) model; ties at the displayed precision are all marked.}
  \label{tab:cond_acc_results}
  \centering
  \small
  \setlength{\tabcolsep}{4pt}
  \begin{tabular}{lccccc}
    \toprule
    Model & Taxi & StackOverflow & Retweet & Amazon & Taobao \\
    \midrule
    \multicolumn{6}{l}{\textit{Poisson processes}} \\
    Homogeneous Poisson & \(44.29_{\pm .00}\) & \(42.50_{\pm .00}\) & \(53.92_{\pm .00}\) & \(29.87_{\pm .00}\) & \(43.57_{\pm .00}\) \\
    Spline Poisson & \(44.26_{\pm .10}\) & \(42.50_{\pm .00}\) & \(54.66_{\pm .06}\) & \(29.87_{\pm .00}\) & \(43.57_{\pm .00}\) \\
    \midrule
    \multicolumn{6}{l}{\textit{Hawkes processes}} \\
    Const./Exp Hawkes & \(85.00_{\pm .08}\) & \(46.16_{\pm .04}\) & \(57.49_{\pm .16}\) & \(35.03_{\pm 2.50}\) & \(60.96_{\pm .05}\) \\
    Spline/Exp Hawkes & \(84.74_{\pm .12}\) & \(46.16_{\pm .08}\) & \underline{\(58.65_{\pm .10}\)} & \(32.48_{\pm .01}\) & \(60.91_{\pm .06}\) \\
    Const./Spline Hawkes & \(83.28_{\pm .09}\) & \underline{\(46.72_{\pm .04}\)} & \(57.24_{\pm .20}\) & \(38.04_{\pm .08}\) & \underline{\(\mathbf{61.55}_{\pm .05}\)} \\
    Spline/Spline Hawkes & \(83.50_{\pm .02}\) & \(46.42_{\pm .17}\) & \(56.55_{\pm .12}\) & \(35.87_{\pm .03}\) & \(61.54_{\pm .06}\) \\
    \midrule
    \multicolumn{6}{l}{\textit{Inhibitive Hawkes processes}} \\
    Inhib. Const./Exp Hawkes & \(90.95_{\pm .04}\) & \(46.41_{\pm .03}\) & \(57.75_{\pm .26}\) & \(32.61_{\pm .96}\) & \(61.06_{\pm .02}\) \\
    Inhib. Spline/Spline Hawkes & \underline{\(92.00_{\pm .02}\)} & \(46.07_{\pm .33}\) & \(57.65_{\pm .08}\) & \underline{\(39.15_{\pm .21}\)} & \(61.23_{\pm .11}\) \\
    \midrule
    \multicolumn{6}{l}{\textit{Neural TPPs}} \\
    NeuralConstTPP & \(91.44_{\pm .06}\) & \(45.24_{\pm .11}\) & \(58.55_{\pm 1.20}\) & \(35.62_{\pm .06}\) & \(60.23_{\pm .16}\) \\
    THP & \(91.47_{\pm .06}\) & \(45.22_{\pm .16}\) & \(\mathbf{59.28}_{\pm .30}\) & \(35.70_{\pm .06}\) & \(60.33_{\pm .17}\) \\
    SAHP & \(92.35_{\pm .27}\) & \(\mathbf{46.79}_{\pm .08}\) & \(58.24_{\pm 2.56}\) & \(32.29_{\pm .75}\) & \(60.34_{\pm .57}\) \\
    NHP & \(92.26_{\pm .32}\) & \(46.76_{\pm .17}\) & \(56.20_{\pm 2.36}\) & \(33.30_{\pm 1.05}\) & \(60.91_{\pm .12}\) \\
    \midrule
    \multicolumn{6}{l}{\textit{Flexible neural TPPs}} \\
    IntensityFree & \(91.49_{\pm .07}\) & \(45.23_{\pm .16}\) & \(58.69_{\pm .94}\) & \(34.87_{\pm .66}\) & \(60.14_{\pm .27}\) \\
    NeuralSplineTPP & \(\mathbf{92.66}_{\pm .05}\) & \(46.61_{\pm .16}\) & \(58.70_{\pm .57}\) & \(\mathbf{39.30}_{\pm .48}\) & \(60.42_{\pm .30}\) \\
    \bottomrule
  \end{tabular}
\end{table}

\begin{table}[t]
  \caption{Marginal top-1 type accuracy \(\mathrm{Acc}\) (\%, higher is better): \(\arg\max_d p(d \mid \mathcal{H})\) from the marginal type distribution (Equation~\ref{eq:type_marginal}), computed by numerical integration. Grid-based metrics are evaluated on the first 1{,}000 test sequences. \(^{\ast}\): mean over 3--4 seeds, as the evaluation of a seed is missing or its RMSE diverged. Spline/Spline Hawkes on Amazon was not evaluated with the grid-based metrics. Mean \(\pm\) std over 5 seeds. \textbf{Bold}: best per dataset; \underline{underline}: best classical (Poisson or Hawkes) model; ties at the displayed precision are all marked.}
  \label{tab:acc_results}
  \centering
  \small
  \setlength{\tabcolsep}{4pt}
  \begin{tabular}{lccccc}
    \toprule
    Model & Taxi & StackOverflow & Retweet & Amazon & Taobao \\
    \midrule
    \multicolumn{6}{l}{\textit{Poisson processes}} \\
    Homogeneous Poisson & \(44.29_{\pm .00}\) & \(42.50_{\pm .00}\) & \(53.92_{\pm .00}\) & \(29.87_{\pm .00}\) & \(43.57_{\pm .00}\) \\
    Spline Poisson & \(44.25_{\pm .28}\) & \(42.50_{\pm .00}\) & \(54.71_{\pm .01}\) & \(29.87_{\pm .00}\) & \(43.57_{\pm .00}\) \\
    \midrule
    \multicolumn{6}{l}{\textit{Hawkes processes}} \\
    Const./Exp Hawkes & \(85.29_{\pm .40}\) & \(44.39_{\pm .06}\) & \(57.62_{\pm .12}\) & \(34.06_{\pm 1.22}\) & \(60.91_{\pm .05}\) \\
    Spline/Exp Hawkes & \(85.50_{\pm .28}\) & \(44.45_{\pm .03}\) & \underline{\(58.71_{\pm .11}\)} & \(33.21_{\pm .02}\) & \(60.79_{\pm .06}\) \\
    Const./Spline Hawkes & \(80.32_{\pm .13}\) & \(44.29_{\pm .05}\) & \(57.44_{\pm .13}\) & \underline{\(35.17_{\pm .06}\)} & \underline{\(\mathbf{61.20}_{\pm .08}\)} \\
    Spline/Spline Hawkes & \(79.27_{\pm .07}\) & \(44.34_{\pm .02}\) & \(57.71_{\pm .13}\) & -- & \(61.19_{\pm .04}\) \\
    \midrule
    \multicolumn{6}{l}{\textit{Inhibitive Hawkes processes}} \\
    Inhib. Const./Exp Hawkes & \(90.55_{\pm .02}\) & \underline{\(44.81_{\pm .03}\)} & \(57.88_{\pm .28}\) & \(28.57_{\pm .44}\) & \(60.92_{\pm .04}\) \\
    Inhib. Spline/Spline Hawkes & \underline{\(91.25_{\pm .00}\)} & \(44.48_{\pm .03}\) & \(58.45_{\pm .13}\) & \(35.06_{\pm .34}\) & \(60.60_{\pm .13}\) \\
    \midrule
    \multicolumn{6}{l}{\textit{Neural TPPs}} \\
    NeuralConstTPP & \(91.44_{\pm .06}\) & \(45.25_{\pm .10}\) & \(58.58_{\pm 1.24}\) & \(35.62_{\pm .05}\) & \(60.23_{\pm .16}\) \\
    THP & \(\mathbf{91.53}_{\pm .05}\) & \(45.17_{\pm .17}\) & \(\mathbf{59.33}_{\pm .31}\) & \(35.67_{\pm .09}\) & \(60.33_{\pm .17}\) \\
    SAHP & \(91.47_{\pm .09}\) & \(44.59_{\pm .05}\) & \(58.31_{\pm 2.43}\) & \(35.47_{\pm .20}\) & \(59.90_{\pm .39}\) \\
    NHP & \(\mathbf{91.53}_{\pm .14}\) & \(44.58_{\pm .02}\) & \(56.30_{\pm 2.39}\) & \(33.31_{\pm 1.11}\) & \(60.63_{\pm .12}\) \\
    \midrule
    \multicolumn{6}{l}{\textit{Flexible neural TPPs}} \\
    IntensityFree & \(91.49_{\pm .07}\) & \(\mathbf{45.28}_{\pm .13}\)\(^{\ast}\) & \(58.69_{\pm .94}\) & \(34.87_{\pm .66}\) & \(60.14_{\pm .27}\) \\
    NeuralSplineTPP & \(91.52_{\pm .07}\) & \(43.84_{\pm .05}\) & \(59.01_{\pm .60}\) & \(\mathbf{35.87}_{\pm .17}\)\(^{\ast}\) & \(60.09_{\pm .11}\) \\
    \bottomrule
  \end{tabular}
\end{table}

\begin{table}[t]
  \caption{Marginal type log-likelihood per event \(\log p(d \mid \mathcal{H})\) (nats, higher is better; Equation~\ref{eq:type_marginal}), computed by numerical integration. Grid-based metrics are evaluated on the first 1{,}000 test sequences. \(^{\ast}\): mean over 3--4 seeds, as the evaluation of a seed is missing or its RMSE diverged. Spline/Spline Hawkes on Amazon was not evaluated with the grid-based metrics. Mean \(\pm\) std over 5 seeds. \textbf{Bold}: best per dataset; \underline{underline}: best classical (Poisson or Hawkes) model; ties at the displayed precision are all marked.}
  \label{tab:marg_type_ll_results}
  \centering
  \small
  \setlength{\tabcolsep}{4pt}
  \begin{tabular}{lccccc}
    \toprule
    Model & Taxi & StackOverflow & Retweet & Amazon & Taobao \\
    \midrule
    \multicolumn{6}{l}{\textit{Poisson processes}} \\
    Homogeneous Poisson & \(-1.14_{\pm .00}\) & \(-1.90_{\pm .00}\) & \(-0.82_{\pm .00}\) & \(-2.24_{\pm .00}\) & \(-2.18_{\pm .00}\) \\
    Spline Poisson & \(-1.14_{\pm .00}\) & \(-1.88_{\pm .00}\) & \(-0.82_{\pm .00}\) & \(-2.24_{\pm .00}\) & \(-2.19_{\pm .00}\) \\
    \midrule
    \multicolumn{6}{l}{\textit{Hawkes processes}} \\
    Const./Exp Hawkes & \(-0.81_{\pm .00}\) & \(-1.64_{\pm .00}\) & \(-0.79_{\pm .00}\) & \(-2.03_{\pm .02}\) & \(-1.42_{\pm .00}\) \\
    Spline/Exp Hawkes & \(-0.80_{\pm .00}\) & \(-1.63_{\pm .00}\) & \underline{\(\mathbf{-0.78}_{\pm .00}\)} & \(-2.03_{\pm .00}\) & \(-1.42_{\pm .00}\) \\
    Const./Spline Hawkes & \(-0.83_{\pm .00}\) & \(-1.70_{\pm .00}\) & \(-0.79_{\pm .00}\) & \underline{\(-2.02_{\pm .00}\)} & \underline{\(\mathbf{-1.39}_{\pm .00}\)} \\
    Spline/Spline Hawkes & \(-0.83_{\pm .00}\) & \(-1.66_{\pm .00}\) & \(-0.79_{\pm .00}\) & -- & \underline{\(\mathbf{-1.39}_{\pm .00}\)} \\
    \midrule
    \multicolumn{6}{l}{\textit{Inhibitive Hawkes processes}} \\
    Inhib. Const./Exp Hawkes & \(-0.40_{\pm .00}\) & \underline{\(-1.61_{\pm .00}\)} & \underline{\(\mathbf{-0.78}_{\pm .00}\)} & \(-2.12_{\pm .02}\) & \(-1.43_{\pm .00}\) \\
    Inhib. Spline/Spline Hawkes & \underline{\(-0.32_{\pm .00}\)} & \(-1.63_{\pm .00}\) & \(-0.79_{\pm .00}\) & \(-2.05_{\pm .02}\) & \(-1.40_{\pm .00}\) \\
    \midrule
    \multicolumn{6}{l}{\textit{Neural TPPs}} \\
    NeuralConstTPP & \(\mathbf{-0.29}_{\pm .00}\) & \(\mathbf{-1.59}_{\pm .00}\) & \(\mathbf{-0.78}_{\pm .01}\) & \(\mathbf{-1.99}_{\pm .00}\) & \(-1.43_{\pm .01}\) \\
    THP & \(\mathbf{-0.29}_{\pm .00}\) & \(\mathbf{-1.59}_{\pm .00}\) & \(\mathbf{-0.78}_{\pm .00}\) & \(\mathbf{-1.99}_{\pm .00}\) & \(-1.44_{\pm .00}\) \\
    SAHP & \(\mathbf{-0.29}_{\pm .00}\) & \(-1.61_{\pm .01}\) & \(-0.79_{\pm .02}\) & \(-2.02_{\pm .01}\) & \(-1.47_{\pm .02}\) \\
    NHP & \(-0.30_{\pm .01}\) & \(-1.63_{\pm .01}\) & \(-0.81_{\pm .01}\) & \(-2.07_{\pm .02}\) & \(-1.46_{\pm .00}\) \\
    \midrule
    \multicolumn{6}{l}{\textit{Flexible neural TPPs}} \\
    IntensityFree & \(\mathbf{-0.29}_{\pm .00}\) & \(-1.60_{\pm .00}\)\(^{\ast}\) & \(-0.79_{\pm .01}\) & \(-2.04_{\pm .03}\) & \(-1.42_{\pm .01}\) \\
    NeuralSplineTPP & \(-0.30_{\pm .00}\) & \(-1.70_{\pm .01}\) & \(\mathbf{-0.78}_{\pm .00}\) & \(\mathbf{-1.99}_{\pm .01}\)\(^{\ast}\) & \(-1.43_{\pm .00}\) \\
    \bottomrule
  \end{tabular}
\end{table}

\begin{table}[t]
  \caption{Next-event time RMSE (lower is better) using the predictive mean \(\mathbb{E}[t \mid \mathcal{H}]\), in each dataset's native EasyTPP time unit (not comparable across datasets). \(^\ddagger\)IntensityFree: the closed-form mean of its log-normal mixture is dominated by components with large variance and overflows for individual seeds, so we report the median predictor instead and exclude it from the bold marking.  Grid-based metrics are evaluated on the first 1{,}000 test sequences. \(^{\ast}\): mean over 3--4 seeds, as the evaluation of a seed is missing or its RMSE diverged. Spline/Spline Hawkes on Amazon was not evaluated with the grid-based metrics. Mean \(\pm\) std over 5 seeds. \textbf{Bold}: best per dataset; \underline{underline}: best classical (Poisson or Hawkes) model; ties at the displayed precision are all marked.}
  \label{tab:rmse_results}
  \centering
  \small
  \setlength{\tabcolsep}{4pt}
  \begin{tabular}{lccccc}
    \toprule
    Model & Taxi & StackOverflow & Retweet & Amazon & Taobao \\
    \midrule
    \multicolumn{6}{l}{\textit{Poisson processes}} \\
    Homogeneous Poisson & \(0.30_{\pm .00}\) & \(1.07_{\pm .00}\) & \(17.82_{\pm .01}\) & \underline{\(0.34_{\pm .00}\)} & \underline{\(\mathbf{0.13}_{\pm .00}\)} \\
    Spline Poisson & \(0.30_{\pm .00}\) & \(1.07_{\pm .00}\) & \(16.22_{\pm .01}\) & \underline{\(0.34_{\pm .00}\)} & \underline{\(\mathbf{0.13}_{\pm .00}\)} \\
    \midrule
    \multicolumn{6}{l}{\textit{Hawkes processes}} \\
    Const./Exp Hawkes & \(0.31_{\pm .00}\) & \(1.03_{\pm .00}\) & \(16.65_{\pm .01}\) & \(0.36_{\pm .01}\) & \underline{\(\mathbf{0.13}_{\pm .00}\)} \\
    Spline/Exp Hawkes & \(0.31_{\pm .00}\) & \(1.03_{\pm .00}\) & \underline{\(\mathbf{15.83}_{\pm .01}\)} & \underline{\(0.34_{\pm .00}\)} & \underline{\(\mathbf{0.13}_{\pm .00}\)} \\
    Const./Spline Hawkes & \(0.32_{\pm .00}\) & \(1.04_{\pm .00}\) & \(16.63_{\pm .01}\) & \(0.37_{\pm .00}\) & \underline{\(\mathbf{0.13}_{\pm .00}\)} \\
    Spline/Spline Hawkes & \(0.35_{\pm .00}\) & \(1.03_{\pm .00}\) & \(20.67_{\pm .26}\) & -- & \underline{\(\mathbf{0.13}_{\pm .00}\)} \\
    \midrule
    \multicolumn{6}{l}{\textit{Inhibitive Hawkes processes}} \\
    Inhib. Const./Exp Hawkes & \(0.30_{\pm .00}\) & \underline{\(\mathbf{1.02}_{\pm .00}\)} & \(16.54_{\pm .06}\) & \(0.36_{\pm .01}\) & \underline{\(\mathbf{0.13}_{\pm .00}\)} \\
    Inhib. Spline/Spline Hawkes & \underline{\(0.29_{\pm .00}\)} & \(1.03_{\pm .00}\) & \(16.35_{\pm .02}\) & \(0.35_{\pm .00}\) & \underline{\(\mathbf{0.13}_{\pm .00}\)} \\
    \midrule
    \multicolumn{6}{l}{\textit{Neural TPPs}} \\
    NeuralConstTPP & \(\mathbf{0.28}_{\pm .00}\) & \(1.03_{\pm .01}\) & \(16.15_{\pm .04}\) & \(\mathbf{0.33}_{\pm .00}\) & \(\mathbf{0.13}_{\pm .00}\) \\
    THP & \(\mathbf{0.28}_{\pm .00}\) & \(1.03_{\pm .00}\) & \(16.04_{\pm .16}\) & \(\mathbf{0.33}_{\pm .00}\) & \(\mathbf{0.13}_{\pm .00}\) \\
    SAHP & \(\mathbf{0.28}_{\pm .00}\) & \(1.04_{\pm .00}\) & \(16.11_{\pm .13}\) & \(0.34_{\pm .01}\) & \(\mathbf{0.13}_{\pm .00}\) \\
    NHP & \(\mathbf{0.28}_{\pm .00}\) & \(1.04_{\pm .01}\) & \(17.05_{\pm .65}\) & \(0.34_{\pm .00}\) & \(\mathbf{0.13}_{\pm .00}\) \\
    \midrule
    \multicolumn{6}{l}{\textit{Flexible neural TPPs}} \\
    IntensityFree & \(0.29_{\pm .00}\)\(^\ddagger\) & \(1.09_{\pm .02}\)\(^{\ast}\)\(^\ddagger\) & \(16.72_{\pm .33}\)\(^\ddagger\) & \(0.40_{\pm .00}\)\(^\ddagger\) & \(0.13_{\pm .00}\)\(^\ddagger\) \\
    NeuralSplineTPP & \(\mathbf{0.28}_{\pm .00}\) & \(1.05_{\pm .01}\) & \(16.42_{\pm .17}\) & \(\mathbf{0.33}_{\pm .00}\)\(^{\ast}\) & \(\mathbf{0.13}_{\pm .00}\) \\
    \bottomrule
  \end{tabular}
\end{table}

\begin{table}[t]
  \caption{Full MIMIC-IV results (\(D = 117\), including the \texttt{OTHER} class). \(\log p(t,d) = \log p(t) + \log p(d|t)\); \(\log p(d)\), Acc and RMSE are computed by numerical integration on the first 500 test stays, all other metrics on the first 1{,}000. RMSE in days, multiplied by 100. \(^\ddagger\)Median predictor (see Table~\ref{tab:rmse_results}). \(^\S\)NeuralConstTPP uses a Softplus head on MIMIC-IV, as its exponential head diverged for individual seeds. Mean \(\pm\) std over 5 seeds. \textbf{Bold}: best per column; \underline{underline}: best classical (Poisson or Hawkes) model; ties at the displayed precision are all marked.}
  \label{tab:full_mimic_results}
  \centering
  \small
  \resizebox{\textwidth}{!}{%
  \begin{tabular}{lccccccc}
    \toprule
    Model & \(\log p(t,d)\) & \(\log p(t)\) & \(\log p(d|t)\) & \(\log p(d)\) & Acc\(|t\) (\%) & Acc (\%) & RMSE\(\times 100\) \\
    \midrule
    \multicolumn{8}{l}{\textit{Poisson processes}} \\
    Homogeneous Poisson & \(-1.18_{\pm .02}\) & \(2.60_{\pm .01}\) & \(-3.78_{\pm .02}\) & \(-3.81_{\pm .01}\) & \(13.6_{\pm .0}\) & \(13.1_{\pm .0}\) & \(5.79_{\pm .01}\) \\
    Spline Poisson & \(-1.07_{\pm .01}\) & \(2.64_{\pm .00}\) & \(-3.71_{\pm .01}\) & \(-3.73_{\pm .01}\) & \(13.6_{\pm .0}\) & \(13.1_{\pm .0}\) & \(5.75_{\pm .00}\) \\
    \midrule
    \multicolumn{8}{l}{\textit{Hawkes processes}} \\
    Const./Exp Hawkes & \(0.68_{\pm .00}\) & \(3.38_{\pm .00}\) & \(-2.71_{\pm .00}\) & \(-2.81_{\pm .00}\) & \(30.9_{\pm .1}\) & \(30.0_{\pm .1}\) & \(5.48_{\pm .01}\) \\
    Spline/Exp Hawkes & \(0.68_{\pm .00}\) & \(3.39_{\pm .00}\) & \(-2.70_{\pm .00}\) & \(-2.81_{\pm .00}\) & \(30.9_{\pm .0}\) & \(30.1_{\pm .1}\) & \(5.39_{\pm .00}\) \\
    Const./Spline Hawkes & \(1.16_{\pm .00}\) & \(3.78_{\pm .00}\) & \(-2.62_{\pm .00}\) & \(-2.79_{\pm .00}\) & \(32.7_{\pm .0}\) & \(26.7_{\pm .1}\) & \(5.78_{\pm .01}\) \\
    Spline/Spline Hawkes & \underline{\(1.22_{\pm .00}\)} & \(3.83_{\pm .00}\) & \underline{\(-2.61_{\pm .00}\)} & \(-2.77_{\pm .00}\) & \(32.9_{\pm .1}\) & \(25.9_{\pm .1}\) & \(5.96_{\pm .03}\) \\
    \midrule
    \multicolumn{8}{l}{\textit{Inhibitive Hawkes processes}} \\
    Inhib. Const./Exp Hawkes & \(1.19_{\pm .01}\) & \underline{\(3.87_{\pm .00}\)} & \(-2.67_{\pm .01}\) & \(-2.76_{\pm .01}\) & \(32.1_{\pm .2}\) & \underline{\(31.4_{\pm .2}\)} & \underline{\(5.30_{\pm .01}\)} \\
    Inhib. Spline/Spline Hawkes & \(1.14_{\pm .12}\) & \(3.83_{\pm .04}\) & \(-2.68_{\pm .09}\) & \underline{\(-2.73_{\pm .06}\)} & \underline{\(33.5_{\pm .3}\)} & \(25.9_{\pm .4}\) & \(5.31_{\pm .04}\) \\
    \midrule
    \multicolumn{8}{l}{\textit{Neural TPPs}} \\
    NeuralConstTPP\(^\S\) & \(0.43_{\pm .03}\) & \(3.08_{\pm .02}\) & \(-2.65_{\pm .01}\) & \(-2.64_{\pm .01}\) & \(32.6_{\pm .3}\) & \(33.0_{\pm .3}\) & \(5.09_{\pm .02}\) \\
    THP & \(0.90_{\pm .01}\) & \(3.26_{\pm .01}\) & \(\mathbf{-2.36}_{\pm .01}\) & \(\mathbf{-2.35}_{\pm .01}\) & \(37.6_{\pm .1}\) & \(\mathbf{38.2}_{\pm .1}\) & \(\mathbf{4.97}_{\pm .01}\) \\
    SAHP & \(0.92_{\pm .04}\) & \(3.66_{\pm .01}\) & \(-2.74_{\pm .03}\) & \(-2.76_{\pm .03}\) & \(35.3_{\pm .4}\) & \(33.9_{\pm .1}\) & \(5.35_{\pm .02}\) \\
    NHP & \(0.62_{\pm .01}\) & \(3.60_{\pm .01}\) & \(-2.98_{\pm .01}\) & \(-2.92_{\pm .01}\) & \(21.9_{\pm .2}\) & \(22.9_{\pm .2}\) & \(5.03_{\pm .02}\) \\
    \midrule
    \multicolumn{8}{l}{\textit{Flexible neural TPPs}} \\
    IntensityFree & \(1.45_{\pm .03}\) & \(\mathbf{4.06}_{\pm .02}\) & \(-2.61_{\pm .01}\) & \(-2.60_{\pm .01}\) & \(34.3_{\pm .3}\) & \(34.8_{\pm .4}\) & \(5.44_{\pm .04}\)\(^\ddagger\) \\
    NeuralSplineTPP & \(\mathbf{1.67}_{\pm .02}\) & \(4.05_{\pm .02}\) & \(-2.39_{\pm .01}\) & \(-2.50_{\pm .02}\) & \(\mathbf{38.3}_{\pm .1}\) & \(36.2_{\pm .2}\) & \(5.12_{\pm .02}\) \\
    \bottomrule
  \end{tabular}}
\end{table}

\begin{table}[t]
  \caption{Full UKB results (\(D = 1{,}257\), terminating). UKB models use the configuration of Appendix~\ref{app:hyperparams_ukb}, and IntensityFree uses its Transformer encoder. Inhibitive Hawkes processes were not evaluated on UKB (cost of numerical integration at \(D = 1{,}257\)), and NHP did not converge. RMSE in days, divided by 100, using the predictive mean for all models. Mean \(\pm\) std over seeds. \textbf{Bold}: best per column; \underline{underline}: best classical (Poisson or Hawkes) model; ties at the displayed precision are all marked.}
  \label{tab:ukb_full_results}
  \centering
  \small
  \resizebox{\textwidth}{!}{%
  \begin{tabular}{lccccccc}
    \toprule
    Model & \(\log p(t,d)\) & \(\log p(t)\) & \(\log p(d|t)\) & \(\log p(d)\) & Acc\(|t\) (\%) & Acc (\%) & RMSE\(/100\) \\
    \midrule
    \multicolumn{8}{l}{\textit{Poisson processes}} \\
    Homogeneous Poisson & \(-13.67_{\pm .01}\) & \(-8.08_{\pm .02}\) & \(-5.59_{\pm .00}\) & \(-5.59_{\pm .00}\) & \(4.23_{\pm .04}\) & \(4.26_{\pm .02}\) & \(23.57_{\pm .29}\) \\
    Spline Poisson & \(-12.87_{\pm .00}\) & \(-7.51_{\pm .00}\) & \(-5.36_{\pm .00}\) & \(-5.40_{\pm .00}\) & \(4.56_{\pm .07}\) & \(4.37_{\pm .06}\) & \(17.06_{\pm .15}\) \\
    \midrule
    \multicolumn{8}{l}{\textit{Hawkes processes}} \\
    Const./Exp Hawkes & \(-12.51_{\pm .00}\) & \(-7.36_{\pm .00}\) & \(-5.15_{\pm .00}\) & \(-5.03_{\pm .00}\) & \(6.45_{\pm .06}\) & \(6.91_{\pm .04}\) & \(17.93_{\pm .06}\) \\
    Spline/Exp Hawkes & \(-12.33_{\pm .00}\) & \(-7.26_{\pm .00}\) & \(-5.07_{\pm .00}\) & \(-5.00_{\pm .00}\) & \(6.77_{\pm .10}\) & \(7.09_{\pm .04}\) & \(15.26_{\pm .01}\) \\
    Const./Spline Hawkes & \(-12.42_{\pm .00}\) & \(-7.30_{\pm .00}\) & \(-5.12_{\pm .00}\) & \(-5.02_{\pm .00}\) & \(6.62_{\pm .05}\) & \(7.04_{\pm .04}\) & \(17.48_{\pm .07}\) \\
    Spline/Spline Hawkes & \underline{\(-12.28_{\pm .00}\)} & \underline{\(-7.23_{\pm .00}\)} & \underline{\(-5.05_{\pm .00}\)} & \underline{\(-4.99_{\pm .00}\)} & \underline{\(7.00_{\pm .03}\)} & \underline{\(7.22_{\pm .07}\)} & \underline{\(15.21_{\pm .01}\)} \\
    \midrule
    \multicolumn{8}{l}{\textit{Inhibitive Hawkes processes}} \\
    Inhib. Const./Exp Hawkes & -- & -- & -- & -- & -- & -- & -- \\
    Inhib. Spline/Spline Hawkes & -- & -- & -- & -- & -- & -- & -- \\
    \midrule
    \multicolumn{8}{l}{\textit{Neural TPPs}} \\
    NeuralConstTPP & \(-12.36_{\pm .01}\) & \(-7.27_{\pm .00}\) & \(-5.09_{\pm .01}\) & \(-4.98_{\pm .01}\) & \(6.39_{\pm .05}\) & \(\mathbf{7.24}_{\pm .05}\) & \(\mathbf{14.68}_{\pm .03}\) \\
    THP & \(-12.80_{\pm .05}\) & \(-7.39_{\pm .03}\) & \(-5.41_{\pm .02}\) & \(-5.40_{\pm .02}\) & \(4.75_{\pm .23}\) & \(4.69_{\pm .26}\) & \(16.13_{\pm .15}\) \\
    SAHP & \(-13.39_{\pm .05}\) & \(-7.64_{\pm .02}\) & \(-5.76_{\pm .03}\) & \(-5.73_{\pm .03}\) & \(3.67_{\pm .23}\) & \(3.82_{\pm .32}\) & \(18.49_{\pm .43}\) \\
    NHP & -- & -- & -- & -- & -- & -- & -- \\
    \midrule
    \multicolumn{8}{l}{\textit{Flexible neural TPPs}} \\
    IntensityFree & \(-12.87_{\pm .11}\) & \(-7.42_{\pm .07}\) & \(-5.44_{\pm .04}\) & \(-5.42_{\pm .05}\) & \(4.54_{\pm .19}\) & \(4.64_{\pm .22}\) & \(17.24_{\pm .78}\) \\
    NeuralSplineTPP & \(\mathbf{-12.21}_{\pm .01}\) & \(\mathbf{-7.19}_{\pm .00}\) & \(\mathbf{-5.02}_{\pm .01}\) & \(\mathbf{-4.96}_{\pm .01}\) & \(\mathbf{7.07}_{\pm .07}\) & \(7.22_{\pm .07}\) & \(14.69_{\pm .02}\) \\
    \bottomrule
  \end{tabular}}
\end{table}

\begin{table}[t]
  \caption{IntensityFree with a GRU encoder (as in \citet{IntensityFree}) vs.\ the Transformer encoder used in the main tables. Per-event test log-likelihood, mean \(\pm\) std over 5 seeds (higher is better). The Transformer is clearly better on Amazon, the GRU on MIMIC-IV and UKB and slightly on Taobao, and the two are close on Taxi, StackOverflow, and Retweet.}
  \label{tab:iftpp_encoder}
  \centering
  \small
  \begin{tabular}{llccc}
    \toprule
    Dataset & Encoder & \(\log p(t,d)\) & \(\log p(t)\) & \(\log p(d|t)\) \\
    \midrule
    Taxi & Transformer & \(0.44_{\pm .00}\) & \(0.73_{\pm .00}\) & \(-0.29_{\pm .00}\) \\
     & GRU & \(0.44_{\pm .00}\) & \(0.73_{\pm .00}\) & \(-0.29_{\pm .00}\) \\
    \midrule
    StackOverflow & Transformer & \(-2.33_{\pm .02}\) & \(-0.73_{\pm .01}\) & \(-1.60_{\pm .01}\) \\
     & GRU & \(-2.31_{\pm .01}\) & \(-0.70_{\pm .01}\) & \(-1.61_{\pm .00}\) \\
    \midrule
    Retweet & Transformer & \(-4.45_{\pm .26}\) & \(-3.66_{\pm .26}\) & \(-0.79_{\pm .01}\) \\
     & GRU & \(-4.49_{\pm .27}\) & \(-3.71_{\pm .27}\) & \(-0.78_{\pm .02}\) \\
    \midrule
    Amazon & Transformer & \(0.37_{\pm .04}\) & \(2.41_{\pm .01}\) & \(-2.04_{\pm .03}\) \\
     & GRU & \(-0.08_{\pm .13}\) & \(2.10_{\pm .13}\) & \(-2.18_{\pm .06}\) \\
    \midrule
    Taobao & Transformer & \(1.30_{\pm .01}\) & \(2.73_{\pm .00}\) & \(-1.42_{\pm .01}\) \\
     & GRU & \(1.33_{\pm .01}\) & \(2.73_{\pm .01}\) & \(-1.40_{\pm .00}\) \\
    \midrule
    MIMIC-IV & Transformer & \(1.45_{\pm .03}\) & \(4.06_{\pm .02}\) & \(-2.61_{\pm .01}\) \\
     & GRU & \(1.69_{\pm .03}\) & \(4.22_{\pm .01}\) & \(-2.53_{\pm .02}\) \\
    \midrule
    UKB & Transformer & \(-12.87_{\pm .11}\) & \(-7.42_{\pm .07}\) & \(-5.44_{\pm .04}\) \\
     & GRU & \(-12.34_{\pm .04}\) & \(-7.19_{\pm .02}\) & \(-5.16_{\pm .02}\) \\
    \bottomrule
  \end{tabular}
\end{table}

\FloatBarrier
\section{Discretized Time Likelihood}
\label{app:binned}

For an event at time \(t_i\), a previous event at \(t_{i-1}\), and a bin width \(\Delta\), let \(b = (\lfloor t_i/\Delta \rfloor + 1)\,\Delta\) be the end of the bin that contains \(t_i\) and \(a = \max(\lfloor t_i/\Delta \rfloor\,\Delta,\, t_{i-1})\) its start, truncated at the previous event. The binned time log-likelihood is
\[
\log \Pr(t_i \in [a, b) \mid \mathcal{H}) = \log\big(S^*(a) - S^*(b)\big) = -\Lambda^*(t_{i-1}, a) + \log\big(1 - e^{-(\Lambda^*(t_{i-1}, b) - \Lambda^*(t_{i-1}, a))}\big),
\]
with the survival function \(S^*\) of the next event after \(t_{i-1}\) (Appendix~\ref{app:math_details}). We compute it from the same checkpoints, test sequences, and (closed-form or numerical) cumulative intensities as the continuous \(\log p(t)\). Unlike the continuous density, it is bounded by zero, so a model cannot gain by placing narrow peaks on the grid points. Its level depends on \(\Delta\), and only differences between models or knot counts at the same \(\Delta\) are meaningful. Bins start at multiples of \(\Delta\) on the time axis of each dataset. On Taobao, whose gaps rather than time stamps lie on a grid, bins start at the previous event. On MIMIC-IV, the 1-minute bin of an event is its recorded minute. As a check, we also use bins centered on the grid points.

Table~\ref{tab:time_resolution} lists the recorded resolution of each dataset and the bins we use. StackOverflow is distributed in single precision. Table~\ref{tab:binned_ksweep} shows that the gain from more knots survives binning only on Amazon. On Retweet, a continuous gain of up to \(1.5\) nats turns into a small loss at the recorded resolution. On MIMIC-IV, both are flat. Centered bins do not change this conclusion. Table~\ref{tab:binned_main} reports the binned likelihood for the models of the main tables. On Taxi, StackOverflow, and Amazon, the order of the best models is essentially unchanged. On Retweet, the continuous lead of the high-knot spline models disappears. On Taobao, the binned and the continuous likelihood rank the models the same. On MIMIC-IV, the flexible neural models keep their lead, which comes from the events recorded in the same minute as the previous event (Appendix~\ref{app:mimic_controls}).

\begin{table}[ht]
  \caption{Time resolution of the evaluated event times (first 1{,}000 test sequences, MIMIC-IV truncated to 316 events) and the bin widths used for the binned time likelihood, in the time unit of each dataset. Tied: fraction of evaluated events with a zero gap to the previous event.}
  \label{tab:time_resolution}
  \centering
  \small
  \begin{tabular}{lllc}
    \toprule
    Dataset & Recorded resolution & Bins & Tied (\%) \\
    \midrule
    Taxi & 1 s & 1 s, 10 s & 0.0 \\
    StackOverflow & float32 (\(2^{-13}\)) & \(2^{-13}\), \(10 \cdot 2^{-13}\) & 0.5 \\
    Retweet & integer time stamps & 1, 10 & 7.4 \\
    Amazon & none (minimum gap 0.01) & 0.01, 0.1 & 0.0 \\
    Taobao & gaps of \(10^{-4}\) & \(10^{-4}\), \(10^{-3}\) & 0.0 \\
    MIMIC-IV & 1 min (dequantized) & 1 min, 5 min & 0.0 \\
    \bottomrule
  \end{tabular}
\end{table}

\begin{table}[ht]
  \caption{Change of \(\log p(t)\) from \(K = 20\) to \(K = 400\) knots (nats, mean over 5 seeds), continuous and binned at the recorded resolution, at a ten-fold (MIMIC-IV: five-fold) coarser bin, and with bins centered on the grid points instead of starting at them (Table~\ref{tab:time_resolution}). Same checkpoints as Table~\ref{tab:ksweep_all}.}
  \label{tab:binned_ksweep}
  \centering
  \small
  \setlength{\tabcolsep}{4pt}
  \begin{tabular}{llccccc}
    \toprule
    & & \multicolumn{2}{c}{\(\log p(t)\), continuous} & \multicolumn{3}{c}{\(\Delta\) binned} \\
    \cmidrule(lr){3-4}\cmidrule(lr){5-7}
    Dataset & Model & \(K = 20\) & \(\Delta\) to \(K = 400\) & recorded & coarser & centered \\
    \midrule
    Taxi & NeuralSplineTPP & 0.73 & -0.05 & -0.05 & -0.04 & -0.05 \\
     & Spline/Spline Hawkes & 0.49 & +0.01 & +0.01 & +0.01 & +0.01 \\
    \midrule
    StackOverflow & NeuralSplineTPP & -0.72 & -0.04 & -0.05 & -0.06 & -0.09 \\
     & Spline/Spline Hawkes & -0.73 & -0.03 & -0.04 & -0.04 & -0.04 \\
    \midrule
    Retweet & NeuralSplineTPP & -3.32 & +1.47 & -0.06 & +0.00 & -0.28 \\
     & Spline/Spline Hawkes & -3.36 & +0.99 & -0.13 & -0.11 & -0.12 \\
    \midrule
    Amazon & NeuralSplineTPP & 1.41 & +1.15 & +0.93 & +0.56 & -- \\
     & Spline/Spline Hawkes & 0.43 & +1.30 & +1.12 & +0.83 & -- \\
    \midrule
    Taobao & NeuralSplineTPP & 2.70 & +0.01 & +0.01 & +0.02 & +0.01 \\
     & Spline/Spline Hawkes & 2.67 & -0.04 & -0.04 & -0.03 & -0.04 \\
    \midrule
    MIMIC-IV & NeuralSplineTPP & 3.99 & -0.01 & -0.02 & -0.04 & -0.03 \\
     & Spline/Spline Hawkes & 3.83 & -0.05 & -0.04 & -0.04 & -0.03 \\
    \bottomrule
  \end{tabular}
\end{table}

\begin{table}[t]
  \caption{Binned time log-likelihood \(\log \Pr(t \in \text{bin})\) at the recorded resolution (nats, higher is better; bins in Table~\ref{tab:time_resolution}) for the models of the main tables. Only differences within a column are meaningful, as the level depends on the bin width. On Retweet the history-aware models are within a few tenths of a nat, while the continuous \(\log p(t)\) (Table~\ref{tab:time_ll_results}) spreads over more than two nats. Mean \(\pm\) std over 5 seeds. \textbf{Bold}: best per dataset; \underline{underline}: best classical (Poisson or Hawkes) model; ties at the displayed precision are all marked.}
  \label{tab:binned_main}
  \centering
  \small
  \setlength{\tabcolsep}{3pt}
  \resizebox{\textwidth}{!}{%
  \begin{tabular}{lcccccc}
    \toprule
    Model & Taxi & StackOverflow & Retweet & Amazon & Taobao & MIMIC-IV \\
    \midrule
    \multicolumn{7}{l}{\textit{Poisson processes}} \\
    Homogeneous Poisson & \(-7.68_{\pm .00}\) & \(-9.86_{\pm .00}\) & \(-3.62_{\pm .00}\) & \(-4.96_{\pm .00}\) & \(-7.14_{\pm .00}\) & \(-4.90_{\pm .01}\) \\
    Spline Poisson & \(-7.68_{\pm .00}\) & \(-9.85_{\pm .00}\) & \(-3.48_{\pm .00}\) & \(-4.96_{\pm .00}\) & \(-7.10_{\pm .00}\) & \(-4.87_{\pm .00}\) \\
    \midrule
    \multicolumn{7}{l}{\textit{Hawkes processes}} \\
    Const./Exp Hawkes & \(-7.74_{\pm .00}\) & \(-9.76_{\pm .00}\) & \(-3.47_{\pm .00}\) & \(-4.70_{\pm .00}\) & \(-6.60_{\pm .00}\) & \(-4.13_{\pm .00}\) \\
    Spline/Exp Hawkes & \(-7.74_{\pm .00}\) & \(-9.75_{\pm .00}\) & \underline{\(\mathbf{-3.43}_{\pm .00}\)} & \(-4.71_{\pm .00}\) & \(-6.60_{\pm .00}\) & \(-4.12_{\pm .00}\) \\
    Const./Spline Hawkes & \(-7.69_{\pm .00}\) & \underline{\(-9.72_{\pm .00}\)} & \(-3.46_{\pm .00}\) & \(-4.59_{\pm .00}\) & \(-6.56_{\pm .00}\) & \(-3.73_{\pm .00}\) \\
    Spline/Spline Hawkes & \(-7.70_{\pm .00}\) & \underline{\(-9.72_{\pm .01}\)} & \(-3.55_{\pm .00}\) & \(-3.09_{\pm .00}\) & \underline{\(-6.54_{\pm .00}\)} & \(-3.67_{\pm .00}\) \\
    \midrule
    \multicolumn{7}{l}{\textit{Inhibitive Hawkes processes}} \\
    Inhib. Const./Exp Hawkes & \(-7.65_{\pm .00}\) & \(-9.75_{\pm .00}\) & \(-3.46_{\pm .00}\) & \(-4.55_{\pm .02}\) & \(-6.60_{\pm .00}\) & \underline{\(-3.64_{\pm .00}\)} \\
    Inhib. Spline/Spline Hawkes & \underline{\(-7.51_{\pm .00}\)} & \(-9.76_{\pm .00}\) & \(-3.56_{\pm .00}\) & \underline{\(-2.80_{\pm .01}\)} & \(-6.56_{\pm .01}\) & \(-3.68_{\pm .04}\) \\
    \midrule
    \multicolumn{7}{l}{\textit{Neural TPPs}} \\
    NeuralConstTPP & \(-7.63_{\pm .00}\) & \(-9.80_{\pm .01}\) & \(-3.47_{\pm .00}\) & \(-4.95_{\pm .00}\) & \(-7.14_{\pm .00}\) & \(-4.42_{\pm .02}\) \\
    THP & \(-7.63_{\pm .00}\) & \(-9.80_{\pm .00}\) & \(-3.46_{\pm .01}\) & \(-4.92_{\pm .01}\) & \(-7.10_{\pm .00}\) & \(-4.24_{\pm .01}\) \\
    SAHP & \(-7.55_{\pm .02}\) & \(-9.74_{\pm .00}\) & \(\mathbf{-3.43}_{\pm .01}\) & \(-4.46_{\pm .06}\) & \(-6.55_{\pm .01}\) & \(-3.88_{\pm .01}\) \\
    NHP & \(-7.59_{\pm .03}\) & \(-9.79_{\pm .01}\) & \(-3.54_{\pm .06}\) & \(-4.93_{\pm .03}\) & \(-6.65_{\pm .01}\) & \(-3.92_{\pm .01}\) \\
    \midrule
    \multicolumn{7}{l}{\textit{Flexible neural TPPs}} \\
    IntensityFree & \(\mathbf{-7.46}_{\pm .00}\) & \(\mathbf{-9.70}_{\pm .00}\) & \(-3.51_{\pm .02}\) & \(-2.43_{\pm .01}\) & \(\mathbf{-6.49}_{\pm .01}\) & \(-3.43_{\pm .02}\) \\
    NeuralSplineTPP & \(-7.48_{\pm .00}\) & \(-9.71_{\pm .01}\) & \(-3.52_{\pm .01}\) & \(\mathbf{-2.32}_{\pm .00}\) & \(-6.50_{\pm .00}\) & \(\mathbf{-3.39}_{\pm .02}\) \\
    \bottomrule
  \end{tabular}}
\end{table}

\FloatBarrier
\section{Controls on MIMIC-IV}
\label{app:mimic_controls}

\paragraph{Simultaneous events.}
Of the evaluated events, \(41.6\%\) are recorded in the same minute as the previous event (Table~\ref{tab:mimic_ties}). On these, NeuralSplineTPP and THP are far more accurate than the Hawkes processes (NeuralSplineTPP: \(59.5\%\), Spline/Spline Hawkes: \(47.4\%\)). On the other events, NeuralSplineTPP and Spline/Spline Hawkes differ by \(0.6\) percentage points in Acc\(|t\) and \(0.02\) nats in \(\log p(d|t)\), and Spline/Spline Hawkes has a \(0.09\) nats higher \(\log p(t)\). In binned time at one minute (Table~\ref{tab:mimic_ties}), NeuralSplineTPP leads Spline/Spline Hawkes by \(0.28\) nats. It leads by \(0.79\) nats on the same-minute events and trails by \(0.09\) nats on the other events, where Inhib.\ Spline/Spline Hawkes has the best binned \(\log p(t)\).

\begin{table}[t]
  \caption{MIMIC-IV results split by whether the evaluated event was recorded in the same minute as the previous event (same, \(41.6\%\) of the evaluated events) or not (other). Same checkpoints and test sequences as Table~\ref{tab:full_mimic_results}. Binned: \(\log \Pr(t \in \text{bin})\) at one minute (Appendix~\ref{app:binned}). Mean \(\pm\) std over 5 seeds. \textbf{Bold}: best per column; \underline{underline}: best classical (Poisson or Hawkes) model; ties at the displayed precision are all marked.}
  \label{tab:mimic_ties}
  \centering
  \small
  \setlength{\tabcolsep}{3pt}
  \resizebox{\textwidth}{!}{%
  \begin{tabular}{lccccccccccc}
    \toprule
    & \multicolumn{2}{c}{\(\log p(t)\)} & \multicolumn{3}{c}{Binned \(\log p(t)\)} & \multicolumn{3}{c}{\(\log p(d|t)\)} & \multicolumn{3}{c}{Acc\(|t\) (\%)} \\
    \cmidrule(lr){2-3}\cmidrule(lr){4-6}\cmidrule(lr){7-9}\cmidrule(lr){10-12}
    Model & all & other & all & other & same & all & other & same & all & other & same \\
    \midrule
    Homogeneous Poisson & \(2.60_{\pm .01}\) & \(1.71_{\pm .04}\) & \(-4.90_{\pm .01}\) & \(-5.57_{\pm .04}\) & \(-3.97_{\pm .04}\) & \(-3.78_{\pm .02}\) & \(-3.91_{\pm .02}\) & \(-3.60_{\pm .02}\) & \(13.6_{\pm .0}\) & \(12.7_{\pm .0}\) & \(14.7_{\pm .0}\) \\
    Spline Poisson & \(2.64_{\pm .00}\) & \(1.79_{\pm .03}\) & \(-4.87_{\pm .00}\) & \(-5.49_{\pm .03}\) & \(-3.99_{\pm .03}\) & \(-3.71_{\pm .01}\) & \(-3.84_{\pm .01}\) & \(-3.53_{\pm .01}\) & \(13.6_{\pm .0}\) & \(12.7_{\pm .0}\) & \(14.7_{\pm .0}\) \\
    \midrule
    Const./Exp Hawkes & \(3.38_{\pm .00}\) & \(2.05_{\pm .00}\) & \(-4.13_{\pm .00}\) & \(-5.24_{\pm .00}\) & \(-2.57_{\pm .00}\) & \(-2.71_{\pm .00}\) & \(-3.12_{\pm .00}\) & \(-2.13_{\pm .00}\) & \(30.9_{\pm .1}\) & \(20.1_{\pm .1}\) & \(46.2_{\pm .2}\) \\
    Spline/Exp Hawkes & \(3.39_{\pm .00}\) & \(2.05_{\pm .00}\) & \(-4.12_{\pm .00}\) & \(-5.23_{\pm .00}\) & \(-2.57_{\pm .00}\) & \(-2.70_{\pm .00}\) & \(-3.12_{\pm .00}\) & \underline{\(-2.12_{\pm .00}\)} & \(30.9_{\pm .0}\) & \(20.1_{\pm .1}\) & \(46.2_{\pm .1}\) \\
    Const./Spline Hawkes & \(3.78_{\pm .00}\) & \(2.11_{\pm .00}\) & \(-3.73_{\pm .00}\) & \(-5.17_{\pm .00}\) & \(-1.71_{\pm .01}\) & \(-2.62_{\pm .00}\) & \(-2.95_{\pm .00}\) & \(-2.15_{\pm .00}\) & \(32.7_{\pm .0}\) & \(22.4_{\pm .0}\) & \(47.3_{\pm .0}\) \\
    Spline/Spline Hawkes & \(3.83_{\pm .00}\) & \(2.15_{\pm .00}\) & \(-3.67_{\pm .00}\) & \(-5.13_{\pm .00}\) & \(-1.61_{\pm .00}\) & \underline{\(-2.61_{\pm .00}\)} & \underline{\(-2.93_{\pm .00}\)} & \(-2.17_{\pm .00}\) & \(32.9_{\pm .1}\) & \(22.5_{\pm .1}\) & \(47.4_{\pm .1}\) \\
    \midrule
    Inhib. Const./Exp Hawkes & \underline{\(3.87_{\pm .00}\)} & \(2.10_{\pm .00}\) & \underline{\(-3.64_{\pm .00}\)} & \(-5.19_{\pm .00}\) & \underline{\(-1.47_{\pm .01}\)} & \(-2.67_{\pm .01}\) & \(-3.07_{\pm .01}\) & \underline{\(-2.12_{\pm .00}\)} & \(32.1_{\pm .2}\) & \(21.3_{\pm .3}\) & \(47.3_{\pm .1}\) \\
    Inhib. Spline/Spline Hawkes & \(3.83_{\pm .04}\) & \underline{\(\mathbf{2.24}_{\pm .01}\)} & \(-3.68_{\pm .04}\) & \underline{\(\mathbf{-5.04}_{\pm .02}\)} & \(-1.77_{\pm .07}\) & \(-2.68_{\pm .09}\) & \(-3.04_{\pm .12}\) & \(-2.19_{\pm .03}\) & \underline{\(33.5_{\pm .3}\)} & \underline{\(22.9_{\pm .5}\)} & \underline{\(48.3_{\pm .1}\)} \\
    \midrule
    NeuralConstTPP & \(3.08_{\pm .02}\) & \(2.18_{\pm .01}\) & \(-4.42_{\pm .02}\) & \(-5.10_{\pm .01}\) & \(-3.47_{\pm .05}\) & \(-2.65_{\pm .01}\) & \(-3.07_{\pm .01}\) & \(-2.06_{\pm .01}\) & \(32.6_{\pm .3}\) & \(19.5_{\pm .2}\) & \(51.1_{\pm .6}\) \\
    THP & \(3.26_{\pm .01}\) & \(2.11_{\pm .01}\) & \(-4.24_{\pm .01}\) & \(-5.16_{\pm .01}\) & \(-2.95_{\pm .03}\) & \(\mathbf{-2.36}_{\pm .01}\) & \(\mathbf{-2.87}_{\pm .00}\) & \(\mathbf{-1.65}_{\pm .01}\) & \(37.6_{\pm .1}\) & \(21.9_{\pm .1}\) & \(\mathbf{59.6}_{\pm .2}\) \\
    SAHP & \(3.66_{\pm .01}\) & \(2.00_{\pm .01}\) & \(-3.88_{\pm .01}\) & \(-5.28_{\pm .01}\) & \(-1.91_{\pm .02}\) & \(-2.74_{\pm .03}\) & \(-3.27_{\pm .03}\) & \(-2.00_{\pm .05}\) & \(35.3_{\pm .4}\) & \(21.0_{\pm .4}\) & \(55.6_{\pm .9}\) \\
    NHP & \(3.60_{\pm .01}\) & \(1.86_{\pm .01}\) & \(-3.92_{\pm .01}\) & \(-5.42_{\pm .01}\) & \(-1.80_{\pm .02}\) & \(-2.98_{\pm .01}\) & \(-2.97_{\pm .01}\) & \(-2.99_{\pm .02}\) & \(21.9_{\pm .2}\) & \(22.0_{\pm .2}\) & \(21.6_{\pm .7}\) \\
    \midrule
    IntensityFree & \(\mathbf{4.06}_{\pm .02}\) & \(2.05_{\pm .02}\) & \(-3.43_{\pm .02}\) & \(-5.23_{\pm .02}\) & \(-0.90_{\pm .02}\) & \(-2.61_{\pm .01}\) & \(-3.09_{\pm .01}\) & \(-1.93_{\pm .02}\) & \(34.3_{\pm .3}\) & \(19.1_{\pm .3}\) & \(55.7_{\pm .4}\) \\
    NeuralSplineTPP & \(4.05_{\pm .02}\) & \(2.06_{\pm .02}\) & \(\mathbf{-3.39}_{\pm .02}\) & \(-5.22_{\pm .02}\) & \(\mathbf{-0.83}_{\pm .02}\) & \(-2.39_{\pm .01}\) & \(-2.91_{\pm .01}\) & \(\mathbf{-1.65}_{\pm .01}\) & \(\mathbf{38.3}_{\pm .1}\) & \(\mathbf{23.1}_{\pm .1}\) & \(59.5_{\pm .2}\) \\
    \bottomrule
  \end{tabular}}
\end{table}

\paragraph{Flexibility and capacity.}
Table~\ref{tab:ksweep_mimic} sweeps the number of spline knots of Spline/Spline Hawkes and NeuralSplineTPP on MIMIC-IV from 10 to 400. For both families, \(\log p(t)\) is flat for \(K \geq 20\). The \(\log p(d|t)\) of Spline/Spline Hawkes decreases slightly with \(K\), and that of NeuralSplineTPP decreases strongly, with a large spread over seeds. The type-conditional advantage of NeuralSplineTPP is therefore largest at \(K = 10\). We attribute the decrease of \(\log p(d|t)\) with growing \(K\) to the higher variance of the ratio \(\lambda_d^*(t) / \sum_{d'} \lambda_{d'}^*(t)\) as the intensities become spikier.

\paragraph{Non-stationarity.}
An alternative explanation for the MIMIC-IV gap is a drift of the dynamics over the ICU stay, which a stationary kernel cannot represent. We therefore built non-stationary (NS) variants of Spline/Exp and Spline/Spline Hawkes whose kernel parameters are themselves linear splines over absolute time within the stay (\(K_{ns} \in \{3, 5\}\) knots). The cumulative intensity stays in closed form, and the model reduces exactly to the stationary kernel for constant heights. We trained 7 settings with 5 seeds each (Table~\ref{tab:ns_control}). No NS setting improves the joint log-likelihood over the stationary model with the nearest knot count by more than \(0.002\) nats. The best NS setting, selected on validation likelihood, lowers it by \(0.12\) nats and widens the gap to NeuralSplineTPP from \(0.45\) to \(0.57\) nats. The largest improvement of any component is \(0.03\) nats in \(\log p(t)\), for \(K_{\Delta t} = 12\) against \(K = 10\). Within-stay non-stationarity therefore does not explain the gap.

\begin{table}[ht]
  \caption{Knot-count sweep on all datasets: \(\log p(t)\,/\,\log p(d\mid t)\) (nats, mean over 5 seeds) for NeuralSplineTPP (head knots \(K\)) and Spline/Spline Hawkes (kernel knots \(K\), baseline fixed at 100 knots). The sweep runs are separate from the reported runs (20{,}000 steps on the standard benchmarks, learning rates without per-model adjustment), so values differ from the main tables.}
  \label{tab:ksweep_all}
  \centering
  \footnotesize
  \setlength{\tabcolsep}{3pt}
  \resizebox{\textwidth}{!}{%
  \begin{tabular}{rcccccc}
    \toprule
    \(K\) & Taxi & StackOverflow & Retweet & Amazon & Taobao & MIMIC-IV \\
    \midrule
    \multicolumn{7}{l}{\textit{NeuralSplineTPP}} \\
    10 & \(0.72/{-}0.21\) & \({-}0.73/{-}1.58\) & \({-}3.32/{-}0.78\) & \(1.63/{-}1.90\) & \(2.66/{-}1.43\) & \(3.99/{-}2.49\) \\
    20 & \(0.73/{-}0.22\) & \({-}0.72/{-}1.59\) & \({-}3.32/{-}0.78\) & \(1.41/{-}1.89\) & \(2.70/{-}1.45\) & \(3.99/{-}2.61\) \\
    50 & \(0.71/{-}0.23\) & \({-}0.73/{-}1.64\) & \({-}3.13/{-}0.79\) & \(2.22/{-}1.92\) & \(2.72/{-}1.47\) & \(3.97/{-}2.82\) \\
    100 & \(0.70/{-}0.24\) & \({-}0.74/{-}1.69\) & \({-}2.87/{-}0.78\) & \(2.43/{-}1.90\) & \(2.73/{-}1.49\) & \(3.97/{-}2.89\) \\
    200 & \(0.67/{-}0.26\) & \({-}0.75/{-}1.74\) & \({-}2.44/{-}0.79\) & \(2.53/{-}1.91\) & \(2.72/{-}1.54\) & \(3.95/{-}3.20\) \\
    400 & \(0.67/{-}0.27\) & \({-}0.76/{-}1.76\) & \({-}1.85/{-}0.79\) & \(2.56/{-}1.91\) & \(2.71/{-}1.60\) & \(3.98/{-}3.03\) \\
    \midrule
    \multicolumn{7}{l}{\textit{Spline/Spline Hawkes}} \\
    10 & \(0.48/{-}0.71\) & \({-}0.73/{-}1.60\) & \({-}3.36/{-}0.78\) & \(0.14/{-}1.98\) & \(2.66/{-}1.40\) & \(3.69/{-}2.60\) \\
    20 & \(0.49/{-}0.70\) & \({-}0.73/{-}1.60\) & \({-}3.36/{-}0.78\) & \(0.43/{-}1.99\) & \(2.67/{-}1.41\) & \(3.83/{-}2.61\) \\
    50 & \(0.48/{-}0.70\) & \({-}0.73/{-}1.62\) & \({-}3.29/{-}0.78\) & \(1.29/{-}2.04\) & \(2.67/{-}1.43\) & \(3.83/{-}2.62\) \\
    100 & \(0.48/{-}0.71\) & \({-}0.74/{-}1.65\) & \({-}3.14/{-}0.79\) & \(1.66/{-}2.03\) & \(2.67/{-}1.46\) & \(3.82/{-}2.64\) \\
    200 & \(0.49/{-}0.72\) & \({-}0.76/{-}1.68\) & \({-}2.84/{-}0.79\) & \(1.70/{-}2.04\) & \(2.66/{-}1.51\) & \(3.81/{-}2.65\) \\
    400 & \(0.49/{-}0.74\) & \({-}0.76/{-}1.71\) & \({-}2.37/{-}0.80\) & \(1.73/{-}2.04\) & \(2.64/{-}1.55\) & \(3.79/{-}2.68\) \\
    \bottomrule
  \end{tabular}}
\end{table}

\begin{table}[t]
  \caption{Knot-count sweep on MIMIC-IV: Spline/Spline Hawkes (kernel knots \(K\); baseline fixed at 100 knots) vs.\ NeuralSplineTPP (head knots \(K\)). Mean \(\pm\) std over 5 seeds. \(\log p(t)\) is flat in \(K\) for both families, and \(\log p(d|t)\) does not improve. \#params: trainable parameters of the Hawkes model and of the NeuralSplineTPP head (its Transformer encoder adds 208{,}832).}
  \label{tab:ksweep_mimic}
  \centering
  \small
  \setlength{\tabcolsep}{3pt}
  \resizebox{\textwidth}{!}{%
  \begin{tabular}{r rccc rccc}
    \toprule
    & \multicolumn{4}{c}{Spline/Spline Hawkes} & \multicolumn{4}{c}{NeuralSplineTPP} \\
    \cmidrule(lr){2-5}\cmidrule(lr){6-9}
    \(K\) & \#params & \(\log p(t)\) & \(\log p(d|t)\) & Acc\(|t\) (\%) & \#params & \(\log p(t)\) & \(\log p(d|t)\) & Acc\(|t\) (\%) \\
    \midrule
    10 & 148{,}591 & \(3.69_{\pm .00}\) & \(-2.60_{\pm .00}\) & \(32.5_{\pm .0}\) & 77{,}220 & \(3.99_{\pm .02}\) & \(-2.49_{\pm .11}\) & \(36.9_{\pm 1.3}\) \\
    20 & 285{,}481 & \(3.83_{\pm .00}\) & \(-2.61_{\pm .00}\) & \(32.9_{\pm .1}\) & 154{,}440 & \(3.99_{\pm .04}\) & \(-2.61_{\pm .21}\) & \(35.4_{\pm 2.7}\) \\
    50 & 696{,}151 & \(3.83_{\pm .00}\) & \(-2.62_{\pm .00}\) & \(32.8_{\pm .0}\) & 386{,}100 & \(3.97_{\pm .02}\) & \(-2.82_{\pm .11}\) & \(32.7_{\pm 1.4}\) \\
    100 & 1{,}380{,}601 & \(3.82_{\pm .00}\) & \(-2.64_{\pm .00}\) & \(32.8_{\pm .0}\) & 772{,}200 & \(3.97_{\pm .05}\) & \(-2.89_{\pm .37}\) & \(30.9_{\pm 7.4}\) \\
    200 & 2{,}749{,}501 & \(3.81_{\pm .00}\) & \(-2.65_{\pm .00}\) & \(32.7_{\pm .0}\) & 1{,}544{,}400 & \(3.95_{\pm .08}\) & \(-3.20_{\pm .50}\) & \(23.5_{\pm 10.0}\) \\
    400 & 5{,}487{,}301 & \(3.79_{\pm .00}\) & \(-2.68_{\pm .00}\) & \(32.6_{\pm .0}\) & 3{,}088{,}800 & \(3.98_{\pm .09}\) & \(-3.03_{\pm .46}\) & \(28.2_{\pm 9.2}\) \\
    \bottomrule
  \end{tabular}}
\end{table}

\begin{table}[t]
  \caption{Non-stationary Hawkes control on MIMIC-IV. The kernel heights of the NS variants are themselves linear splines over absolute stay time with \(K_{ns}\) knots (\(K_{ns}=1\): the stationary model); \(K_{\Delta t}\): knots of the kernel over elapsed time. Mean \(\pm\) std over 5 seeds. The NS runs use the configuration of Appendix~\ref{app:hyperparams_final} with learning rate 0.05 for the NS Spline/Spline kernels. The stationary Spline/Spline rows are taken from the knot sweep (Table~\ref{tab:ksweep_mimic}). At \(K_{\Delta t} = 12\), their nearest counterparts are \(K = 10\) and \(20\) there. No non-stationary cell improves the joint log-likelihood of its nearest stationary counterpart by more than 0.002 nats, and all remain below NeuralSplineTPP (\(\log p(t) = 4.05\), \(\log p(d|t) = -2.39\)).}
  \label{tab:ns_control}
  \centering
  \small
  \begin{tabular}{lrrccc}
    \toprule
    Model & \(K_{\Delta t}\) & \(K_{ns}\) & \(\log p(t)\) & \(\log p(d|t)\) & Acc\(|t\) (\%) \\
    \midrule
    Spline/Exp Hawkes & -- & 1 & \(3.39_{\pm .00}\) & \(-2.70_{\pm .00}\) & \(30.9_{\pm .0}\) \\
    \quad + NS kernel & -- & 5 & \(3.36_{\pm .00}\) & \(-2.69_{\pm .00}\) & \(31.1_{\pm .0}\) \\
    \midrule
    Spline/Spline Hawkes + NS kernel & 12 & 3 & \(3.72_{\pm .00}\) & \(-2.63_{\pm .00}\) & \(32.9_{\pm .0}\) \\
    Spline/Spline Hawkes + NS kernel & 12 & 5 & \(3.71_{\pm .00}\) & \(-2.63_{\pm .00}\) & \(33.0_{\pm .0}\) \\
    Spline/Spline Hawkes & 20 & 1 & \(3.83_{\pm .00}\) & \(-2.61_{\pm .00}\) & \(32.9_{\pm .1}\) \\
    \quad + NS kernel & 20 & 3 & \(3.76_{\pm .00}\) & \(-2.66_{\pm .00}\) & \(32.9_{\pm .1}\) \\
    \quad + NS kernel & 20 & 5 & \(3.74_{\pm .00}\) & \(-2.66_{\pm .00}\) & \(32.9_{\pm .1}\) \\
    Spline/Spline Hawkes & 100 & 1 & \(3.82_{\pm .00}\) & \(-2.64_{\pm .00}\) & \(32.8_{\pm .0}\) \\
    \quad + NS kernel & 100 & 3 & \(3.73_{\pm .00}\) & \(-2.67_{\pm .00}\) & \(32.9_{\pm .0}\) \\
    \quad + NS kernel & 100 & 5 & \(3.71_{\pm .00}\) & \(-2.68_{\pm .00}\) & \(32.9_{\pm .0}\) \\
    \bottomrule
  \end{tabular}
\end{table}

\FloatBarrier
\section{Calibration}
\label{app:calibration}
As a goodness-of-fit check beyond likelihood and accuracy, Table~\ref{tab:calibration} reports time-rescaling calibration~\citep{brown2002timerescaling} and the calibration of the type distribution for four representative models with a closed-form cumulative intensity (so that the compensator is exact). The flexible models are best calibrated in time on four of the five standard datasets. On MIMIC-IV, NeuralSplineTPP is well calibrated in time, whereas the other models are not.

\begin{table}[t]
  \caption{Calibration of four representative models with a closed-form cumulative intensity (seed 0). Each cell is the time-rescaling Kolmogorov--Smirnov distance \(\sup|F_{\mathrm{emp}}(u) - u|\) of the probability integral transform \(u_i = 1 - \exp(-\Lambda^*(t_{i-1}, t_i))\) (events with a zero inter-event time excluded) and the top-label expected calibration error of \(p(d \mid t, \mathcal{H})\) (15 bins); lower is better for both. Up to 10{,}000 test events per cell (MIMIC-IV: 5{,}000). For Spline/Spline Hawkes on StackOverflow, the evaluated checkpoint uses learning rate 0.1 instead of the adjusted rate of Table~\ref{tab:joint_ll_results}.}
  \label{tab:calibration}
  \centering
  \small
  \setlength{\tabcolsep}{3pt}
  \resizebox{\textwidth}{!}{%
  \begin{tabular}{lcccccc}
    \toprule
    Model (KS / ECE) & Taxi & StackOverflow & Retweet & Amazon & Taobao & MIMIC-IV \\
    \midrule
    Const./Exp Hawkes & .151 / .274 & .033 / .023 & .060 / .008 & .267 / .045 & .063 / .022 & .315 / .040 \\
    Spline/Spline Hawkes & .128 / .221 & .027 / .025 & .111 / .021 & .137 / .018 & .035 / .020 & .176 / .056 \\
    NeuralConstTPP & .057 / .007 & .066 / .024 & .048 / .021 & .344 / .012 & .343 / .015 & .395 / .031 \\
    NeuralSplineTPP & .040 / .008 & .009 / .027 & .083 / .009 & .025 / .015 & .037 / .016 & .021 / .024 \\
    \bottomrule
  \end{tabular}}
\end{table}

\FloatBarrier
\newpage


\begin{thebibliography}{50}

\bibitem[Bacry et~al.(2018)]{bacry2017tick}
Emmanuel Bacry, Martin Bompaire, Philip Deegan, St{\'e}phane Ga{\"i}ffas, and S{\o}ren V. Poulsen.
\newblock Tick: a Python library for statistical learning, with an emphasis on Hawkes processes and time-dependent models.
\newblock \emph{Journal of Machine Learning Research}, 18(214):1--5, 2018.

\bibitem[Bosser and Ben~Taieb(2023)]{bosser2023predictive}
Tanguy Bosser and Souhaib Ben~Taieb.
\newblock On the Predictive Accuracy of Neural Temporal Point Process Models for Continuous-time Event Data.
\newblock \emph{Transactions on Machine Learning Research}, 2023.

\bibitem[Br{\'e}maud and Massouli{\'e}(1996)]{bremaud1996nonlinear}
Pierre Br{\'e}maud and Laurent Massouli{\'e}.
\newblock Stability of nonlinear Hawkes processes.
\newblock \emph{The Annals of Probability}, 24(3):1563--1588, 1996.

\bibitem[Brown et~al.(2002)]{brown2002timerescaling}
Emery N. Brown, Riccardo Barbieri, Valentina Ventura, Robert E. Kass, and Loren M. Frank.
\newblock The time-rescaling theorem and its application to neural spike train data analysis.
\newblock \emph{Neural Computation}, 14(2):325--346, 2002.

\bibitem[Chang et~al.(2025)]{S2P2}
Yuxin Chang, Alex Boyd, Cao Xiao, Taha Kass-Hout, Parminder Bhatia, Padhraic Smyth, and Andrew Warrington.
\newblock Deep Continuous-Time State-Space Models for Marked Event Sequences.
\newblock \emph{Advances in Neural Information Processing Systems}, 2025.

\bibitem[Du et~al.(2016)]{RMTPP}
Nan Du, Hanjun Dai, Rakshit Trivedi, Utkarsh Upadhyay, Manuel Gomez-Rodriguez, and Le Song.
\newblock Recurrent Marked Temporal Point Processes: Embedding Event History to Vector.
\newblock \emph{Proceedings of the 22nd ACM SIGKDD International Conference on Knowledge Discovery and Data Mining}, pages 1555--1564, 2016.

\bibitem[Goldberger et~al.(2000)]{goldberger2000physionet}
Ary L. Goldberger, Luis A. N. Amaral, Leon Glass, Jeffrey M. Hausdorff, Plamen Ch. Ivanov, Roger G. Mark, Joseph E. Mietus, George B. Moody, Chung-Kang Peng, and H. Eugene Stanley.
\newblock PhysioBank, PhysioToolkit, and PhysioNet: Components of a new research resource for complex physiologic signals.
\newblock \emph{Circulation}, 101(23):e215--e220, 2000.

\bibitem[Hawkes(1971)]{hawkes1971spectra}
A.~G. Hawkes.
\newblock Spectra of some self-exciting and mutually exciting point processes.
\newblock \emph{Biometrika}, 58(1):83--90, 1971.

\bibitem[Isham and Westcott(1979)]{SelfCorrecting}
Valerie Isham and Mark Westcott.
\newblock A self-correcting point process.
\newblock \emph{Stochastic Processes and their Applications}, 8:335--347, 1979.

\bibitem[Jiang et~al.(2025)]{DanBench}
Yue Jiang, Jichu Li, Yang Liu, Dingkang Yang, Feng Zhou, and Quyu Kong.
\newblock Danmaku{TPPB}ench: A Multi-modal Benchmark for Temporal Point Process Modeling and Understanding.
\newblock \emph{Advances in Neural Information Processing Systems, Datasets and Benchmarks Track}, 2025.

\bibitem[Johnson et~al.(2023)]{mimic-iv}
Alistair E. W. Johnson, Lucas Bulgarelli, Lu Shen, Alvin Gayles, Ayad Shammout, Steven Horng, Tom J. Pollard, Sicheng Hao, Benjamin Moody, Brian Gow, Li-wei H. Lehman, Leo A. Celi, and Roger G. Mark.
\newblock MIMIC-IV, a freely accessible electronic health record dataset.
\newblock \emph{Scientific Data}, 10(1):1, 2023.

\bibitem[Johnson et~al.(2024)]{mimiciv31}
Alistair Johnson, Lucas Bulgarelli, Tom Pollard, Brian Gow, Benjamin Moody, Steven Horng, Leo Anthony Celi, and Roger Mark.
\newblock MIMIC-IV (version 3.1).
\newblock PhysioNet, 2024.
\newblock \url{https://doi.org/10.13026/kpb9-mt58}.

\bibitem[Kerrigan et~al.(2026)]{EventFlow}
Gavin Kerrigan, Kai Nelson, and Padhraic Smyth.
\newblock EventFlow: Forecasting Temporal Point Processes with Flow Matching.
\newblock \emph{International Conference on Artificial Intelligence and Statistics (AISTATS)}, 2026.

\bibitem[Leskovec and Krevl(2014)]{leskovec2014snap}
Jure Leskovec and Andrej Krevl.
\newblock SNAP Datasets: Stanford Large Network Dataset Collection.
\newblock 2014.
\newblock \url{http://snap.stanford.edu/data}.

\bibitem[Li et~al.(2020)]{behrt}
Yikuan Li, Shishir Rao, Jose Roberto Ayala Solares, Abdelaali Hassaine, Rema Ramakrishnan, Dexter Canoy, Yajie Zhu, Kazem Rahimi, and Gholamreza Salimi-Khorshidi.
\newblock {BEHRT}: Transformer for Electronic Health Records.
\newblock \emph{Scientific Reports}, 10:7155, 2020.

\bibitem[Liguori et~al.(2025)]{TPPSurvey}
Angelica Liguori, Luciano Caroprese, Marco Minici, Bruno Veloso, Francesco Spinnato, Mirco Nanni, Giuseppe Manco, and Jo\~{a}o Gama.
\newblock Modeling events and interactions through temporal processes: {A} survey.
\newblock \emph{Neurocomputing}, 653:131191, 2025.

\bibitem[Lin et~al.(2022)]{DiffusionTPP}
Haitao Lin, Lirong Wu, Guojiang Zhao, Pai Liu, and Stan Z. Li.
\newblock Exploring generative neural temporal point process.
\newblock \emph{Transactions on Machine Learning Research}, 2022.

\bibitem[Lin et~al.(2025)]{lin2025survey}
Haitao Lin, Cheng Tan, Lirong Wu, Zicheng Liu, Zhangyang Gao, and Stan Z. Li.
\newblock An Extensive Survey With Empirical Studies
on Deep Temporal Point Process.
\newblock \emph{IEEE Transactions on Knowledge and Data Engineering},
37(4):1599--1619, 2025.

\bibitem[L{\"u}dke et~al.(2023)]{ludke2023addthin}
David L{\"u}dke, Marin Bilo{\v{s}}, Oleksandr Shchur, Marten Lienen,
and Stephan G{\"u}nnemann.
\newblock Add and Thin: Diffusion for Temporal Point Processes.
\newblock \emph{Advances in Neural Information Processing Systems}, 2023.

\bibitem[L{\"u}dke et~al.(2026)]{ludke2026editpp}
David L{\"u}dke, Marten Lienen, Marcel Kollovieh, and Stephan G{\"u}nnemann.
\newblock Edit-Based Flow Matching for Temporal Point Processes.
\newblock \emph{International Conference on Learning Representations}, 2026.

\bibitem[Mei and Eisner(2017)]{nhp2017}
Hongyuan Mei and Jason Eisner.
\newblock The Neural Hawkes Process: A Neurally Self-Modulating Multivariate Point Process.
\newblock \emph{Advances in Neural Information Processing Systems}, 2017.

\bibitem[Mukherjee et~al.(2026)]{mukherjee2026adiff4tpp}
Amartya Mukherjee, Ruizhi Deng, He Zhao, Yuzhen Mao, Leonid Sigal,
and Frederick Tung.
\newblock {ADiff4TPP}: Asynchronous Diffusion Models
for Temporal Point Processes.
\newblock \emph{Transactions on Machine Learning Research}, 2026.

\bibitem[Ni et~al.(2019)]{ni2019amazon}
Jianmo Ni, Jiacheng Li, and Julian McAuley.
\newblock Justifying Recommendations using Distantly-Labeled Reviews and Fine-Grained Aspects.
\newblock In \emph{Proceedings of EMNLP-IJCNLP}, pages 188--197, 2019.

\bibitem[Ogata(1981)]{ogata1981lewis}
Yosihiko Ogata.
\newblock On Lewis' simulation method for point processes.
\newblock \emph{IEEE Transactions on Information Theory}, 27(1):23--31, 1981.

\bibitem[Ogata(1988)]{ogata1988earthquake}
Yosihiko Ogata.
\newblock Statistical models for earthquake occurrences and residual analysis for point processes.
\newblock \emph{Journal of the American Statistical Association}, 83(401):9--27, 1988.

\bibitem[Omi et~al.(2019)]{omi2019fullynn}
Takahiro Omi, Naonori Ueda, and Kazuyuki Aihara.
\newblock Fully Neural Network based Model
for General Temporal Point Processes.
\newblock \emph{Advances in Neural Information Processing Systems}, 201

\bibitem[Rasmy et~al.(2021)]{medbert}
Laila Rasmy, Yang Xiang, Ziqian Xie, Cui Tao, and Degui Zhi.
\newblock {Med-BERT}: pretrained contextualized embeddings on large-scale structured electronic health records for disease prediction.
\newblock \emph{npj Digital Medicine}, 4(1):86, 2021.

\bibitem[Renc et~al.(2024)]{ethos}
Pawe{\l} Renc, Yugang Jia, Anthony E. Samir, Jaros{\l}aw W{\k{a}}s, Quanzheng Li, David W. Bates, and Arkadiusz Sitek.
\newblock Zero shot health trajectory prediction using transformer.
\newblock \emph{npj Digital Medicine}, 7:256, 2024.

\bibitem[Shchur et~al.(2020a)]{IntensityFree}
Oleksandr Shchur, Marin Bilo{\v{s}}, and Stephan G{\"u}nnemann.
\newblock Intensity-Free Learning of Temporal Point Processes.
\newblock \emph{International Conference on Learning Representations}, 2020.

\bibitem[Shchur et~al.(2020b)]{shchur2020tritpp}
Oleksandr Shchur, Nicholas Gao, Marin Bilo{\v{s}}, and Stephan G{\"u}nnemann.
\newblock Fast and Flexible Temporal Point Processes
with Triangular Maps.
\newblock \emph{Advances in Neural Information Processing Systems}, 2020.

\bibitem[Shchur et~al.(2021)]{shchur2021review}
Oleksandr Shchur, Ali Caner T{\"u}rkmen, Tim Januschowski, and Stephan G{\"u}nnemann.
\newblock Neural Temporal Point Processes: A Review.
\newblock In \emph{Proceedings of the 30th International Joint Conference on Artificial Intelligence (IJCAI)}, pages 4585--4593, 2021.

\bibitem[Shmatko et~al.(2025)]{Delphi}
Artem Shmatko, Alexander Wolfgang Jung, Kumar Gaurav, S{\o}ren Brunak, Laust Hvas Mortensen, Ewan Birney, Tom Fitzgerald, and Moritz Gerstung.
\newblock Learning the natural history of human disease with generative transformers.
\newblock \emph{Nature}, 647:248--256, 2025.

\bibitem[Soen et~al.(2021)]{soen2021unipoint}
Alexander Soen, Alexander Mathews, Daniel Grixti-Cheng, and Lexing Xie.
\newblock {UNIPoint}: Universally Approximating
Point Processes Intensities.
\newblock \emph{Proceedings of the AAAI Conference
on Artificial Intelligence}, 35(11):9685--9694, 2021.

\bibitem[Song et~al.(2024)]{song2024decoupled}
Yujee Song, Donghyun Lee, Rui Meng, and Won Hwa Kim.
\newblock Decoupled Marked Temporal Point Process
using Neural Ordinary Differential Equations.
\newblock \emph{International Conference on Learning Representations},

\bibitem[Steinberg et~al.(2024)]{MOTOR}
Ethan Steinberg, Jason Fries, Yizhe Xu, and Nigam Shah.
\newblock MOTOR: A Time-to-Event Foundation Model For Structured Medical Records.
\newblock \emph{International Conference on Learning Representations}, 2024.

\bibitem[Sudlow et~al.(2015)]{sudlow2015ukb}
Cathie Sudlow, John Gallacher, Naomi Allen, Valerie Beral, Paul Burton, John Danesh, Paul Downey, Paul Elliott, Jane Green, Martin Landray, Bette Liu, Paul Matthews, George Ong, Jill Pell, Alan Silman, Alan Young, Tom Sprosen, Tim Peakman, and Rory Collins.
\newblock UK Biobank: an open access resource for identifying the causes of a wide range of complex diseases of middle and old age.
\newblock \emph{PLOS Medicine}, 12(3):e1001779, 2015.

\bibitem[Whong(2014)]{whong2014taxi}
Chris Whong.
\newblock FOILing NYC's Taxi Trip Data.
\newblock 2014.
\newblock \url{https://chriswhong.com/open-data/foil_nyc_taxi/}.

\bibitem[Wornow et~al.(2023)]{EHRSHOT}
Michael Wornow, Rahul Thapa, Ethan Steinberg, Jason Fries, and Nigam Shah.
\newblock {EHRSHOT}: An {EHR} Benchmark for Few-Shot Evaluation of Foundation Models.
\newblock \emph{Advances in Neural Information Processing Systems, Datasets and Benchmarks Track}, 2023.

\bibitem[Xiao et~al.(2017)]{xiao2017intensity}
Shuai Xiao, Junchi Yan, Xiaokang Yang, Hongyuan Zha, and Stephen M. Chu.
\newblock Modeling the Intensity Function of Point Process
via Recurrent Neural Networks.
\newblock \emph{Proceedings of the AAAI Conference
on Artificial Intelligence}, 31(1), 2017.

\bibitem[Xue et~al.(2022)]{xue2022hypro}
Siqiao Xue, Xiaoming Shi, James Y. Zhang, and Hongyuan Mei.
\newblock HYPRO: A Hybridly Normalized Probabilistic Model for Long-Horizon Prediction of Event Sequences.
\newblock \emph{Advances in Neural Information Processing Systems}, 2022.

\bibitem[Xue et~al.(2024)]{xue2024easytpp}
Siqiao Xue, Xiaoming Shi, Zhixuan Chu, Yan Wang, Hongyan Hao, Fan Zhou, Caigao Jiang, Chen Pan, James Y. Zhang, Qingsong Wen, Jun Zhou, and Hongyuan Mei.
\newblock EasyTPP: Towards Open Benchmarking Temporal Point Processes.
\newblock \emph{International Conference on Learning Representations}, 2024.

\bibitem[Yang et~al.(2022)]{AttNHP}
Chenghao Yang, Hongyuan Mei, and Jason Eisner.
\newblock Transformer Embeddings of Irregularly Spaced Events and Their Participants.
\newblock \emph{International Conference on Learning Representations}, 2022.

\bibitem[Zhang et~al.(2020)]{SAHP}
Qiang Zhang, Aldo Lipani, Omer Kirnap, and Emine Yilmaz.
\newblock Self-Attentive Hawkes Processes.
\newblock \emph{International Conference on Machine Learning}, 2020.

\bibitem[Zhao and Engelhard(2025)]{zhao2025embedded}
Yuankang Zhao and Matthew M. Engelhard.
\newblock Balancing Interpretability and Flexibility in Modeling Diagnostic Trajectories with an Embedded Neural Hawkes Process Model.
\newblock In \emph{Machine Learning for Healthcare}, volume 298 of \emph{Proceedings of Machine Learning Research}, 2025.

\bibitem[Zhou et~al.(2013)]{zhou2013triggering}
Ke Zhou, Hongyuan Zha, and Le Song.
\newblock Learning Triggering Kernels for Multi-dimensional Hawkes Processes.
\newblock \emph{Proceedings of the 30th International Conference on Machine Learning}, pages 1301--1309, 2013.

\bibitem[Zhou and Yu(2023)]{zhou2023autoint}
Zihao Zhou and Rose Yu.
\newblock Automatic Integration for Fast and Interpretable
Neural Point Processes.
\newblock In \emph{Learning for Dynamics and Control Conference}, volume~211 of
\emph{Proceedings of Machine Learning Research}, 2023.

\bibitem[Zuo et~al.(2020)]{THP}
Simiao Zuo, Haoming Jiang, Zichong Li, Tuo Zhao, and Hongyuan Zha.
\newblock Transformer Hawkes Process.
\newblock \emph{International Conference on Machine Learning}, 2020.

\end{thebibliography}
\end{document}